\documentclass[3p, 11.0pt, a4paper]{elsarticle}

\usepackage{caption}
\usepackage{float}
\usepackage{multirow}
\usepackage{subcaption}
\usepackage[utf8]{inputenc} 
\usepackage{acronym}
\usepackage{setspace}
\usepackage{multicol}
\usepackage{amsmath}
\usepackage{amsfonts}
\usepackage{amssymb}
\usepackage{url}
\usepackage{color}
\usepackage{makecell}
\usepackage{amssymb}
\usepackage{amsthm}
\usepackage{subcaption}
\usepackage{booktabs}
\newtheorem{definition}{Definition}
\newtheorem{proposition}{Proposition}
\usepackage[figuresright]{rotating}
\usepackage{bbm}
\usepackage{amsmath}
\usepackage{mathtools}
\usepackage{algorithm, algorithmicx, algpseudocode}
\usepackage[T1]{fontenc}

\begin{document}

    \onehalfspacing	
	
	\begin{frontmatter}

\title{Data-driven Preference Learning Methods\\ for Sorting Problems with~Multiple Temporal Criteria}
		
\author[1]{Yijun Li \fnref{fn1}}
\address[1]{School of Data Science, City University of Hong Kong, 83 Tat Chee Avenue, Kowloon Tong, Kowloon, Hong Kong SAR}
\author[2]{Mengzhuo Guo \corref{cor1} \fnref{fn1}}
\address[2]{Business School, Sichuan University, No.24 South Section 1, Yihuan Road, Chengdu, China, 610065}
\cortext[cor1]{Corresponding author, Email address: mengzhguo@scu.edu.cn}
\fntext[fn1]{Both authors contributed equally to this work.}
\author[4]{Mi{\l}osz Kadzi{\'n}ski}
\address[4]{Institute of Computing Science, Poznan University of Technology, Piotrowo 2, 60-965 Pozna\'n, Poland}
\author[3]{Qingpeng Zhang}
\address[3]{Musketeers Foundation Institute of Data Science and LKS Faculty of Medicine, The University of Hong Kong, Hong Kong, China}
		
		\begin{abstract}
        \noindent We present novel preference learning approaches for sorting problems with multiple temporal criteria. They leverage an additive value function as the basic preference model, adapted for accommodating time series data. Given assignment examples concerning reference alternatives, we learn such a model using convex quadratic programming. It is characterized by fixed-time discount factors and operates within a regularization framework. This approach enables the consideration of temporal interdependencies between timestamps while mitigating the risk of overfitting. To enhance scalability and accommodate learnable time discount factors, we introduce a novel monotonic Recurrent Neural Network (mRNN). It captures the evolving dynamics of preferences over time, while upholding critical properties inherent to MCS problems. These include criteria monotonicity, preference independence, and the natural ordering of classes. The proposed mRNN can describe the preference dynamics by depicting piecewise linear marginal value functions and personalized time discount factors along with time. Thus, it effectively combines the interpretability of traditional sorting methods with the predictive potential offered by deep preference learning models. We comprehensively assess the proposed models on synthetic data scenarios and a real-case study centered on classifying valuable users within a mobile gaming app based on their historical in-game behavioral sequences. Empirical findings underscore the notable performance improvements achieved by the proposed models when compared to a spectrum of baseline methods, spanning machine learning, deep learning, and conventional multiple criteria sorting approaches. 
        \end{abstract}
		
		\end{frontmatter}

\section{Introduction}

\noindent The realm of Multiple Criteria Sorting (MCS) problems, also known as ordinal classification, pertains to evaluating alternatives based on their performance across multiple criteria. Subsequently, these alternatives are categorized into pre-defined classes that inherently embody a preference order~\citep{roy1996multicriteria}. This field has emerged as pivotal within Multiple Criteria Decision Aiding (MCDA)~\citep{dyer1992multiple,wallenius2008multiple}. It finds widespread use across a diverse spectrum of applications, including nanotechnology assessments \citep{kadzinski2018co}, consumer preference analysis \citep{guo2020consumer}, credit rating \citep{doumpos2019multicriteria}, and estimating migration potential \citep{arandarenko2020multiple}.

A contemporary trend in sorting problems has emerged, necessitating the Decision Makers' (DMs') involvement in expressing preferences through assignment examples. Subsequently, such holistic preferences are utilized to construct a~compatible preference model within a~preference disaggregation framework, as outlined in, e.g., \citep{devaud1980utadis,zopounidis2002multicriteria}. Thus inferred model can be subsequently employed to assign non-reference alternatives using a value-driven threshold-based sorting procedure.

The underlying idea in data-driven preference discovery or learning is similar. That is, a preference model is autonomously constructed using a batch of historical decisions (training samples) to predict outcomes for new data samples \citep{furnkranz2010preference}. However, diverging from traditional MCDA methods for preference learning, these approaches leverage the capabilities of information technology to manage high-dimensional data. Consequently, their primary focus revolves around the ability to capture intricate patterns between input and output data and to excel in the realm of scalable decision-making \citep{corrente2013robust}. Notably, these methods accommodate non-monotonic criteria \citep{ghaderi2017linear,guo2019progressive,liu2019preference}, consider interactions between criteria \citep{guo2021deciphering,guo2021hybrid,liu2021data}, and are adaptable to extracting criteria from user-generated content \citep{guo2020consumer,liu2023physician, wu2021modeling}. This flexibility renders them well-suited to capture the complexities inherent in real-world decision-making scenarios.

While data-driven preference learning methods have made significant strides in various decision problems, they face limitations in handling scenarios where the decision criteria involve time series data. Incorporating temporal criteria enables DMs to discern preference-changing patterns over time, a crucial aspect in numerous real-world applications. For instance, in credit risk evaluation, a DM may need to assess consumer default risk based on historical behavioral data encompassing purchase and loan repayment records \citep{babaev2019rnn, lin2021online}. Also, time series contribute to developing robust long-term electricity demand forecasts in energy forecasting, as exemplified in \citep{angelopoulos2019disaggregating}. Further, continuous recording of a patient's health status in the healthcare sector generates a temporal data series. This chronicle of information unveils patterns in readmission and elucidates the behavioral trajectories associated with chronic diseases \citep{lekwijit2023evaluating}. The temporal dimension in these instances proves instrumental in extracting valuable insights and enhancing predictive modeling.

To preserve temporal information in MCDA problems, one approach is to apply traditional preference learning models directly to time series. For example, \citet{angelopoulos2019disaggregating} assume each time series has a quantitative impact on the decision and evaluates time series by a value-based disaggregation model. However, this approach neglects crucial temporal information, such as the fact that yesterday's behaviors may be more relevant to the current decision than those from a week ago. Alternatively, fixed time discount factors or some statistics, such as the time series average, tendency, and seasonality, are introduced to account for the varying importance of nearby behaviors. For instance, \citet{campello2023improving} use descriptive statistics to study the temporal criteria impacts on a multi-objective preference learning model.

This paper tackles the MCS problems where the criteria are allowed to be time series, hence further bridging the gap between MCDA and temporal criteria. We aim to address the flaws of the existing approaches in this stream by treating performance for each timestamp as a criterion associated with a time discount factor. The adopted preference model comprises a collection of additive piecewise linear value functions. However, using such functions to approximate each timestamp's contribution to the alternative's comprehensive value would result in a highly complex preference model that would be difficult to interpret. Moreover, while excelling in replicating the provided assignment examples, it would exhibit poor predictive performance when applied to non-reference alternatives. This phenomenon is commonly referred to as over-fitting \citep{liu2019preference}. Also, as data volume and time series length increase, along with the inclusion of learnable time discount factors, such a model would demand substantial storage space and computational resources. This resource requirement can become particularly prohibitive when dealing with large datasets and long time series. Therefore, to facilitate modeling performance values at each timestamp of the criteria, we partition them into multiple sub-intervals delineated by characteristic points and propose dedicated preference learning algorithms. 

Our methodological contribution is two-fold. First, we introduce a convex quadratic programming model, named the Temporal Preference Learning (TPL) model, to approximate additive piecewise linear functions for each timestamp of the criteria. This model operates within a regularization framework, effectively addressing the challenge of over-fitting. It applies a fixed discount factor to consecutive timestamps, accounting for the importance of nearby timestamps. While the optimization problem at hand can be directly solved, it bears the risk of incurring substantial computational costs as the scale of the data increases.

Second, we extend the proposed mathematical programming model by modifying the a \textit{monotonic recurrent neural network} (mRNN) to suit the MCS problem context with preference-ordered classes and accommodate learnable discount factors. The resulting model can adaptively learn personalized time discount factors based on individual data sample characteristics, hence capturing the dynamics of changing preferences over time. Furthermore, the mRNN retains the intuitive and comprehensible nature of MCDA approaches. It accomplishes this by elucidating changes in time discount factors and by portraying obtained marginal value functions that adhere to pre-defined properties like monotonicity and preference independence.

Our most essential application-oriented contribution consists of using the proposed methods in a real-world problem in the gaming industry. Specifically, we distinguish high- and low-value mobile game users based on their historical behavior sequences, i.e., in-game purchases and action records. The study provides an in-depth analysis of individual users' evolving preferences and the learned time discount factors. In this way, we confirm that the proposed methodologies facilitate a deeper understanding of users' evolving behavior patterns for the DM.

We also rigorously evaluate the performance of the proposed models and a~spectrum of baseline models in terms of their predictive performance. The considered approaches involve machine learning, deep learning, and conventional MCDA approaches. The results -- quantified in terms of accuracy, precision, recall, and F-score -- are further verified on synthetic data with various complexities of time series, different characteristics of data generators, and assumptions underlying an assumed value-based model.

\section{Related Work}

\subsection{Generalizations of Value-based Preference Learning Methods}

\noindent Value-based preference learning methods have gained prominence due to their ability to disaggregate indirect preferences, reducing the cognitive burden on the DM, and using an intuitive preference model~\citep{corrente2013robust}. Extensive research in this domain has addressed various aspects, including robust recommendations~\citep{GRECO20101455,kadzinski2014preferential}, representative preference model selection~\citep{greco2011selection}, and contingent preference disaggregation~\citep{kadzinski2020contingent,liu2023modeling}.

Typically, these methods assumed that criteria are static, monotonic, and preference-independent, simplifying the description of decision problems. Recent advancements have focused on relaxing these premises to extend the applicability of value-based MCDA methods. One avenue of research involves allowing the marginal value functions to be non-monotonic. For instance, \citet{ghaderi2017linear} introduced a~linear fractional programming model to account for preference inflections. \citet{guo2019progressive} utilized a~progressive algorithm to adaptively adjust non-monotonic marginal value functions by constraining the variations of their slopes. Researchers have also incorporated regularization frameworks to automatically smooth marginal value functions for approximating piecewise linear functions using efficient learning algorithms \citep{liu2019preference}. Additionally, \citet{guo2021hybrid} proposed a~hybrid model that permits the preference model to be expressed as polynomials, enabling a machine learning-enabled, non-monotonic, and expressive MCDA approach for complex real-world applications.

In scenarios involving interacting criteria, \citet{angilella2014musa} considered both positive and negative interactions, including an additional term to the comprehensive value. Within the framework of robust ordinal regression, bonus and penalty values were introduced for the same purpose~\citep{greco2014robust}. \citet{guo2021deciphering} developed a model based on \textit{factorization machines} to manually model criteria interactions, proving effective in capturing flexible marginal value function patterns. Recent work addressed the challenges posed by data-intensive tasks with interacting criteria by incorporating a regularization scheme into value-based MCDA methods \citep{liu2021data}.

The advanced models mentioned above have been designed for static criteria, where performance values remain constant over time. In contrast, handling time series criteria introduces temporal dynamics into decision-making. Some efforts have been made in this direction, such as \citet{yan2015dynamic} analyzing trends in alternatives by aggregating data across periods and \citet{banamar2019interpolation} extending the PROMETHEE method to incorporate the importance of temporality. Further, \citet{angelopoulos2019disaggregating} have assumed that each time series can output a quantitative measure to the decision and used a preference disaggregation framework to obtain robust forecasts for long-term electricity demand. \citet{thesari2019decision} have collected statistical data about public resource distribution to determine a timeline of budget data and employed a classic MCDA approach to analyze long-term manager behaviors. Such forecasting methods have also been employed to predict future rankings under uncertainty \citep{campello2022dealing}. A recent approach uses descriptive statistics measuring the time series trend and seasonality to help understand the DM's preference in the long term under a preference disaggregation scheme \citep{campello2023improving}. Finally, a variant of the TOPSIS method with a tensorial formulation that embeds time as the third dimension has been proposed to capture temporal information in criteria \citep{campello2023exploiting}.

In summary, the developments mentioned above either treat the time series directly as the criteria values or use some manual statistics to describe their temporal characteristics. The first stream of methods may underscore the temporal relationship. In turn, the other stream may fail to obtain explicit time-dependent patterns of preferences, which is a crucial advantage of value function-based MCDA methods~\citep{GRECO20101455}. We aim to address these shortcomings in the paper.

\subsection{Deep Preference Learning in Multiple Criteria Decision Aiding}

\noindent Artificial neural networks (ANN) and their efficient learning algorithms, known as deep learning (DL), have found increasing adoption in business analytics and operational research due to their remarkable performance in predicting high-dimensional and scalable data \citep{kraus2020deep}. For instance, in the context of MCDA, researchers have explored the application of feedforward neural networks in addressing decision problems within a supervised learning framework \citep{wang1992feedforward}. These studies have introduced various desirable properties related to multiattribute preference models. In other instances, rather than adhering to multi-attribute value theory, \citet{hu2009bankruptcy} designed an outranking-based single-layer perceptron for bankruptcy prediction. They employed a genetic algorithm to determine the parameters of the preference model based on indirect pairwise comparisons among the alternatives.

However, one of the primary challenges in deploying ANN or DL in MCDA problems lies in ensuring the interpretability of the models. The DM often requires a deeper understanding of the decision process and the relationships between criteria and recommended decisions. To address this concern, \citet{guo2021hybrid} proposed a hybrid DL-based model. It seeks to enhance predictive performance while also delivering interpretable results. More recent research systematically integrated DL into six classic MCDA models, introducing a \textit{monotonic block} to ensure that criteria assumptions are satisfied \citep{martyn2023deep}. The authors aimed to bridge the gap between DL and MCDA by incorporating novel network structures that cater to the interpretability requirements of MCDA.

In summary, deep preference learning in MCDA has primarily focused on enhancing interpretability in decision models \citep{guo2021hybrid} or adapting neural networks to meet specific decision-making requirements \citep{martyn2023deep}. However, these efforts have not extended to decision contexts involving criteria with temporal information. Addressing this temporal dimension in decision-making remains an area for development.

\section{The Proposed Temporal Preference Learning Model} 
\label{sec-pre}

\subsection{Problem Description}
\noindent Let us the following notation to describe MCS problems in the presence of temporal criteria:
\begin{itemize}
    \item $A^R=\{a,b,\ldots \}$ -- a set of reference alternatives. These alternatives can be regarded as training samples, and their classifications are known.
    \item $A=\{a_1,a_2,\ldots\}$ -- a set of non-reference alternatives to be classified.
    \item $Cl=\{Cl_1,\ldots, Cl_H\}$ -- a set of pre-defined decision classes exhibiting a natural ordering, $Cl_h \succ Cl_{h-1}, h = 2,\dots,H$, indicating that the alternatives in class $Cl_h$ are preferred to those in class $Cl_{h-1}$. In the threshold-based MCDA methods, class $Cl_{h}$ is bounded by real-valued thresholds $\theta_{h}$ and $\theta_{h-1}$.
    \item $G=\{g_1,\ldots, g_j,\ldots, g_m\}$ -- a family of $m$ temporal criteria, and $\mathbf{g}_j(a)$ is a time series containing $T_j$ timestamps, i.e., $\mathbf{g}_j(a)=\left(g^1_j(a),\ldots,g^t_j(a),\ldots,g^{T_j}_j(a) \right) \in \mathbb{R}^{T_j}, a\in A\cup A^R$, where $g^t_j(a)$ is the performance value of alternative $a$ on criterion $g_j$ at $t$-th timestamp.
\end{itemize}

\noindent We aim to learn a preference model from the training samples $a\in A^R$ given their assignments $Cl(a)\in Cl$ and then deduce the classification of non-reference alternatives $a'\in A$. A widely used preference model is an additive value function $U(\cdot)$ aggregating the performances of each alternative $a\in A\cup A^R$ on all criteria into a comprehensive score $U(a)$:
\begin{equation}
 U(a)=\sum_{j=1}^m u_j^{T_j}(\mathbf{g}_j(a)), \label{eq-globalu}
\end{equation}
where $u_j^{T_j}(\mathbf{g}_j(a)) $ is the marginal value on criterion $g_j$ at the last timestamp, i.e., the ${T_j}$-th timestamp of the $j$-th time series. In the value-based MCDA methods, the criteria are assumed to be preferentially independent and monotonic~\citep{GRECO20101455}. We define the monotonicity of temporal criteria as follows:
\begin{definition}
A temporal criterion $g_j$ is monotonic if $u_j^{T_j}(\mathbf{g}_j(a))> u_j^{T_j}(\mathbf{g}_j(b))$, given $g_j^{t'}(a)> g_j^{t'}(b), \forall t'\in T^c $ at some timestamps, and $g_j^{t}(a)= g_j^{t}(b), \forall t\in T $ at the other timestamps, where $T\cap T^c = \varnothing, T\cup T^c = \{1,\ldots,T_j\}$.
\label{def-monoto}
\end{definition}

\begin{figure}[h]
    \centering
    \includegraphics[scale=0.45]{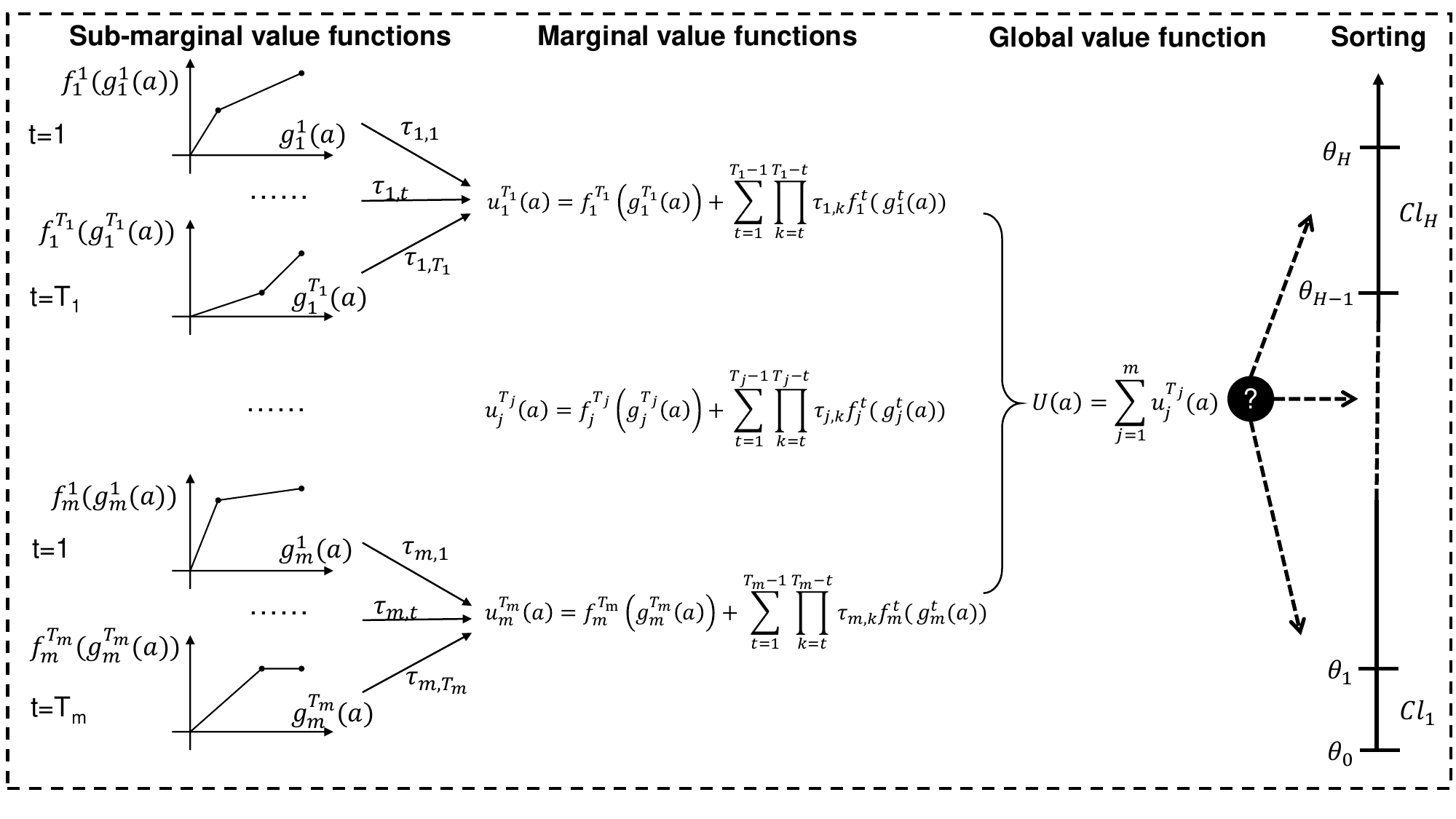}
    \caption{Threshold-based sorting procedure with sub-marginal value functions. The sub-marginal value functions $f_j^t(\cdot)$ of timestamps aggregate the marginal value functions $u_j^{T_j}(\cdot)$ considering time discount factors. Note that the time discount factor $\tau_{j,t}$ can be fixed to $\tau$ (Section \ref{sec-pre}) or learnable across criteria and timestamps (Section \ref{sec-dl}), and the length of the time series can be fixed or not. The comprehensive value of an alternative is determined by $m$ marginal values, and the assignment is decided by comparing the comprehensive value to the delimiting class thresholds.}     
    \label{fig-thresholdsorting}
\end{figure}

\noindent Definition \ref{def-monoto} establishes that in the context of two time series featuring the $j$-th criterion, diverging only at a singular performance value occurring at the $t$-th timestamp, a preference is accorded to the marginal value associated with the superior performance. As stated in Figure \ref{fig-thresholdsorting}, the marginal value is defined as the last timestamp's value, which is related to each timestamp's performance. Then, the comprehensive value aggregates the marginal values associated with the last timestamps' for all criteria. Therefore, alternative $a$ is at least as good as alternative $b$ on a given criterion if only one timestamp's value of $a$ is greater or equal to that for $b$, with all values at other timestamps equal. This preference aligns with the monotonicity principles delineated in static criteria, as expounded upon in~\cite{GRECO20101455}.

We approximate marginal value function $u_j^{T_j}(\cdot)$ using piecewise linear one, aggregating a set of functions $u_j^t(\cdot)$ corresponding to each timestamp $t$. Let $\alpha_j^t=\min_{a\in \{A\cup A^R\}} g_j^t(a)$ and $\beta_j^t=\max_{a\in \{ A\cup A^R\}} g_j^t(a)$ be the worst and best performance on criterion $g_j$ at timestamp $t$. We then divide $\left[\alpha_j^t, \beta_j^t \right]$ into $\gamma_j^t\ge 1$ sub-intervals, i.e., $\left[x_j^{t,0}, x_j^{t,1} \right]$, $\left[x_j^{t,1}, x_j^{t,2} \right], \ldots,$ $\left[x_j^{t,\gamma_j^t-1}, x_j^{t,\gamma_j^t} \right]$, where $x_j^{t,k} = \alpha_j^t + \frac{k}{\gamma_j^t}(\beta_j^t-\alpha_j^t), k=0,1,\ldots,\gamma_j^t$. We define a sub-marginal value $f_j^t(a)$ of alternative $a$ on criterion $g_j$ at $t$-th timestamp:
\begin{equation}
    f_j^t(g_j^t(a)) = f_j^t(x_j^{t,k_j}) + \frac{g^t_j(a)-x^{t, k_j}_j}{x^{t, k_j+1}_j-x^{t, k_j}_j}\left( f_j^t(x_j^{t,k_j+1})-f_j^t(x_j^{t,k_j})\right),\quad \mbox{if}\quad g_j^t(a)\in \left[x_j^{t,k_j},x_j^{t,k_j+1} \right]. \label{eq-f1}
\end{equation}
Note that $u_j^t(\cdot)$ is related to $f_j^t(\cdot)$ at $t$-th timestamp and the values before $t$, but the correlation should be decayed along with the time. We assume there is a time discount factor $\tau_{j,t}$ for $j$-th criterion at $t$-th timestamp, which is set to be fixed in this section, i.e., $\tau=\tau_{1,1}=\tau_{1,2}=\ldots=\tau_{m,T_m}$. The $t$-th marginal value is only associated with the $(t-1)$-th timestamp:
\begin{align*}
    u_j^1(g_j^{1}(a)) &= f_j^{1}(g_j^{1}(a)), \\
    &\cdots \\
    u_j^{t}(g_j^{t}(a)) &= f_j^{t}(g_j^{t}(a)) + \tau\cdot u_j^{t-1}(g_j^{t-1}(a))
     = f_j^{t}(g_j^{t}(a)) + \sum_{k=1}^{t-1} \tau^{t-k}f_j^k(g_j^{k}(a)) ,\quad \forall t\in\{2,\ldots,T_j\},
\end{align*}
where $\tau\in\left[0,1\right]$ indicates how much information should be remembered at the last timestamp. Note that when $\tau=0$, the temporal information is not considered. Then, the comprehensive value of alternative $a$ is:
\begin{equation}
    \begin{aligned}
     U(a)&=\sum_{j=1}^m u_j^{T_j}(\mathbf{g}_j(a)) 
&=\sum_{j=1}^m\left[ f_j^{T_j}(g_j^{T_j}(a)) + \sum_{t=1}^{T_j-1}\tau^{T_j-t} f_j^{t}(g_j^t(a)) \right],
\end{aligned}\label{eq-globalu2}
\end{equation}
where $f_j^t(\cdot)$ is the sub-marginal value function since it does not consider the time discount factor. Denote $\Delta f_j^{t,k}=f_j^t(x_j^{t,k}) - f_j^t(x_j^{t,k-1}), t=1,\ldots,T_j, k=1,\ldots,\gamma_j^t$, then Eq.~(\ref{eq-f1}) can be rewritten as:
\begin{equation}
    f_j^{t}(g_j^{t}(a))=\sum_{k=1}^{k_j} \Delta f_j^{t,k} + \frac{g^t_j(a)-x^{t, k_j}_j}{x^{t, k_j+1}_j-x^{t, k_j}_j} \Delta f_j^{t,k_j+1} \quad if \quad g_j^t(a)\in \left[x_j^{t,k_j},x_j^{t,k_j+1} \right].
\label{eq-9}
\end{equation}
Let $\mathbf{f}_j^t = (\Delta f_j^{t,1},\ldots,\Delta f_j^{t,\gamma_j^t})'$ and $\mathbf{v}_j^t(a)=(v_j^{t,1}(a),\ldots,v_j^{t,\gamma_j^t}(a))'$ be the sub-marginal value and performance value vectors for criterion $g_j$ at $t$-th timestamp, respectively, where 
\begin{equation}
    v_j^{t,k}(a)={ \left\{ \begin{array}{*{20}{l}}{1,\quad if \quad g_j^t(a) > x_j^{t,k},}\\{\frac{g_j^{t}(a)-x_j^{t,k-1}}{x_j^{t,k}-x_j^{t,k-1}},\quad if \quad  x_j^{t,k-1} \le g_j^t(a)\le x_j^{t,k},}\\{0,\quad if \quad g_j^t(a)< x_j^{t,k-1}.}\end{array}\right. }
\end{equation}
Hence, Eq.~(\ref{eq-globalu2}) be expressed as:
\begin{equation}
    \begin{aligned}
        U(a) &= \mathbf{u}' \mathbf{v}(a)  
        =\sum_{j=1}^m \mathbf{u}_j' \Lambda_j \mathbf{v}_j(a),
\end{aligned}\label{eq-globaluv}
\end{equation}
where $\mathbf{u}=(\mathbf{u}_1',\ldots, \mathbf{u}_m')'$ and $\mathbf{v}(a) = (\mathbf{v}_1(a)' \Lambda_1,\ldots,\mathbf{v}_m(a)'\Lambda_m)'$. In particular, $\Lambda_j = { \left[ {{\begin{array}{*{20}{c}}{{ \left( {{\tau }} \right) }\mathop{{}}\nolimits^{{T_j-1}}}&{ \cdots }&{0}\\{ \vdots }&{ \ddots }&{ \vdots }\\{0}&{ \cdots }&{ \left( \tau  \left) \mathop{{}}\nolimits^{{0}}\right. \right. }\end{array}}} \right] }$, $\mathbf{u}_j = ((\mathbf{f}_j^1)',\ldots,(\mathbf{f}_j^{T_j})')'$, and $\mathbf{v}_j(a) = (\mathbf{v}_j^1(a)',\ldots,\mathbf{v}_j^{T_j}(a)')'$. Thus computed comprehensive value $U(a)$ is incorporated into a threshold-based sorting procedure, where the alternative's assignment is decided based on the comparison of its score with the delimiting class thresholds (see Figure \ref{fig-thresholdsorting}).

\subsection{Temporal Preference Learning Model}
\label{subsec-cqo}

\noindent The class assignments for reference alternatives can be translated into pairwise comparisons so that $a$ is preferred to $b$ ($a \succ b$) if the desired class of $a$ is more preferred than the desired class of $b$. Suppose there are total $N$ pairs of alternatives in training set $\mathcal{D}$. Denote each pair as $\{(a,b), y_i\}_{i=1}^N $, where $y_i=1$ if $a \succ b$ and $y_i=-1$ if $b \succ a$. 
For $i$-th pair $(a,b)\in A^R \times A^R$ such that $y_i = 1$, we require that the desired relation is reproduced as follows:
\begin{equation}
    U(a) > U(b) \Leftrightarrow  \mathbf{u}' \mathbf{v}(a) >  \mathbf{u}' \mathbf{v}(b)  \Leftrightarrow \mathbf{u}'\mathbf{v}_i \ge -\xi_i, \label{eq-cons1}
\end{equation}
where $\xi_i\ge 0$, $\mathbf{v}_i = \mathbf{v}(a)-\mathbf{v}(b)$, and putting the minimization of $\sum_i^N \xi_i$ to the objective function leads to a model with the minimum inconsistency. Since we need to approximate a sub-marginal value function $f_j^t(\cdot)$ for each timestamp, there are total $\sum\nolimits_{j=1}^m \gamma_m$ functions. To avoid over-fitting when $\gamma_j$ increases~\citep{liu2019preference}, we propose a~regularized variant of the Temporal Preference Learning (TPL) model, aiming to minimize the total inconsistency:
\begin{align}
    (P0) \quad & \min \frac{1}{2} \left\Vert {\mathbf{u}} \right\Vert^2_2 + C \cdot \sum\nolimits_{i=1}^N \xi_i \nonumber \\
    s.t.:\quad & y_i\mathbf{u}'\mathbf{v}_i \ge 1- \xi_i,\\
    & \xi_i  \ge 0, i=1,\ldots,N, \\
    & \Delta f_j^{t,k} \ge 0, j=1,\ldots,m, t=1,\ldots, T_j, k = 1,\ldots,\gamma_j^{t} .
\end{align}


\noindent There are $N+\sum\nolimits_{j=1}^m\sum\nolimits_{t=1}^{T_j} \gamma_j^t$ constraints in the optimization problem $P0$. Such a high number can prevent obtaining a~solution effectively when there are too many pairwise alternatives, i.e., $N \gg \sum\nolimits_{j=1}^m\sum\nolimits_{t=1}^{T_j} \gamma_j^t$. To this end, we transform $P0$ to the following optimization problem:
\begin{align}
    {(P1)} \quad & \min \frac{1}{2} \left\Vert {\mathbf{u}} \right\Vert^2_2 - \sum\nolimits_{i=1}^N \mu_i \nonumber \\
    s.t.:\quad & \Delta f_j^{t,k} - \sum\nolimits_i^N y_i\mu_i V_{i,j}^{t,k} \ge 0, j=1,\ldots,m, t=1,\ldots, T_j, k = 1,\ldots,\gamma_j^{t}, \\
    & C\ge \mu_i  \ge 0, i=1,\ldots,N.
\end{align}
The proof of the transformation is provided in \ref{app-proof}. Denote the optimal solution of problem $P1$ by $\mathbf{u}^*$. In the threshold-based value-driven methods, a non-reference alternative $a$ is assigned to class $Cl_h$ if $\theta_{h-1}\le U(a) < \theta_h, h=1,\dots, H$. Following \cite{herbrich1999support}, we set threshold $\theta_h$ between $Cl_{h+1}$ and $Cl_h$ in the middle between comprehensive values of two alternatives whose distance is the shortest among all pairs accurately assigned to the respective classes:
\begin{equation}
    \theta_h = \frac{{\mathbf{u}^*}'\mathbf{v}(a)+{\mathbf{u}^*}'\mathbf{v}(b)}{2},  \label{eq-threshold}
\end{equation}
where
\begin{equation}
    (a,b) = \mathop{\arg\min}\limits_{{{{{ \left\{ {\text{(}a,b\text{)} \in \mathop{{A}}\nolimits^{{R}} \times \mathop{{A}}\nolimits^{{R}} \left| {\mathbf{u}^*}' \mathbf{v} \left( a \left)  \ge {\mathbf{u}^*}' \mathbf{v} \left( b \left) , \; a \in A\mathop{{}}\nolimits_{{R}}^{{h+1}}, \; b \in A\mathop{{}}\nolimits_{{R}}^{{h}},h=1, \ldots, H-1 \right. \right. \right. \right. \right.} \right\} }}}}} {\mathbf{u}^*}'(\mathbf{v}(a) - \mathbf{v}(b)),
\end{equation} 
where $A\mathop{{}}\nolimits_{{R}}^{{h}} \subset A^R$, $h=1, \ldots, H$ is the subset of reference alternatives assigned to class $C_h$.
The Hessian matrix of the objective function in $P1$ is positive semidefinite, and the constraint functions are also convex. Thus, $P1$ is a~convex quadratic programming problem, which can be solved by standard software packages, such as Cplex\footnote{{https://www.ibm.com/products/ilog-cplex-optimization-studio/cplex-optimizer}}, python CVXOPT package\footnote{{https://cvxopt.org/}}, and R\footnote{{https://cran.r-project.org/web/packages/e1071/index.html}}.

In some decision problems, to interpret and explain the trade-offs between time series., we can normalize the sub-marginal value functions of all timestamps. Since the temporal criteria are assumed to be monotonic and preference-independent, we apply the following transformations:
\begin{equation}
    \begin{aligned}
     f_j^t(x_j^{t,k_j})' &= \frac{f_j^t(x_j^{t,\gamma_j^t}) - f_j^t(x_j^{t,0})}{\sum\nolimits_{j,t} (f_j^t(x_j^{t,\gamma_j^t}) - f_j^t(x_j^{t,0})) } \times \frac{f_j^t(x_j^{t,k_j}) - f_j^t(x_j^{t,0})}{f_j^t(x_j^{t,\gamma_j^t}) - f_j^t(x_j^{t,0})} 
     =\frac{f_j^t(x_j^{t,k_j}) - f_j^t(x_j^{t,0})}{\sum\nolimits_{j,t} (f_j^t(x_j^{t,\gamma_j^t}) - f_j^t(x_j^{t,0})) }.
\end{aligned}\label{eq-normalization}
\end{equation}
Eq.~(\ref{eq-normalization}) ensures the minimal sub-marginal value, i.e., $f_j^t(x_j^{t,0})'$, is zero, and the sum of maximal sub-marginal values, i.e., $\sum\nolimits_{j,t} f_j^t(x_j^{t,\gamma_j^t}) $, is one. Meanwhile, the obtained thresholds also need to be transformed:
\begin{proposition}
    If alternative $a$ is assigned to class $Cl_h$, i.e., $\theta_{h-1}\le U(a) < \theta_h$, then the transformed threshold should satisfy $\theta_h'=\frac{\theta_h - \sum_{j=1}^m{\left[ f_j^{T_j}(x_j^{T_j,0}) + \sum_{t=1}^{T_j-1} \tau^{T_j-t}f_j^t(x_j^{t,0}) \right]}}{\sum\nolimits_{j,t} (f_j^t(x_j^{t,\gamma_j^t}) - f_j^t(x_j^{t,0})) }$ to ensure $\theta_{h-1}'\le U'(a) < \theta_h'$, where $U'(\cdot)$ is the comprehensive value computed based on the transformed sub-marginal value functions. \label{prop-1}
\end{proposition}
The proof of Proposition \ref{prop-1} is available in \ref{app-prop1}. If we normalized the sub-marginal value functions, the optimized thresholds should also be transformed to ensure the classifications are consistent. Note that the thresholds are related to the maximal/minimal marginal values and the time discount factor.

\section{Deep Preference Learning with Monotonic Recurrent Neural Network}
\label{sec-dl}
\noindent The solution of the quadratic optimization-based TPL model introduced in Section~\ref{subsec-cqo} can require high computational effort. It may be too challenging for the existing optimization techniques when the number of assignment examples and timestamps in the temporal criteria scale up and even become unsolvable if the time discount factors are not fixed. Accumulating vast data, it is reasonable to foster the intersection of MCDA and deep learning to help the DM deal with large-scale decision problems~\citep{guo2021hybrid,martyn2023deep}. To this end, we formulate the standard RNNs to ensure the monotonicity in Definition \ref{def-monoto} and propose a novel monotonic recurrent neural network (mRNN) for MCS problems with temporal criteria.

\subsection{Recurrent Neural Network}

\noindent RNN is an artificial neural network designed to process sequential data and time-series information due to the ability to capture temporal dependencies and patterns within sequences effectively. In contrast to traditional feed-forward neural networks, RNN introduces recurrent connections, allowing information to persist across time, which endows them with the capacity to maintain an internal state or memory. This dynamic state enables RNN to contextualize current inputs based on the history of previously encountered elements in the sequence, effectively modeling long-range dependencies.

The computation at each timestamp involves a non-linear activation function ($\tanh(\cdot)$) applied to a combination of the current input and the previous hidden state., i.e.,
\begin{align}
    \mathbf{h}^t(a) = \tanh(\mathbf{W}_g \mathbf{g}^t(a)+ \mathbf{W}_h\mathbf{h}^{t-1}(a)+\mathbf{b}), \label{eqn:update of RNNs}
\end{align}
In Eq.~(\ref{eqn:update of RNNs}), $\mathbf{h}^t(a) \in \mathbb{R}^{c\times 1}$ is the hidden state at $t$-th timestamp with pre-defined dimension $c$, which comprises three parts. The first part comes from the input vector $\mathbf{g}^t(a)=(g_1^t(a), g_2^t(a), \cdots, g_m^t(a))'\in \mathbb{R}^{m\times1}$ containing $m$ temporal criteria at $t$-th timestamp, and $\mathbf{W}_g \in \mathbb{R}^{c\times m}$ is the input-to-hidden weight matrix. The second part contains the information from the last timestamp $\mathbf{h}^{t-1}(a)$, and $\mathbf{W}_h \in \mathbb{R}^{c\times c}$ is the hidden-to-hidden weight matrix that captures the hidden state transitions within the recurrent connections in the RNN, allowing the network to retain information from the past timestamps and model sequential dependencies. The last part $\mathbf{b} \in \mathbb{R}^{c}$ is the bias vector associated with the hidden state of the RNN, enabling the model to introduce shifts in the learned representations and can have a~significant impact on the network's ability to fit the data and generalize to test samples. In this way, the hidden state $\mathbf{h}^t(a)$ retains information about criteria from the current and previous timestamps, including $\mathbf{g}^1(a),\cdots,\mathbf{g}^t(a)$, and carries it forward to influence future predictions.

The parameters in Eq.~(\ref{eqn:update of RNNs}) can be optimized by standard gradient descent algorithms. The hyperbolic tangent activation function, $\tanh(\cdot)$, squashes the input values into the range $(-1, 1)$, ensuring non-linear transformations, which is crucial for the RNN to learn complex patterns and dependencies in sequential data~\citep{lecun2015deep}.

Standard RNN cannot be directly applied to MCS problems due to the following three issues. First, the non-linear activation functions cannot ensure the monotonic relationship as illustrated in Definition~\ref{def-monoto}. It is essential in most MCDA problems since the preferences are usually stable and have an obvious positive/negative impact on the decisions. Second, RNN inherently processes all criteria jointly, assuming full correlation among them, which, however, contradicts the essence of preference independence assumption, where criteria are processed independently to help the DM simplify the decision problem and comprehend it better. Third, the output of standard RNN for classification problems is usually a probability for assignments, indicating no inherent superiority or inferiority between classes. Consequently, the cross-entropy loss function -- the standard choice for classification problems -- is not suitable for capturing the natural ordering in classes for the MCS problems. Moreover, as stated in Eq.~(\ref{eq-threshold}), the optimized thresholds are outside the proposed TPL model, which may raise issues if one sets them in the same range, but better results would be obtained close to either class. Although this determination of thresholds is used in \cite{herbrich1999support}, the proposed mRNN can also handle such a problem.

\subsection{Ensuring Monotonicity and Preference Independence}
\noindent The sub-marginal value vector $\mathbf{f}^t(a)=(f_1^t(a), f_2^t(a), \cdots, f_m^t(a))'\in \mathbb{R}^{m \times 1}$ for all $m$ criteria at the $t$-th timestamp can be derived as a linear mapping from the hidden state $\mathbf{h}^{t}(a)$: 
\begin{align}
    \mathbf{f}^t(a) = \mathbf{W}_f \mathbf{h}^{t} (a),
    \label{eqn:utility function}
\end{align}
where $\mathbf{W}_f \in \mathbb{R}^{m\times c}$ is a learnable weight matrix. 
The derivative of the sub-marginal value functions on all criteria at the $t$-th timestamp should satisfy the following condition: 
\begin{align}
    \frac{\partial \mathbf{f}^t(a)}{\partial \mathbf{g}^t(a)} &= \frac{\partial \mathbf{f}^t(a)}{\partial \mathbf{h}^t(a)}\frac{\partial \mathbf{h}^t(a)}{\partial \mathbf{g}^t(a)}=\mathbf{W}_f \tanh^{'}(\cdot)\mathbf{W}_g \ge 0.
    \label{eq-der}
\end{align}
Given Eq.~(\ref{eq-der}), since the derivative of $\tanh(\cdot)$ is strictly positive for all inputs, one only needs to impose a constraint during the training process to prohibit the training parameters, $\mathbf{W}_f$ and $\mathbf{W}_g$, from negative values. Specifically, we bypass $\mathbf{W}_f$ and $\mathbf{W}_g$ through an activation function $\phi(\cdot)$ that guarantees the non-negativity of the output. This study employs the Rectified Linear Unit (ReLU), which sets all negative input values to zero and leaves positive input values unchanged, i.e., $\phi(x)=\max(x, 0)$. The advantages of adopting ReLU are its simplicity and computational efficiency and its ability to mitigate the vanishing gradient problem encountered during backpropagation in ANN~\citep{glorot2011deep}. In this way, we can guarantee that the sub-marginal value functions are monotonically correlated with the temporal criteria values. 

\begin{figure}
    \centering
    \includegraphics[trim=0 60 0 0,clip, scale=.5]{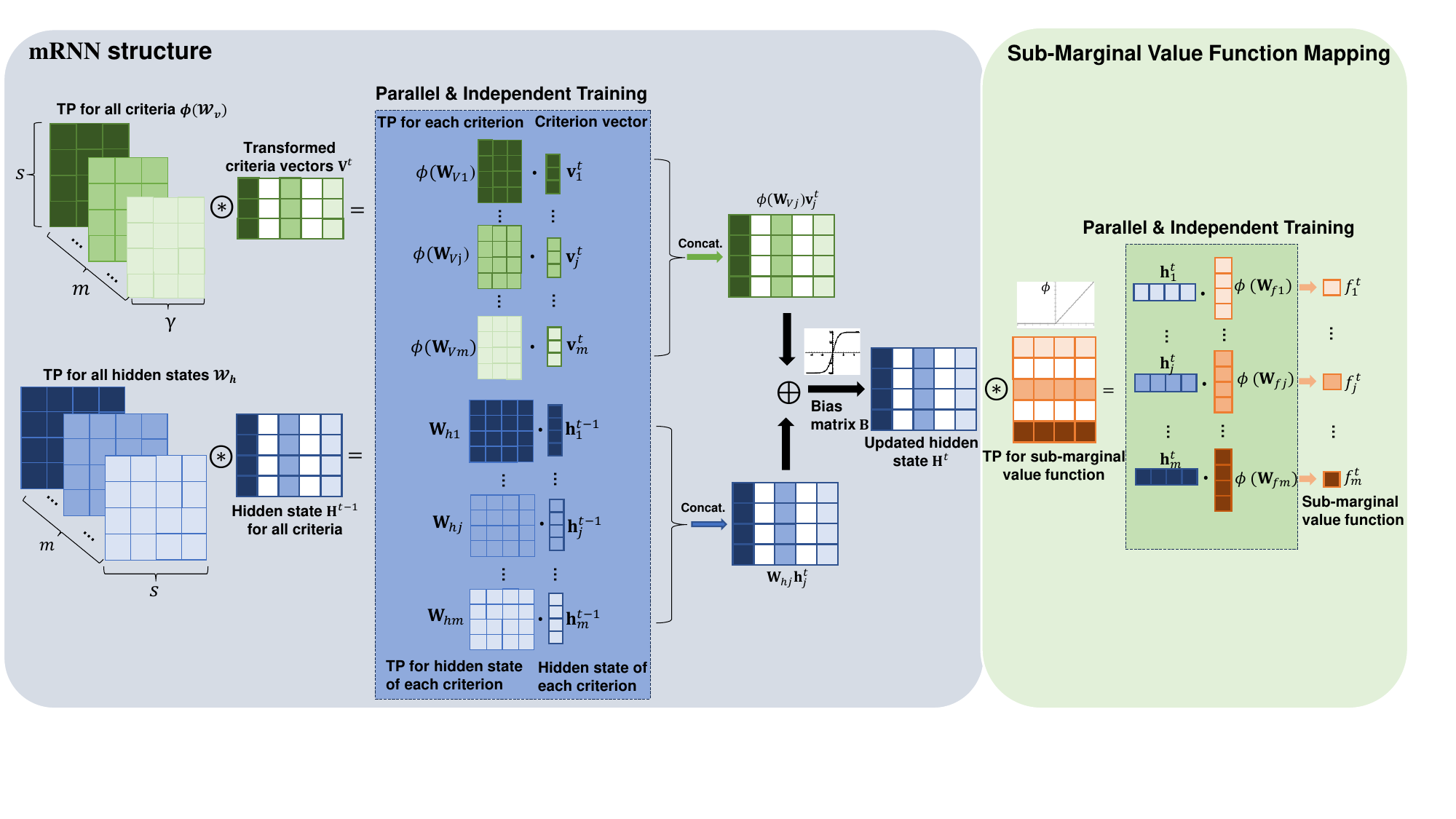}
    \caption{The illustration of mRNN and the training process of sub-marginal value functions. TP is short for training parameters. The notations correspond to Eqs.~(\ref{eq-der})--(\ref{eq-21}). In particular, $\mathcal{W}_v$ and $\mathcal{W}_h$ are the transformed learnable parameter matrices, $\mathbf{V}^t(a)$ represents the input transformed performance values of all criteria at timestamp $t$, and $\mathbf{H}^{t-1}(a)$ is the hidden state of the previous timestamp $t-1$.}
    \label{fig:operation}
\end{figure}

To cater to the need for preference independence, we develop a novel architecture that processes each criterion in a parallel manner. Instead of sharing a common hidden state across all criteria, we allocate a unique one to each criterion whose parameters are trained independently. By adopting this approach, the hidden state of one criterion remains isolated from the others throughout the entire training procedure, ensuring that the obtained sub-marginal value function is not related to others. For simplicity, we assume the numbers of sub-intervals and timestamps are fixed, i.e., $\gamma_j^t=\gamma$ and $T_j=T, j=1,\ldots,m. $ The parameter update process for the proposed mRNN is as follows: 
\begin{align}
    \mathbf{H}^t(a) = \tanh(\phi(\mathcal{W}_v) \circledast \mathbf{V}^t(a)+ \mathcal{W}_h \circledast \mathbf{H}^{t-1}(a)+\mathbf{B}), 
\label{eqn:update of mRNNs}
\end{align}
where $\mathbf{V}^t(a)=[\mathbf{v}^t_1(a), \mathbf{v}^t_2(a), \cdots, \mathbf{v}^t_m(a)] \in \mathbb{R}^{m \times \gamma \times 1}$ are the transformed performance values on all criteria at timestamp $t$. In this way, $\mathbf{H}^{t}(a)=[\mathbf{h}_1^{t}(a), \cdots, \mathbf{h}_j^{t}(a), \cdots, \mathbf{h}_m^{t}(a)] \in \mathbb{R}^{m\times s \times 1}$ considers both the performance values at the current $t$-th timestamp and the historical information of $\mathbf{H}^{t-1}(a)$. The operation $\circledast$ is defined as the tensor-wise multiplication: 
\begin{align}
    \phi(\mathcal{W}_v) \circledast \mathbf{V}^t(a)&=[\phi(\mathbf{W}_{v1})\mathbf{v}_1^t(a),\cdots,\phi(\mathbf{W}_{vj})\mathbf{v}_j^t(a),\cdots,\phi(\mathbf{W}_{vm})\mathbf{v}_m^t(a)] \in \mathbb{R}^{m \times s \times 1}, \\
\mathcal{W}_h \circledast \mathbf{H}^{t-1}(a)&=[\mathbf{W}_{h1}\mathbf{h}_1^{t-1}(a),\cdots,\mathbf{W}_{hj}\mathbf{h}_j^{t-1}(a),\cdots,\mathbf{W}_{hm}\mathbf{h}_m^{t-1}(a)]  \in \mathbb{R}^{m \times s \times 1}.
\label{eq-21}
\end{align}
The parameters in $\phi(\mathcal{W}_v), \mathcal{W}_h$ and $\mathbf{B}$ are learnable. In particular, $\phi(\mathcal{W}_v)=[\phi(\mathbf{W}_{v1}), \cdots, \phi(\mathbf{W}_{vj}), \cdots, \phi(\mathbf{W}_{vm})] \in \mathbb{R}^{m\times s\times \gamma}$ contains $m$ matrices with the size of $s \times \gamma$, where $s$ is the hidden unit's size and $\gamma$ is the pre-defined number of sub-intervals. Each matrix $\mathbf{W}_{vj}\in \mathbb{R}^{s\times \gamma}$ represents the sub-marginal values of the characteristic points in $\mathbf{v}^t_j(a)$ at $t$-th timestamp. Moreover, $\mathcal{W}_h=[\mathbf{W}_{h1}, \cdots, \mathbf{W}_{hj}, \cdots, \mathbf{W}_{hm}]\in \mathbb{R}^{m\times s\times s}$ considers each criterion's historical information of the previous $t-1$ timestamps in a preference independence manner by updating $\mathbf{h}_j^t(a) \in \mathbb{R}^{s\times 1}$ separately. At last, $\mathbf{B}=[\mathbf{b}_{1},\cdots, \mathbf{b}_{j}, \cdots,\mathbf{b}_{m}]\in \mathbb{R}^{m\times s \times 1}$ is the matrix of bias. Figure \ref{fig:operation} presents the training process of the proposed mRNN structure.

\subsection{Learning Sub-marginal and Marginal Value Functions}

\begin{figure}
    \centering
    \includegraphics[trim=0 120 0 100, clip, width=1\columnwidth]{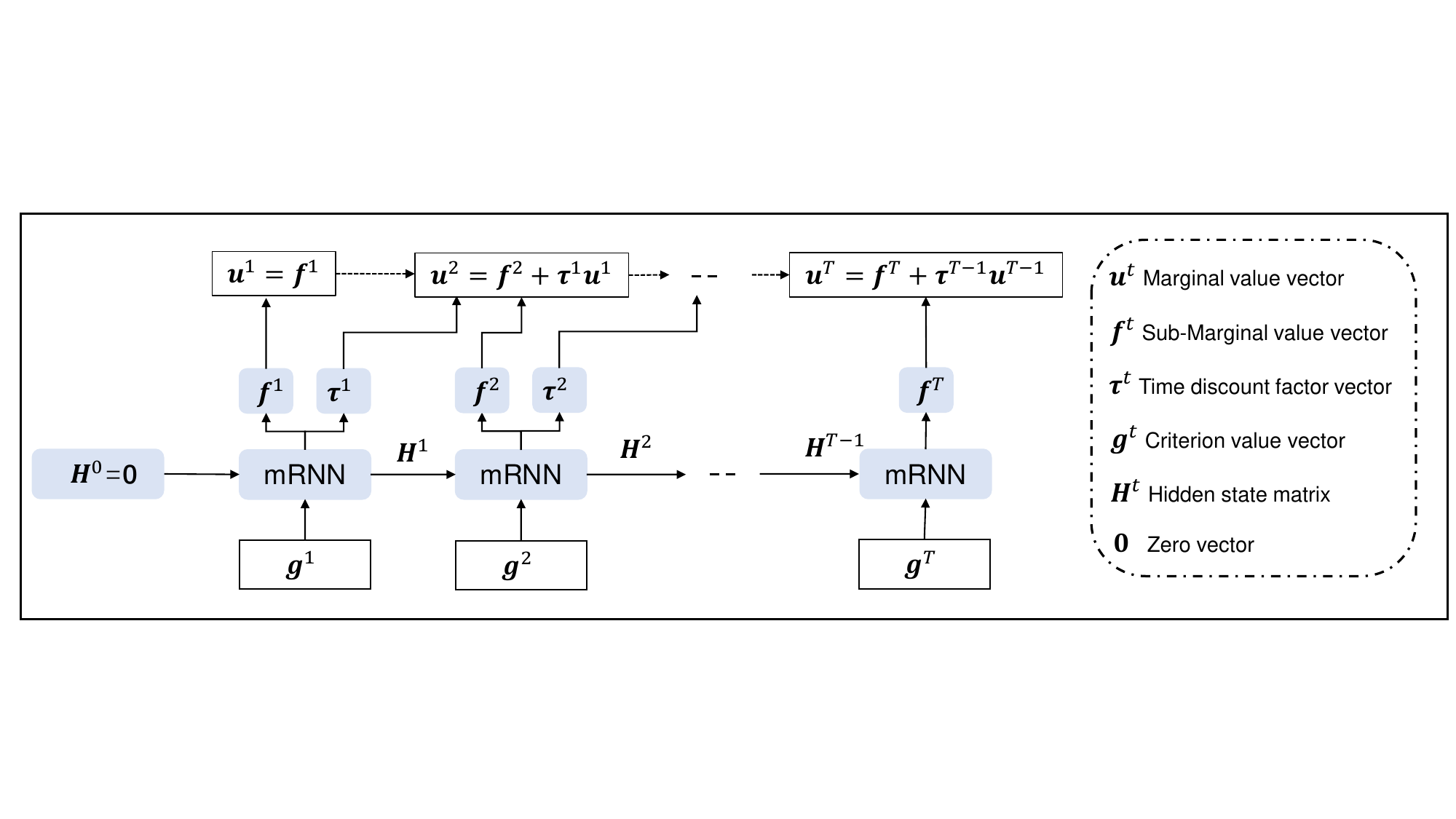}
    \caption{The process of learning comprehensive values aggregating sub-marginal value functions and time discount factors. Note that $\mathbf{u}^t=(u_1^t,\ldots,u_m^t)'\in \mathbb{R}^m, \mathbf{f}^t=(f_1^t,\ldots,f_m^t)' \in \mathbb{R}^m, \boldsymbol{\tau}^t=(\tau_{1,t},\ldots,\tau_{m,t})'\in \mathbb{R}^m,$, and $\mathbf{g}^t=(g_1^t,\ldots,g_m^t)'\in\mathbb{R}^m$ are the vectors at the $t$-th timestamp of all criteria.}
    \label{fig:learning}
\end{figure}

\noindent This section discusses how to depict the DM's preferences by the proposed mRNN structure. In Figure~\ref{fig:learning}, we illustrate the aggregation of the sub-marginal value functions and the time discount factors. Recall that $\mathbf{f}^t(a)=[f^t_1(a), f^t_2(a), \cdots, f^t_m(a)]$ contains the sub-marginal values of all criteria at timestamp $t$:
\begin{align}
    \mathbf{f}^t(a) &= \phi(\mathcal{W}_f^t) \circledast \mathbf{H}^t(a)\\
                    &= [\phi(\mathbf{W}_{f1}^t)\mathbf{h}^t_1(a), \phi(\mathbf{W}_{f2}^t)\mathbf{h}^t_2(a), \cdots, \phi(\mathbf{W}_{fm}^t)\mathbf{h}^t_m(a)], \label{eq:sub-mgn}
\end{align}
where $\phi(\mathcal{W}_f^t)=[\phi(\mathbf{W}_{f1}^t), \phi(\mathbf{W}_{f2}^t), \cdots, \phi(\mathbf{W}_{fm}^t)] \in \mathbb{R}^{m\times s \times 1}$ are learnable parameters and $\phi(\cdot)$ is the RuLU activation function to ensure the monotonicity. We ensure that each criterion's sub-marginal value function and time discount factors are also learned independently. 

We assume that the marginal value function $\mathbf{u}^t(a)=[u_1^t(a), u_2^t(a), \cdots, u_m^t(a)]$ at any given timestamp $t$ is composed of two distinct components. The first one comes from the discounted marginal value functions at timestamp $t-1$, represented by $\boldsymbol{\tau}^{t-1}\odot\mathbf{u}^{t-1}(a)$, where $\boldsymbol{\tau}^{t-1} = (\tau_{1,t-1},\ldots,$ $\tau_{j,t-1}, \ldots$, $\tau_{m,t-1})'$ is a column vector of size $m$ for $(t-1)$-th timestamp, and $\odot$ is the element-wise product. The other component is the sub-marginal value function $\mathbf{f}^t(a)$ at the current timestamp. Instead of using a fixed pre-defined time discount factor, mRNN allows $\boldsymbol{\tau}^{t-1}$ to be learnable:
\begin{align}
    \boldsymbol{\tau}^{t-1}=Q(\mathbf{H}^{t-1}(a)), 
\end{align}
where $Q(\cdot)$ is a feed-forward network with a smaller size. In this way, our model can adaptively and dynamically adjust the impact of the sub-marginal values in the past, and the time discount factor is related to the input of criteria values. The marginal value function $\mathbf{u}^t(a)$ at timestamp $t$ can be obtained by:
\begin{align}
    \mathbf{u}^t(a)=\mathbf{f}^t(a)+\boldsymbol{\tau}^{t-1} \odot \mathbf{u}^{t-1}(a). 
    \label{eq:discount}
\end{align}
The comprehensive value function $U(a)$ can also be derived from the marginal value functions at the last timestamp~$T$:
\begin{align}
    U(a) &= \sum_{j=1}^m u_j^T(g_j^T(a)) \nonumber\\
        &=\sum_{j=1}^m f_j^T(g_j^T(a))+\tau_{j,T-1} u_j^{T-1}(g_j^{T-1}(a)) \\
        &=\sum_{j=1}^m f_j^T(g_j^T(a))+\tau_{j,T-1}f_j^{T-1}(g_j^{T-1}(a))+  \cdots+\tau_{j,T-1}\cdots\tau_{j,1}f_j^{1}(g_j^{1}(a)). \nonumber
\end{align}

\subsection{Loss Function with Ordinal Thresholds}

\noindent Unlike in MCDA, the classification task in machine learning typically involves categories without inherent superiority or inferiority between them. Consequently, applying the cross-entropy loss function cannot capture the inherent orders of classes in MCDA problems, thus making it inefficient for decision-making. 

An alternative solution is setting a decision boundary manually. For instance, such a boundary is often set to $0.5$ in a binary classification scenario, as the predicted probabilities typically range from 0 to 1. However, this configuration may not suit multiple classes, resulting in a sub-optimal outcome. Real-world decision-making problems require adaptive decision boundaries that can better accommodate the inherent complexities and variations in the data distribution.

To address this limitation, we present a novel loss function with ordinal thresholds, allowing the model to learn the classification threshold directly from the data and enabling the proposed mRNN structure to act like the threshold-based MCDA method. Note that the assignment of alternative $a$ to class $\hat{Cl}(a)$ can be considered by comparing its comprehensive value $U(a)$ to the thresholds $\theta_0, \theta_1, \cdots,\theta_{H}$:
\begin{equation}
\hat{Cl}(a)= \begin{cases}Cl_1, & \text { if } \theta_0 < U(a) \leq \theta_1 \\ Cl_2, & \text { if } \theta_1<U(a) \leq \theta_2 \\ \vdots & \\ Cl_H, & \text { if } \theta_{H-1}<U(a)\le \theta_H\end{cases},
\end{equation}
where $\theta_0$ and $\theta_H$ can be set as infinitely small and large numbers. Thus, we can derive the distribution of $\hat{Cl}(a)$ as:
\begin{equation}
\begin{aligned}
P(\hat{Cl}(a)=Cl_h \mid \mathbf{v}(a))  =P\left(\theta_{h-1}<U(a) \leq \theta_k\right)  =\Phi\left(\theta_h-U(a)\right)-\Phi\left(\theta_{h-1}-U(a)\right), \label{eq:threshold}
\end{aligned}
\end{equation}
where $\Phi(\cdot)$ is the cumulative distribution function of the standard normal distribution that can be approximated by the Sigmoid function~\citep{bowling2009logistic}. A loss function can then be constructed to minimize the negative log-likelihood of the predicted results compared to the true labels. The loss function for the proposed mRNN structure~is:
\begin{equation}
    \begin{aligned}
    \mathcal{L} &= \sum\nolimits_{a\in A^R} \mathcal{L}(a) = - \sum\nolimits_{a\in A^R} \sum\nolimits_{h=1}^H \mathbf{1}_{Cl(a)=Cl_h} \cdot \log[\Phi\left(\theta_h-U(a)\right)-\Phi\left(\theta_{h-1}-U(a)\right)],
\end{aligned} \label{eq:loss}
\end{equation}
where $Cl(a)$ is the true class of alternative $a \in A^R$, and the thresholds are learned by the neural networks. In \ref{app-dummyexam}, we present a numerical example illustrating how to utilize the ordinal thresholds in Eq.~(\ref{eq:loss}). Still, the recommended class assignments are established by comparing the values of learned $U(\cdot)$ with class thresholds rather than suggesting a class with the greatest probability defined above.


\section{Experiments on Real-world Data Concerning Gaming Industry}
\label{sec-exp}

\noindent In this section, we showcase the effectiveness of the proposed quadratic optimization models and the mRNN in addressing MCS problems with temporal criteria through a real case study. The objective is to distinguish between high- and low-value users for a Multiplayer Online Battle Arena (MOBA) mobile game app. Our experiments aim to provide insights into the following key questions:
\begin{itemize}
    \item To what extent does modeling temporal relationships contribute to improved performance in solving MCS problems with time series criteria?
    \item To what degree are the proposed optimization and deep learning-based models superior to conventional machine learning and MCDA methods?
    \item How does the performance of the proposed models change if we neglect the initial periods and consider only later timestamps?
    \item Can we extract meaningful interpretations from the predictions and user preferences, especially concerning the time discount factor and marginal value functions?
\end{itemize}

\subsection{Data Description}
\label{subsec-data}

\noindent The dataset utilized in this study is sourced from one of China's most prominent global Internet companies. It~comprises four distinct time series spanning from April 30, 2023, to May 30, 2023. These time series pertain to various aspects of user behavior within a MOBA mobile game app. Specifically, the four-time series encompass the following dimensions: (a) \textbf{Purchase Amounts (CNY)}: This sequence tracks the total payment made by each user at each timestamp, denominated in Chinese Yuan (CNY). It provides insights into users' spending behavior within the game. (b) \textbf{Purchase Frequency}: This sequence records how frequently a user makes purchases within the game at each timestamp. It quantifies the user's transactional frequency. (c) \textbf{Time Spent (hours)}: This sequence captures the total time each user actively engages in the game. The time is measured in hours and reflects users' engagement levels. (d) \textbf{Log-in Frequency}: This sequence documents the frequency with which a user logs into the game. It indicates the user's interaction frequency with the game.

\begin{figure}[ht]
    \centering
    \includegraphics[width=\columnwidth]{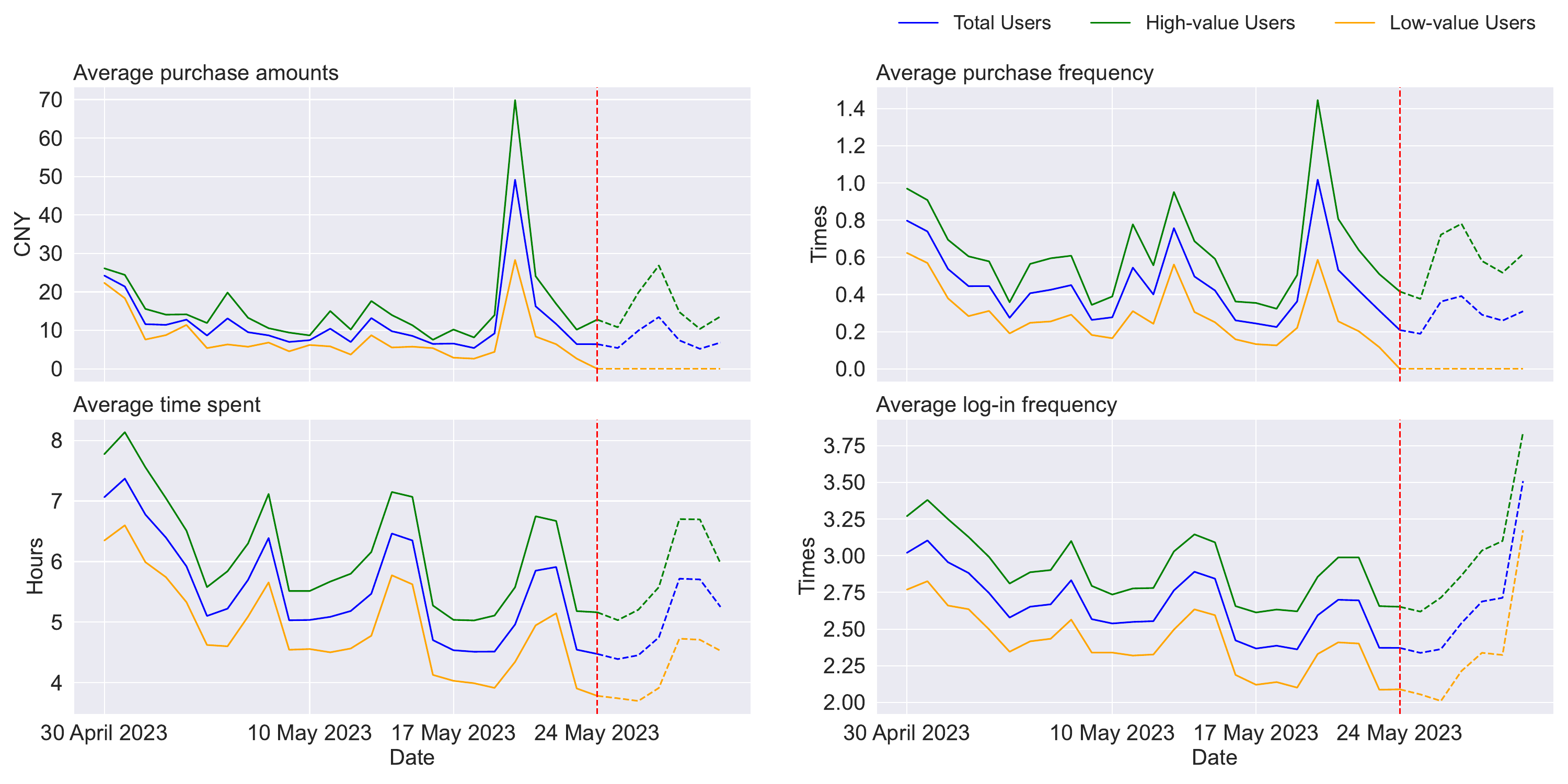}
    \caption{Visualization of averaged criteria values concerning each timestamp of different types of users. The solid line represents the training period, while the dotted line represents the testing data.}
    \label{fig-visualization}
\end{figure}

The dataset encompasses a total of 3,080 users, each of whom is categorized as either ``high-value'' or ``low-value''. This classification is determined based on the sum of each user's in-game purchases over seven days, spanning from May 24, 2023, to May 30, 2023, which is a commonly used time window in user lifetime value evaluation \citep{zhao2023percltv}. Users with a cumulative purchase amount greater than zero during this period are designated as high-value, constituting 47\% of the total user population. The remaining users are classified as low-value.

\begin{table}[tbh]
    \centering\footnotesize
        \caption{The descriptive statistics of different types of users.}
        \label{tab-desp}
        \begin{tabular}{cccccc}
    \toprule
        Type of Users & Criteria & Mean & Standard Deviation & Min & Max   \\ \midrule
        All Users & Purchase amounts & 12.82 & 76.66 & 0 & 3,686  \\ 
        ~ & Purchase frequency & 0.47 & 1.32 & 0 & 38  \\ 
        ~ & Time spent & 5.50 & 5.50 & 0  & 72.59  \\ 
        ~ & Log-in frequency & 2.67 & 2.15 & 0 & 22  \\  \midrule
        High-Value Users & Purchase amounts & 20.00 & 99.87 & 0 & 3,686  \\ 
        ~ & Purchase frequency & 0.70 & 1.62 & 0 &  38  \\ 
        ~ & Time spent &7.35 & 6.15 & 0 &  72.56  \\ 
        ~ & Log-in frequency & 3.45 & 2.37 & 0 & 22  \\ \midrule
        Low-Value Users & Purchase amounts & 5.25 & 37.93 & 0 & 1,958  \\ 
        ~ & Purchase frequency & 0.22 & 0.84 & 0 & 27  \\ 
        ~ & Time spent & 3.55 & 3.85 & 0 & 36.49  \\ 
        ~ & Log-in frequency & 1.86 & 1.51 & 0 & 14  \\ \bottomrule
    \end{tabular}
\end{table}

The descriptive statistics across all timestamps are provided in Table \ref{tab-desp}. We have filtered out the users who never logged in, i.e., those with criteria values at any timestamp being zero. The visual representations in Figure \ref{fig-visualization} underscore significant differences between high-value and low-value users in terms of their historical in-game purchases and time spent within the app. Specifically, high-value users exhibit statistically higher engagement levels in these aspects than their low-value counterparts. These insights are the foundation for the subsequent benchmark models and their respective implementation details, which will be elaborated upon in the relevant subsections.

It is imperative to underscore that the data depicted in Figure \ref{fig-visualization} pertains to average values derived from distinct user groups instead of individualized patterns. Consequently, certain criteria may exhibit apparent correlations. However, it is essential to disentangle the misconception that increased user payment frequency inevitably corresponds to higher aggregate expenditure. This is exemplified in Figure \ref{fig:discountfator}, which delineates the nuanced fluctuations in individual criteria values, demonstrating their non-interdependence. Notably, in mobile gaming, expenditure patterns manifest divergently, with some users making substantial one-time payments and others opting for recurrent, albeit smaller, transactions. Consequently, the statistical averages of purchase amounts and frequency are interlinked, not due to the intrinsic nature of individual behaviors but rather due to the heterogeneous spending practices across user cohorts. It is pivotal to emphasize that the purview of this study centers on individual behaviors, as their impacts can exhibit considerable heterogeneity. Thus, the criteria under consideration are far from redundant within this analytical framework.

The $F-score$, defined as the harmonic mean of a~system's precision and recall values, is used as the primary performance metric in evaluation:
\begin{equation}
    F-score_i = \frac{(1+\beta ^ 2) \cdot recall_i \cdot precision_i}{\beta^2\cdot recall_i + precision_i} \label{eq-f}
\end{equation}
where $recall_i = \frac{m_{ii}}{\sum\nolimits_{j=1}^H m_{ij}}$, $precision_i=\frac{m_{ii}}{\sum\nolimits_{j=1}^H m_{ji}}$, and $m_{ij}, i,j=1,\ldots,H$ is the number of alternatives whose real class is $Cl_{i}$ with prediction being $Cl_j$. The parameter $\beta$ controls the trade-off between recall and precision ($\beta=1$ in most cases). The final $F-score = \frac{\sum\nolimits_{i=1}^H F-score_i}{H}$ is computed as the average $F-score$ across all classes, providing a comprehensive evaluation metric for the models' performance. Apart from the $F$ measure, precision, and recall, we also report $accuracy=\frac{\sum\nolimits_{i=1}^H m_{ii}}{|A|}$ as the most intuitive measure reflecting the performance of predictive models.

All studies and experiments were run on Dell Precision 7920 Workstations with Intel(R) Xeon(R) Gold 6256 CPU at 3.60GHz, and NVIDIA Quadro GV100 GPUs. All models were implemented in Python 3.8. The versions of the main packages of our code are Pytorch 1.8.1+cu102, Scikit-Learn: 0.23.2, Numpy: 1.19.2, Pandas: 1.1.3, Matplotlib: 3.3.2. In addition, Pulp\footnote{\url{https://coin-or.github.io/pulp/}} and Mosek\footnote{\url{https://www.mosek.com/}} libraries were used to implement the optimization-based models.


\subsection{Comparison with Baseline Models}

\noindent We will first verify if conventional machine learning models and standard MCDA methods can successfully discriminate between high- and low-value without considering the time discount factor. This will let us understand the roles played by modeling temporal relationships. In this regard, we will compare the following methods:
\begin{itemize}
    \item Temporal preference learning (TPL) model: This model corresponds to the optimization-based approach introduced in problem statement $P1$. It optimizes the model parameters using all available data simultaneously. The optimization problem is solved using the CVXOPT Python package with the Mosek optimizer.
    
    
    \item The UTilit\'es Additives DIScriminantes (UTADIS) method \citep{zopounidis2002multicriteria}: It is a~classic approach for MCS problems. It~utilizes linear programming to minimize misclassification errors, defined as the average distance of comprehensive values and thresholds corresponding to the desired classes. In this study, each timestamp is treated as an individual criterion in UTADIS, and a threshold-based preference model in the form of piecewise linear functions with equal segments is inferred using the Pulp Python package.

    \item UTADIS-Tensorial (UTADIS-T): It is adapted for MCS problems based on \cite{campello2023improving}. Specifically, we manually derive two descriptive statistics, including mean and slope coefficients, as illustrated in \cite{campello2023improving}. The first indicator can measure the average criteria value in the time series, while the second indicator can describe the volatility of each time series.
    
    \item Logistic regression (LR) \citep{hosmer2013applied}: It is a linear model aggregating input attributes (criteria in this study) using weighted sums. It transforms the output using a~function determining the probability of assigning an alternative to a~desired class. The hyper-parameter $C$ controls the degree of regularization applied to the model.
    
    \item Support vector machines (SVM) \citep{cortes1995support}: It is a supervised learning algorithm that maps training examples to points in space to maximize the margin between two classes. This study uses the criteria values for all timestamps as inputs to SVM. A Radial Basis Function kernel is employed to learn a non-linear classifier. The hyper-parameter $C$ controls the trade-off between maximizing the margin and minimizing classification errors on the training data.

    \item Random forests (RF) \citep{ho1995random}: It is an ensemble learning method that aggregates multiple decision trees during training. It is often used as a black-box model because it can provide predictions across a wide range of data with minimal configuration. Hyper-parameters such as n\_estimators, max\_features, and max\_depth play critical roles in controlling the behavior and performance of the ensemble of decision trees in RF.
        
    \item Extreme gradient boosting (XGB) \citep{chen2016xgboost}: It is a sequential model based on a gradient-boosting algorithm. It~differs from bagging algorithms and can be parallelized. XGB incorporates regularization techniques to generate smaller trees, mitigating overfitting issues. Hyper-parameters like n\_estimators and max\_depth are crucial for controlling the behavior and performance of the ensemble of decision trees in XGB.

    \item Explainable Ordinal Factorization Model (XOFM) \citep{guo2021deciphering}: It is derived from the factorization machine and also uses piecewise linear functions to decipher the criteria impacts on the ordinal regression problems. Moreover, XOFM can account for low-order interacting criteria, which preserves comparable performance.
\end{itemize}
Note that we treat the classic UTADIS as a benchmark MCDA model, serving as a reference point for evaluating the performance of the proposed approaches. UTADIS-T is its recently proposed counterpart considering the temporal information in MCS. In turn, LR, SVM, RF, and XGB are widely used machine learning models for classification tasks where classes may not exhibit a natural order. They are trained to predict the probability of high-value using cross-entropy loss. Finally, XOFM is designed for ordinal classification problems. 

We will also check if the proposed model performs better in user value evaluation problems than standard deep learning methods. The following methods will be compared in the experiments:
\begin{itemize}
    \item Monotonic recurrent neural network (mRNN): It is a novel deep preference learning model designed for temporal criteria with learnable time discount factors. Several adaptations are made to ensure the criteria monotonicity, preference independence, and natural order between classes assumptions are satisfied. This model captures the temporal dynamics in the criteria while maintaining these critical properties.
    \item Multi-layer perceptron (MLP) \citep{rosenblatt1958perceptron}: It is a fully connected ANN. It can capture non-linear transformations and interactions between criteria, making it suitable for distinguishing data that is not linearly separable. This study employs a three-layered MLP as a deep learning-based baseline, with criteria values from all timestamps directly used as input.
    \item Recurrent neural network (RNN) \citep{schuster1997bidirectional}: It is a type of sequential ANN that considers internal states to process sequential data. The output from the current state can influence subsequent inputs. The proposed mRNN is adapted from the standard RNN architecture to better describe and address MCS problems with temporal criteria.
    \item Gated recurrent unit (GRU) \citep{cho2014learning}: It is a RNN variant incorporating a gating mechanism. It features a forget gate, allowing only some information from previous states to pass to the subsequent states. GRU is designed to capture long-range dependencies in sequential data efficiently.
\end{itemize}
The DL-based models, such as MLP, RNN, and GRU, are known for their efficiency in handling large-scale and high-dimensional data \citep{lecun2015deep}. The hyper-parameters and DL model structures are tuned using the ten-fold cross-validation. The best parameters are presented in \ref{app-parameter}. 

Table~\ref{tab-baselineresults} presents the values of all four measures averaged across ten-fold cross-validation. First, despite not explicitly learning temporal information, conventional machine learning models demonstrate comparable performance to the newly introduced, optimization-based TPL model. They treat each timestamp as an independent criterion, leveraging their robust predictive capabilities for handling high-dimensional data. Notably, these models mitigate data overfitting issues associated with increasing data dimensions. The UTADIS method, being a classic MCDA approach for MCS problems, exhibits the best precision performance but the worst recall performance. This indicates that most assignment results may go into the same class. It is less suited for scenarios with many criteria, particularly when each timestamp is treated separately. Consequently, it exhibits the weakest performance when temporal information is not considered. Additionally, it is noteworthy that the two state-of-the-art ordinal regression models, namely XOFM and UTADIS-T, exhibit superior performance solely in the realm of precision when compared with the proposed TPL and mRNN. This distinction arises because XOFM fails to consider temporal relationships inherent in time series data. In parallel, UTADIS-T relies solely on descriptive statistics to encapsulate preference-changing patterns, which proves less efficacious in mobile gaming scenarios characterized by highly stochastic in-game behaviors among users~\citep{zhao2023percltv}.

\begin{table}[h]
    \centering\footnotesize
    \caption{Experimental results for user value evaluation across four metrics. Means and standard deviations are calculated via ten-fold cross-validation. Bold highlights the top-performing results, while stars ($^*$) denote models significantly lower than the best-performing models in each metric (Wilcoxon statistic test $^{***} p<0.01, ^{**} p<0.05, ^{*} p<0.1$).\label{tab-baselineresults}}
    \begin{tabular}{cccccc}
            \toprule
        Method                & Precision  & Recall & F-score & Accuracy  \\ \midrule
        SVM               & 0.811 $\pm$ 0.026$^{***}$   & 0.746 $\pm$ 0.011$^{***}$     & 0.777 $\pm$ 0.016$^{***}$    & 0.781 $\pm$ 0.012$^{***}$  \\ 
        Logistic          & 0.822 $\pm$ 0.028           & 0.722 $\pm$ 0.016$^{***}$     & 0.768 $\pm$ 0.008$^{***}$    & 0.777 $\pm$ 0.009$^{***}$  \\ 
        RF                & 0.815 $\pm$ 0.032           & 0.754 $\pm$ 0.008$^{**}$      & 0.783 $\pm$ 0.017$^{***}$    & 0.786 $\pm$ 0.016$^{**}$  \\ 
        XGB               & 0.788 $\pm$ 0.031$^{***}$   & 0.762 $\pm$ 0.016$^{**}$      & 0.774 $\pm$ 0.017$^{***}$    & 0.773 $\pm$ 0.013$^{***}$  \\ 
        MLP               & 0.819 $\pm$ 0.030           & 0.744 $\pm$ 0.012$^{***}$     & 0.779 $\pm$ 0.012$^{***}$    & 0.784 $\pm$ 0.013$^{***}$  \\ 
        RNN               & 0.820 $\pm$ 0.023           & 0.753 $\pm$ 0.013$^{***}$     & 0.785 $\pm$ 0.013$^{***}$    & 0.789 $\pm$ 0.010$^{**}$  \\ 
        GRU               & 0.830 $\pm$ 0.031           & 0.752 $\pm$ 0.015$^{***}$     & 0.788 $\pm$ 0.014$^{**}$     & 0.793 $\pm$ 0.012  \\ 
        UTADIS            & \textbf{0.831 $\pm$ 0.028}  & 0.436 $\pm$ 0.027$^{***}$     & 0.571 $\pm$ 0.023$^{***}$    & 0.665 $\pm$ 0.015$^{***}$  \\ 
        UTADIS-T          & 0.823 $\pm$ 0.025           & 0.533 $\pm$ 0.019$^{***}$     & 0.647 $\pm$ 0.022$^{***}$    & 0.716 $\pm$ 0.018$^{***}$  \\ 
        XOFM              & 0.813 $\pm$ 0.022           & 0.625 $\pm$ 0.015$^{***}$     & 0.707 $\pm$ 0.015$^{***}$    & 0.727 $\pm$ 0.022$^{***}$  \\ 
        TPL               & 0.757 $\pm$ 0.092           & 0.784 $\pm$ 0.098             & 0.770 $\pm$ 0.023$^{***}$    & 0.745 $\pm$ 0.052$^{***}$  \\ \midrule
        mRNN $(\gamma=2$) & 0.804 $\pm$ 0.024$^{**}$    & 0.784 $\pm$ 0.034             & 0.794 $\pm$ 0.024            & 0.792 $\pm$ 0.018  \\ 
        mRNN $(\gamma=4$) & 0.804 $\pm$ 0.029$^{***}$   & 0.783 $\pm$ 0.017             & 0.793 $\pm$ 0.0183$^{*}$     & 0.791 $\pm$ 0.018$^{**}$  \\ 
        mRNN $(\gamma=6$) & 0.802 $\pm$ 0.042$^{**}$    & \textbf{0.789 $\pm$ 0.032}    & 0.794 $\pm$ 0.020            & 0.791 $\pm$ 0.021  \\ 
        mRNN $(\gamma=8$) & 0.814 $\pm$ 0.023$^{***}$   & 0.781 $\pm$ 0.028             & \textbf{0.797 $\pm$ 0.019}   & \textbf{0.796 $\pm$ 0.016}  \\ 
                \bottomrule
    \end{tabular}
\end{table}

Regarding the performance of DL-based methods, the findings highlight several important observations. First, traditional MCDA methods like UTADIS and standard machine learning techniques demonstrate comparatively lower performance. Their limitations in effectively capturing and utilizing temporal information contribute to this disparity. Second, DL-based models, in general, perform better in handling temporal criteria in MCS problems. The model equipped with the proposed mRNN structure stands out as the top performer in three measures. Its effectiveness can be attributed to integrating prior knowledge about the monotonic preference direction and incorporating both sub-marginal value functions and loss functions tailored for ordinal classes. It is worth noting that increasing the pre-defined number of sub-intervals in the mRNN model does not consistently enhance its efficacy. In fact, there is a risk of overfitting when using a more significant number of sub-intervals. This explains why the mRNN model with $\gamma=6$ outperforms the one with $\gamma=8$, underscoring the importance of carefully selecting hyperparameters to avoid overfitting.

\subsection{Predictive Performance When Considering Different Numbers of Timestamps}

\noindent In this section, we report the performance of the mRNN method when limiting its use to various numbers of timestamps. Specifically, its analysis is limited to the last 7 and 14 timestamps (hence omitting the initial period from consideration) or utilizing the entire time series (i.e., 24 timestamps). Table~\ref{tab-lagresult} presents the outcomes derived from this perspective.

\begin{table}[!ht]
    \centering\footnotesize
    \caption{Results accounting for the last 7 (a week) and 14 (half of a month) timestamps and the entire time series (i.e., 24 timestamps).
    \label{tab-lagresult}}
    \begin{tabular}{ccccc}
        \toprule
        Method & Precision & Recall & F-score & Accuracy  \\ \hline
        ~ &  ~ &$T=7$  & ~ & ~    \\ \hline
        mRNN $(\gamma=2$) & 0.7908 $\pm$ 0.0287$^{***}$  & 0.7483 $\pm$ 0.0236$^{**}$      & 0.7684 $\pm$ 0.0156$^{***}$       & 0.7692 $\pm$ 0.0129$^{***}$  \\ 
        mRNN $(\gamma=4$) & 0.7950 $\pm$ 0.0232$^{***}$  & 0.7516 $\pm$ 0.0165$^{***}$     & 0.7724 $\pm$ 0.0130$^{***}$       & 0.7732 $\pm$ 0.0104$^{***}$  \\ 
        mRNN $(\gamma=6$) & 0.7897 $\pm$ 0.0286$^{***}$  & 0.7627 $\pm$ 0.0244$^{**}$      & 0.7755 $\pm$ 0.0185$^{***}$       & 0.7742 $\pm$ 0.0138$^{***}$  \\ 
        mRNN $(\gamma=8$) & 0.7975 $\pm$ 0.0282$^{***}$  & 0.7542 $\pm$ 0.0247$^{**}$      & 0.7748 $\pm$ 0.0178$^{***}$       & 0.7756 $\pm$ 0.0143$^{***}$  \\ \hline
        ~ &  ~ & $T=14$  &~ & ~   \\ \hline
        mRNN $(\gamma=2$) & 0.8066 $\pm$ 0.0257$^{*}$    & 0.7720 $\pm$ 0.0166$^{*}$       & 0.7885 $\pm$ 0.0113$^{**}$        & 0.7878 $\pm$ 0.0118$^{***}$  \\ 
        mRNN $(\gamma=4$) & 0.8071 $\pm$ 0.0193          & 0.7697 $\pm$ 0.0235$^{**}$      & 0.7876 $\pm$ 0.0134$^{***}$       & 0.7875 $\pm$ 0.0110$^{**}$  \\ 
        mRNN $(\gamma=6$) & 0.8092 $\pm$ 0.0281          & 0.7720 $\pm$ 0.0272             & 0.7894 $\pm$ 0.0137$^{**}$        & 0.7891 $\pm$ 0.0136$^{**}$  \\ 
        mRNN $(\gamma=8$) & \textbf{0.8145 $\pm$ 0.0280} & 0.7689 $\pm$ 0.0215$^{*}$       & 0.7905 $\pm$ 0.0133               & 0.7914 $\pm$ 0.0114  \\ \hline
        ~ &  ~ & $T=24$  &~ & ~    \\ \hline
        mRNN $(\gamma=2$) & 0.8042 $\pm$ 0.0236          & 0.7839 $\pm$ 0.0339             & 0.7935 $\pm$ 0.0236               & 0.7917 $\pm$ 0.0183  \\ 
        mRNN $(\gamma=4$) & 0.8035 $\pm$ 0.0290          & 0.7834 $\pm$ 0.0171             & 0.7930 $\pm$ 0.0175$^{*}$         & 0.7906 $\pm$ 0.0176$^{**}$  \\ 
        mRNN $(\gamma=6$) & 0.8023 $\pm$ 0.0422          & \textbf{0.7890 $\pm$ 0.0321}    & 0.7943 $\pm$ 0.0200               & 0.7907 $\pm$ 0.0205  \\ 
        mRNN $(\gamma=8$) & 0.8137 $\pm$ 0.0225          & 0.7809 $\pm$ 0.0283             & \textbf{0.7966 $\pm$ 0.0192}      & \textbf{0.7961 $\pm$ 0.0163}  \\ 
        \bottomrule
    \end{tabular}
\end{table}

Our findings reveal the optimal performance of mRNN when leveraging the entire time series for analysis. With an augmentation in the time series length, the model gains a more comprehensive understanding of users' in-game purchase patterns, thereby enhancing predictive capabilities. Notably, the performance improvement becomes relatively marginal beyond $T=14$. This can be attributed to the inherent randomness characterizing online gaming users' behaviors, resulting in sparse records dominated by zero data points \citep{zhao2023percltv}. While longer sequences prove advantageous for deep learning-based models \citep{lecun2015deep}, their impact on the current user value is less pronounced compared to recent behaviors. Consequently, the observed improvements in predictions remain marginal.

\subsection{Interpreting User Preferences}

\noindent In this section, we analyze users' preferences, focusing on two distinct perspectives. The primary objective is to explore the user's marginal value function at each temporal timestamp, revealing its dynamic evolution over time. In~\ref{app-compglobalvalue}, we present a representative user's sub-marginal values, time discount factors, and marginal values at each timestamp. This way, we show how to obtain comprehensive values of alternatives and then determine the classification.

\begin{figure}[h]
\centering
\begin{subfigure}{\textwidth}
  \centering
    \includegraphics[width=\textwidth]{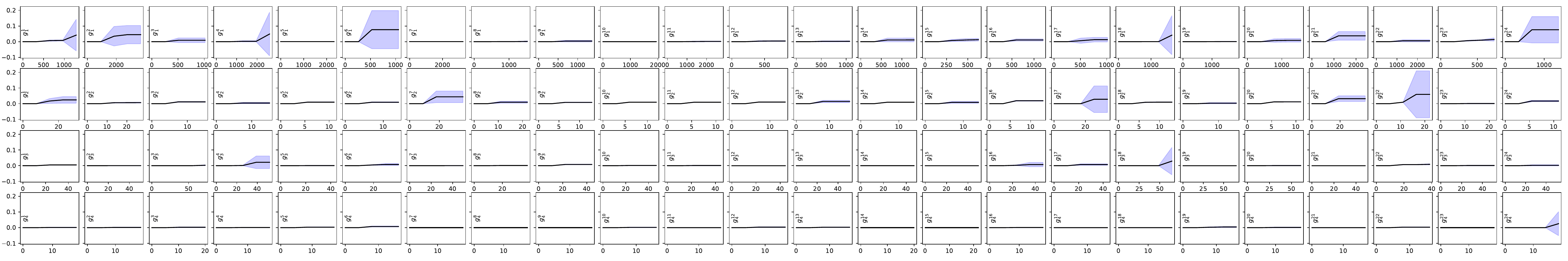}
    \caption{The sub-marginal value functions for each criterion at different timestamps learned by UTADIS ($\gamma=4$).}
  \label{fig:ltv_utadis_mgn_gamma}
\end{subfigure}
\begin{subfigure}{\textwidth}
  \centering
    \includegraphics[width=\textwidth]{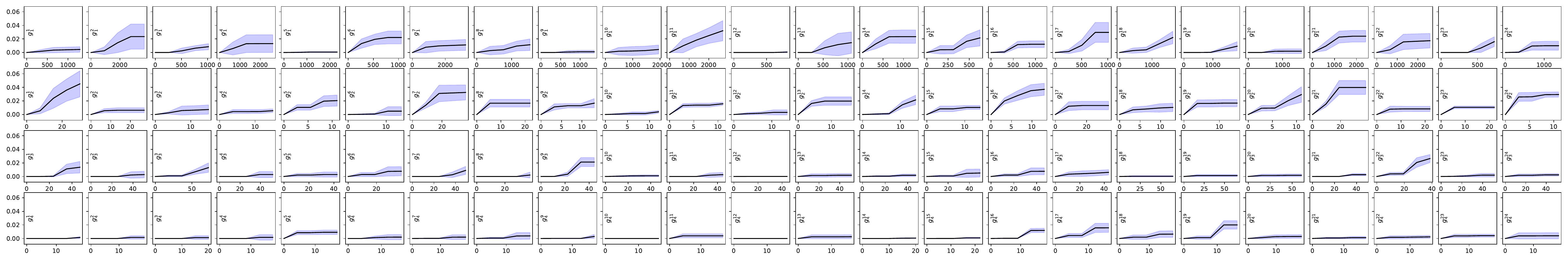}
    \caption{The sub-marginal value functions for each criterion at different timestamps learned by TPL ($\gamma=4$).}
  \label{fig:ltv_stpl_mgn_gamma}
\end{subfigure}
\begin{subfigure}{\textwidth}
  \centering
    \includegraphics[width=\textwidth]{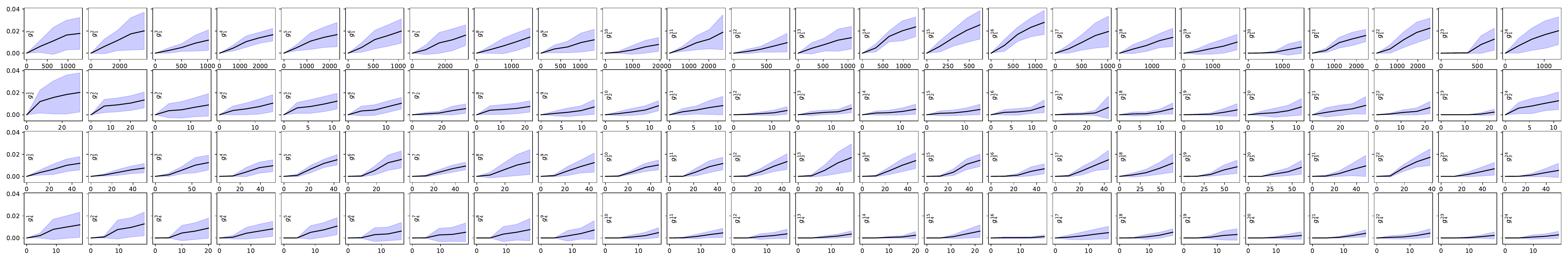}
    \caption{The sub-marginal value functions for each criterion at different timestamps learned by mRNN ($\gamma=4$).}
  \label{fig:ltv_mgn_gamma}
\end{subfigure}
\caption{The sub-marginal value functions for each criterion at different timestamps learned by the proposed approaches and MCDA baselines on the real-case dataset. In each sub-figure, the rows are marginal value functions of purchase amounts, purchase frequency, time spent, and log-in frequency, respectively. The solid line is the average marginal value, and the shaded region is one standard deviation across ten folds. For a fair comparison, we use $\gamma=4$ for all models as it is the optimal hyper-parameter for UTADIS and TPL. Note that the sub-marginal value functions are normalized using Eq.~(\ref{eq-normalization}).}
\label{fig:mgn_func_ltv}
\end{figure}

Figure \ref{fig:ltv_utadis_mgn_gamma} presents the results delivered by UTADIS. Notably, it exhibits a counterfactual pattern, with the marginal values consistently remaining at zero for most timestamps. Only a limited number of timestamps, specifically in the purchase amount criterion, appear to impact user value significantly. This unconventional pattern raises questions about the UTADIS method's ability to capture nuanced temporal information when analyzing user preferences effectively. 

In contrast, the marginal value functions derived from the proposed TPL method display a more coherent and dynamic pattern (see Figure~\ref{fig:ltv_stpl_mgn_gamma}). It assigns greater weight to the criteria purchase amount and purchase frequency, with most timestamps affecting discriminating user value. This aligns with the intuitive expectation that users who have spent more in terms of amount and frequency in the past are more likely to pay more in the future. However, these models neglect the impact of in-game activities, such as time spent in the game and log-in frequency.

Figure \ref{fig:ltv_mgn_gamma} represents the marginal value functions learned by the proposed mRNN. It confirms a more balanced weight to criteria such as in-game time spent and log-in frequency. Also, the marginal value functions do not approach zero as closely as those shown in Figure \ref{fig:ltv_stpl_mgn_gamma}. The flexibility introduced by the learnable time discount factors in the mRNN model allows for the adjustment of relationships across timestamps, taking into account the specific characteristics of each user. This adaptability enables the mRNN model to capture the influence of in-game activities on user preferences, even when such activities do not directly contribute to monetary value.

The analysis underscores the limitations of UTADIS in capturing temporal preferences. Moreover, it highlights the advantages of the proposed TPL and mRNN methods in modeling complex temporal relationships in user preferences. In particular, mRNN demonstrates the capability to consider a broader impact of criteria and timestamps, making it a valuable tool for understanding and predicting user behavior.

\begin{figure}[htbp]
\centering
\begin{subfigure}{.5\textwidth}
  \centering
  \includegraphics[trim=20 70 20 20, clip, width=1.1\columnwidth]{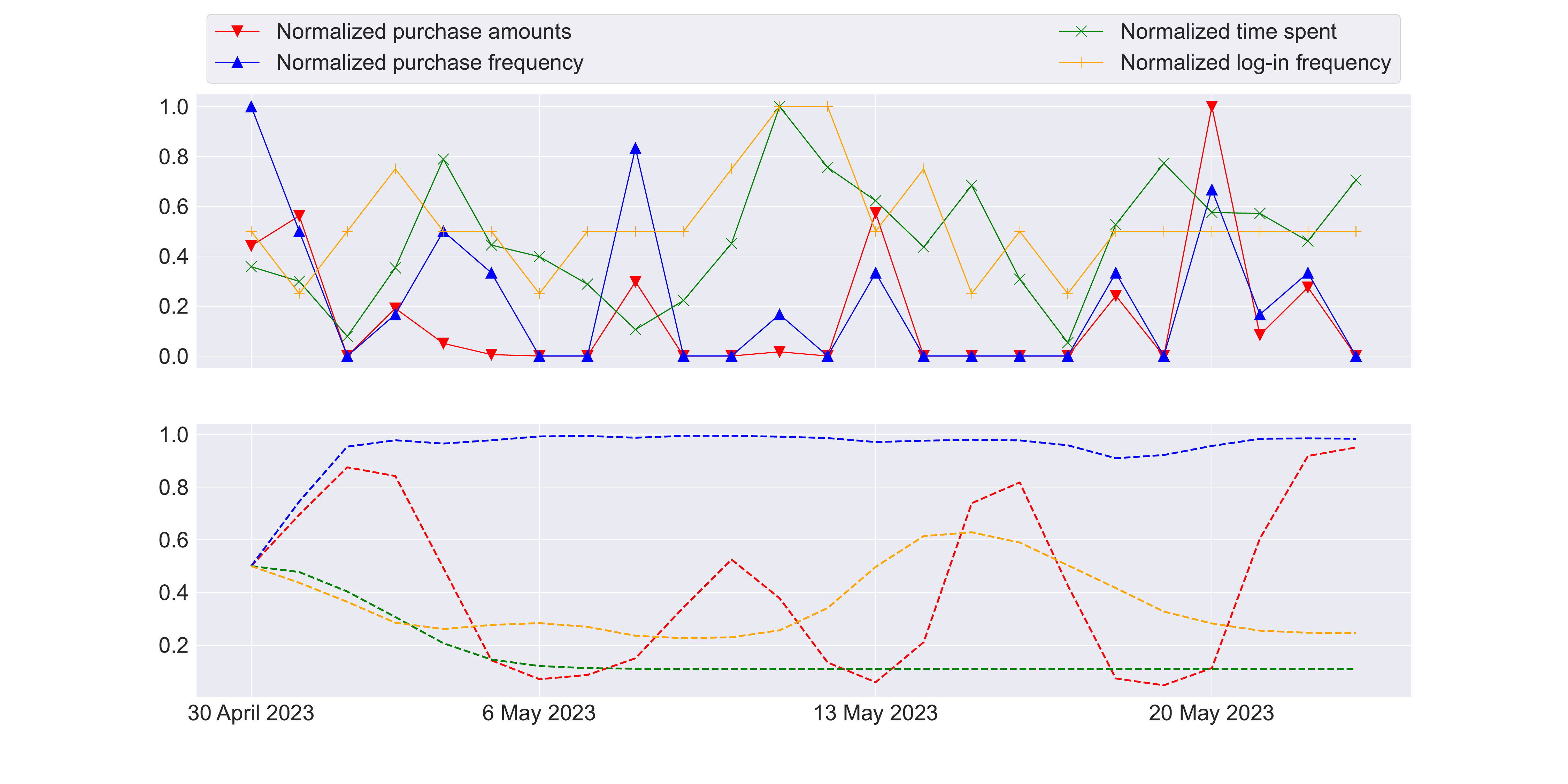}
  \caption{Representative high-value user with high time discount factors.\label{fig: high discount factors}}
  
\end{subfigure}%
\begin{subfigure}{.5\textwidth}
  \centering
    \includegraphics[trim=20 70 20 20, clip, width=1.1\columnwidth]{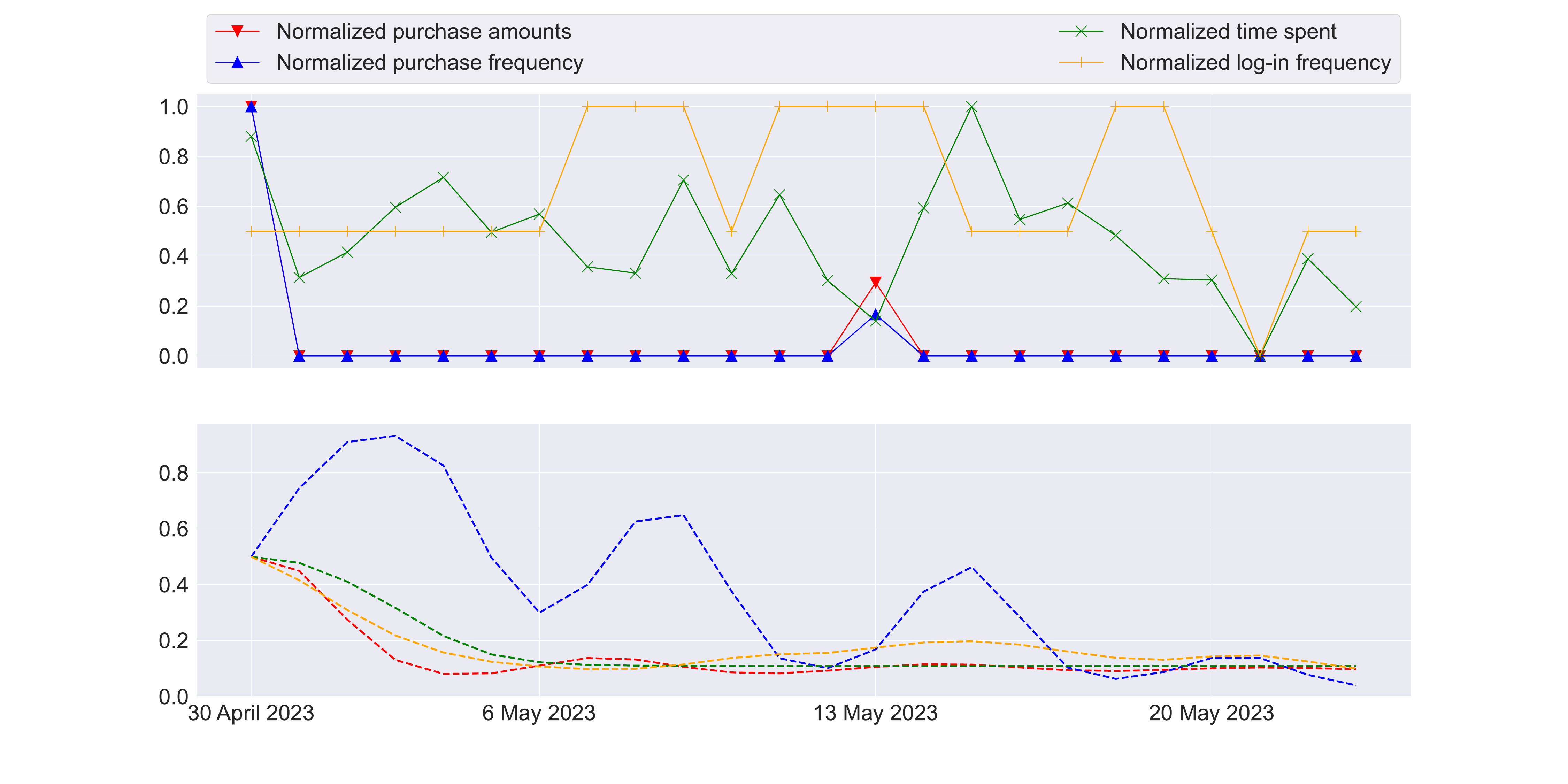}
  \caption{Representative low-value user with low time discount factors.\label{fig:low discount factors}}
\end{subfigure}
\caption{Representative users with different levels of time discount factors, represented by the dashed lines. \label{fig:discountfator}}

\end{figure}

The second aspect of our analysis involves an exploration of learnable time discount factors, affording us a deeper comprehension of users' temporal behaviors. Notably, this investigation is exclusively addressed only by the proposed mRNN model. As elucidated in Figure \ref{fig:discountfator}, we present a visual representation of the learned discount factors, considering distinct users as examples. Meanwhile, we also offer graphical depictions of the normalized purchase amount, purchase frequency, time spent, and log-in frequency to understand the implications of the learned time discount factors. 

Specifically, Figure \ref{fig: high discount factors} shows a high-valued user who frequently makes in-game purchases and log-in activities. In this scenario, the learned time discount factors for the purchase frequency criterion consistently maintain elevated values, reflecting the model's imperative to retain a comprehensive memory of valuable historical information. 
In contrast, Figure \ref{fig:low discount factors} portrays a low-valued user whose involvement in in-game purchases and log-in activities is notably sparse. Here, the learned time discount factors rapidly decline, signifying the scarcity of valuable historical data. This observation is further substantiated by the purchase frequency criterion, which exhibits a discernible fluctuating pattern of decrease, increase, and subsequent decrease, mirroring a different trend observed in Figure \ref{fig: high discount factors}.

\subsection{Managerial Implications}

\noindent Given the ubiquity of precision marketing, identifying high-value users is a pivotal reference for DM. Notably, prevailing MCDA methods exhibit proficiency in efficiently handling static criteria. However, challenges emerge as the data scale expands and becomes dynamic. In this context, user sequences, encapsulating personal behavioral patterns over time, emerge as valuable repositories of long-term insights for each user. The proposed models excel in offering precise predictions and interpretable outcomes pertaining to evolving user preference patterns over time. This capability proves particularly valuable in the following two applications.

Given the temporal analytical capabilities of the proposed models, a notable avenue of enhancement lies in alleviating the constraints associated with fixed recommended displays, thereby enabling what is termed ``popup recommendation''. In practical terms, this implies the flexibility to expose recommendations at any point in time. In the context of our mobile gaming scenario, a pivotal task involves determining the optimal timing and target user for recommendation triggering. Real-time execution of our model allows for identifying users with higher comprehensive values, signifying heightened purchase intentions and, consequently, an augmentation of in-game revenue. Furthermore, the analysis of sub-marginal value functions facilitates an in-depth exploration of the nuanced impacts of diverse historical behaviors on the differentiation of high-value users. This nuanced understanding serves as a valuable reference for DM in crafting personalized and effective marketing strategies.

In mobile gaming, another impactful application emerges in the form of churn intervention. The key to this intervention lies in targeting users predicted to be of high value, yet exhibiting a rapid decline in their time discount factors over the recent period. Such a decline signifies either inactivity or a heightened risk of user departure. Beyond merely discriminating high-value users, the visual representation of learnable time discount factors, exemplified in Figure~\ref{fig:discountfator}, offers a novel avenue for DM to identify those prone to churn. To proactively address this, the DM can devise custom tasks designed to incentivize continuous log-ins, subsequently rewarding users upon task completion. This strategic approach not only aids in user retention but also aligns with a personalized and targeted intervention strategy informed by predictive analytics.

\section{Simulation Experiments}

\noindent To further demonstrate the efficacy of the proposed models, in this section, we report the results of simulation experiments on four synthetic data with different complexity. We aim to answer the following questions: 
\begin{itemize}
\item Do the proposed models perform better than conventional machine learning and MCDA methods when the time series become complex? 
\item Are the learned marginal value functions able to describe the characteristics of the original data generators?
\item How does the proposed mRNN perform concerning different pre-defined numbers of sub-intervals?
\end{itemize}

\subsection{Data Generation Process and Experimental Settings}
\noindent The data generation process (DGP) employed in these experiments utilizes the Fourier decomposition method. This well-established technique allows for generating time series data with varying complexities and temporal patterns. It is based on the fundamental principle that any time series can be represented as a linear combination of sine or cosine basis functions.

\begin{figure}
\centering
\begin{subfigure}{.3\textwidth}
  \centering
  \includegraphics[scale=.3]{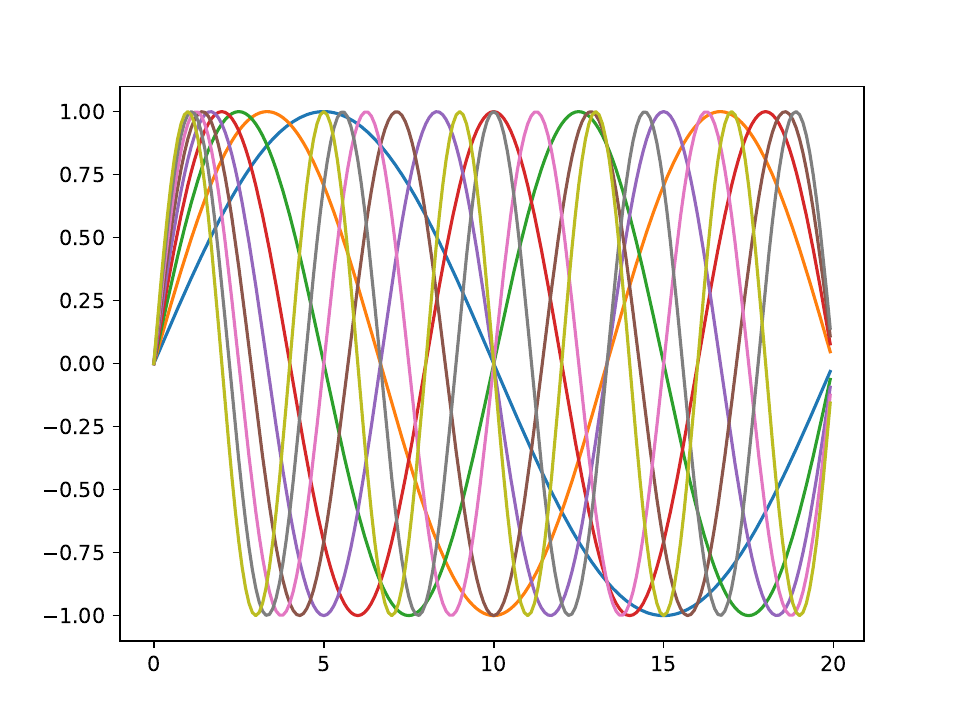}
  \caption{Nine basis functions in DGP.}
  \label{fig:basis function}
\end{subfigure}%
\begin{subfigure}{.3\textwidth}
  \centering
  \includegraphics[scale=.3]{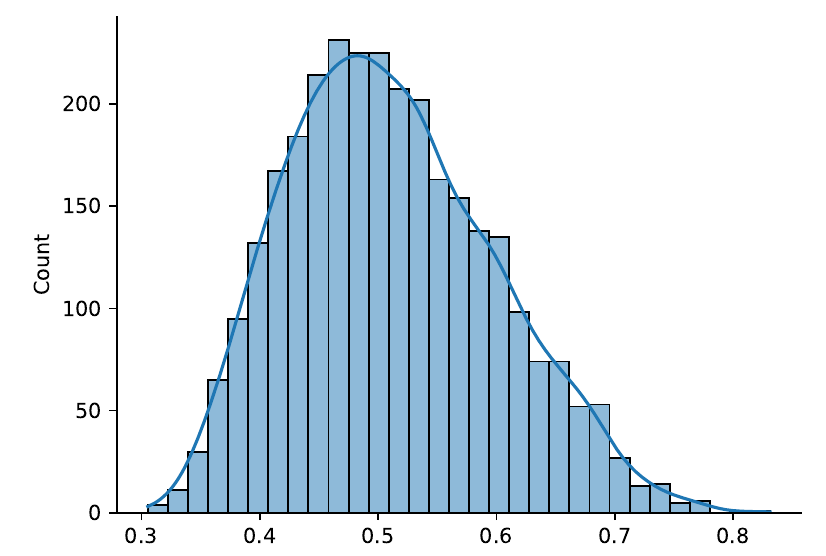}
  \caption{The distribution of comprehensive values. }
  \label{fig:utility function}
\end{subfigure}
\begin{subfigure}{.3\textwidth}
  \centering
  \includegraphics[scale=.3]{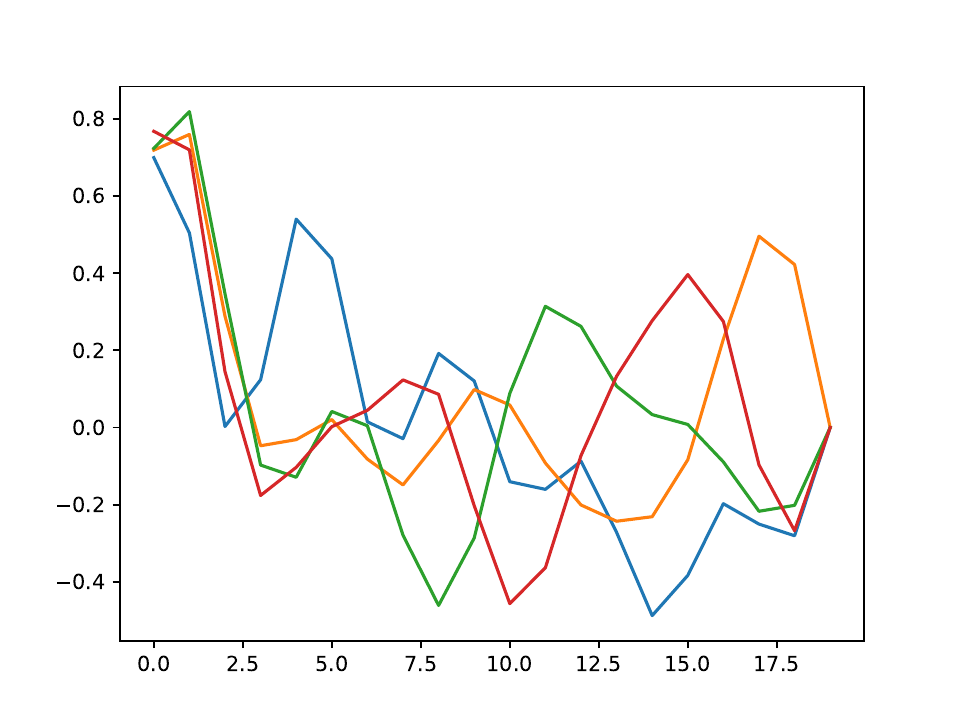}
  \caption{A sample with four distinct time series.}
  \label{fig:sim_sample}
\end{subfigure}
\caption{The basis functions, comprehensive value distribution, and simulated sample in the Basic simulation experiment. }
\label{fig:test}
\end{figure}

Specifically, we create a set of nine basis functions, each corresponding to a sine function with a distinct frequency. These frequencies range from 0.1 to 0.5 with an increment of 0.05, resulting in a set ${\sin(\omega_1 \pi t), \cdots, \sin(\omega_9 \pi t)}$ (see Figure \ref{fig:basis function}). The performance value at each timestamp is generated by a linear combination of these basis functions, with weights randomly sampled from a Dirichlet distribution, ensuring that they sum to one:
\begin{align}
    g_j^t(a)&=\beta_1\sin(\omega_1 \pi t)+\cdots+\beta_9\sin(\omega_9 \pi t),\quad \mbox{s.t.} \quad\sum\nolimits_{i=1}^9 \beta_i =1.
\end{align}
This way, we can generate diverse time series data with different frequencies and temporal patterns, which are essential for verifying the proposed models' capacity to handle temporal information:

The DGP provides a versatile framework for constructing time series with varying frequencies and temporal patterns. We pre-define sub-marginal value functions as the ground model and then define different procedures for generating marginal values and reference assignment examples given these sub-marginal values. The aim is to use the proposed models to restore the simulated assignment examples in different settings and verify whether they can capture the characteristics of the DGP settings. The latter include the time discount factor distributions and the monotonicity of the sub-marginal value functions. To achieve this goal, we have designed four distinct experiments, each catering to specific temporal characteristics:
\begin{itemize}
    \item Basic experiment: It represents the most straightforward scenario, establishing the simplest DGP for time series data. In this case, the sub-marginal value function, denoted as $f_j^t(a)$, is a monotonically transformed function of the performance value $g_j^t(a)$ at the current timestamp. Specifically, we define it as $f_j^t(a) = \tanh(g_j^t(a))$. .
    \item Non-Markovian experiment: It introduces temporal dependencies into the sub-marginal value function, which not only depends on the performance value at the current timestamp but also incorporates the performance value from the previous timestamp. Mathematically, it is defined as $f_j^t(a)=\tanh \left(g_j^t(a)+g^{t-1}_j(a)\right)$. 
    \item Non-monotonic experiment: It incorporates sub-marginal value functions deliberately made non-monotonic. Instead of a simple transformation, the function for criterion $j$ at timestamp $t$ is determined by a sine function: $f_j^t(a) = \sin \left(2^{j-1}\pi g_j^t(a)\right)$. This experiment explores the models' ability to handle non-monotonic relationships in the data.
    \item Non-independent experiment: It represents the most complex scenario, where the assumption of independence between criteria is relaxed. To achieve this, we modify the DGP by introducing an interaction term into the sub-marginal value function. Specifically, it is defined as $f^t(a) = \tanh \left( g_1^t(a)+g_2^t(a) \right) + \tanh \left( g_3^t(a)+g_4^t(a)\right)$. This experiment assesses the models' performance when criteria interact, making the data more intricate.
\end{itemize}
\noindent In each experiment, $u_j^t(a)$ is determined by iteratively applying a time discount factor $\tau_{j,t-1}$ to the previous marginal value $u_j^{t-1}(a)$, along with the current sub-marginal value $f_j^t(a)$, i.e., $u_j^{t}(a)=\tau_{j,t-1}u_j^{t-1}(a)+ (1-\tau_{j,t-1})f_j^t(a)$, where $\tau_{j,t-1}= \frac{1}{1+\exp(-u_j^{t-2}(a))}$, indicating that a higher marginal value exerts a more significant influence on the subsequent period.



Deriving comprehensive values from the temporal marginal value functions involves summing up all the values of marginal functions at the last timestamp. Such a comprehensive value is then transformed to the range between 0 and 1 using a Sigmoid function. The distribution of these comprehensive values for all the data points in the Basic experiment is depicted in Figure \ref{fig:utility function}. To determine class labels, samples with a comprehensive value less than 0.5 are labeled as negative samples, while those with a comprehensive value at least as good as 0.5 are labeled as positive samples. This data generation process ensures that the time series data exhibits dynamic temporal patterns, making it well-suited for training and evaluating models like mRNN.

For each of the four experiments, we generated a total of 3,000 data samples. Each data sample consists of four distinct time series, and all time series have a fixed length of 20 timestamps. Figure~\ref{fig:sim_sample} presents an example data sample. We adopt a robust ten-fold cross-validation approach for data splitting to ensure an unbiased and reliable evaluation of the model's performance. In each fold, the data is further partitioned into three distinct subsets: the training set, the validation set, and the test set, with proportions of 60\%, 20\%, and 20\%, respectively. This systematic approach allows us to assess the models' generalization performance effectively.

\subsection{Experimental Results}

\noindent The experimental results, shown in Table \ref{tab:simulation_res_1}, are averaged across the ten folds to derive the $F-score$, comprehensively evaluating the models' generalization capabilities. The detailed results regarding $precision$, $recall$, and $accuracy$ measures are presented in \ref{app-compresult}.

\begin{table}[tbh]
    \centering\footnotesize
    \caption{Experimental results ($F-score$) for simulation experiment. Means and standard deviations are calculated via ten-fold cross-validation. Bold highlights the top-performing methods, while stars ($^*$) denote models significantly worse than the best-performing models ($^{***} p<0.01, ^{**} p<0.05, ^{*} p<0.1$).\label{tab:simulation_res_1}}
    \begin{tabular}{cccccc}
            \toprule
        Method               & Basic & Non-Markovian & Non-monotonic & Non-independent  \\ \midrule
        SVM               & 0.913 $\pm$ 0.009$^{***}$     & 0.874 $\pm$ 0.016$^{***}$     & 0.588 $\pm$ 0.028$^{***}$     & 0.899 $\pm$ 0.016$^{***}$  \\ 
        Logistic          & 0.906 $\pm$ 0.008$^{***}$     & 0.881 $\pm$ 0.009$^{***}$     & 0.620 $\pm$ 0.024$^{***}$     & 0.925 $\pm$ 0.009$^{***}$  \\ 
        RF                & 0.924 $\pm$ 0.007$^{***}$     & 0.886 $\pm$ 0.019$^{***}$     & 0.623 $\pm$ 0.027$^{***}$     & 0.839 $\pm$ 0.017$^{***}$ \\ 
        XGB               & 0.935 $\pm$ 0.009$^{***}$     & 0.910 $\pm$ 0.007$^{***}$     & \textbf{0.661 $\pm$ 0.024}    & 0.887 $\pm$ 0.011$^{***}$  \\ 
        MLP               & 0.914 $\pm$ 0.006$^{***}$     & 0.878 $\pm$ 0.009$^{***}$     & 0.608 $\pm$ 0.021$^{***}$     & 0.920 $\pm$ 0.012$^{***}$  \\ 
        RNN               & 0.912 $\pm$ 0.008$^{***}$     & 0.882 $\pm$ 0.013$^{***}$     & 0.610 $\pm$ 0.024$^{***}$     & \textbf{0.953 $\pm$ 0.008}  \\ 
        GRU               & 0.912 $\pm$ 0.008$^{***}$     & 0.884 $\pm$ 0.009$^{***}$     & 0.600 $\pm$ 0.019$^{***}$     & 0.952 $\pm$ 0.009  \\ 
        UTADIS            & 0.891 $\pm$ 0.015$^{***}$     & 0.892 $\pm$ 0.011$^{***}$     & 0.585 $\pm$ 0.025$^{***}$     & 0.857 $\pm$ 0.013$^{***}$  \\ 
        UTADIS-T          & 0.897 $\pm$ 0.018$^{***}$     & 0.900 $\pm$ 0.016$^{***}$     & 0.600 $\pm$ 0.025$^{***}$     & 0.865 $\pm$ 0.011$^{***}$  \\ 
        XOFM              & 0.904 $\pm$ 0.019$^{***}$     & 0.911 $\pm$ 0.010$^{***}$     & 0.600 $\pm$ 0.029$^{***}$     & 0.875 $\pm$ 0.021$^{***}$  \\         
        TPL               & 0.893 $\pm$ 0.018$^{***}$     & 0.909 $\pm$ 0.030$^{***}$     & 0.606 $\pm$ 0.048$^{**}$      & 0.894 $\pm$ 0.020$^{***}$\\ \midrule
        mRNN $(\gamma=2$) & 0.943 $\pm$ 0.011$^{***}$     & 0.920 $\pm$ 0.013$^{***}$     & 0.634 $\pm$ 0.029$^{**}$      & 0.906 $\pm$ 0.011$^{***}$ \\ 
        mRNN $(\gamma=4$) & 0.952 $\pm$ 0.017$^{*}$       & 0.940 $\pm$ 0.016$^{**}$      & 0.636 $\pm$ 0.025$^{**}$      & 0.901 $\pm$ 0.014$^{***}$\\ 
        mRNN $(\gamma=6$) & \textbf{0.961 $\pm$ 0.005}    & 0.948 $\pm$ 0.012             & 0.638 $\pm$ 0.024$^{**}$      & 0.902 $\pm$ 0.015$^{***}$\\ 
        mRNN $(\gamma=8$) & 0.959 $\pm$ 0.008             & \textbf{0.952 $\pm$ 0.006}    & 0.624 $\pm$ 0.031$^{***}$     & 0.899 $\pm$ 0.016$^{***}$\\ 
                \bottomrule
    \end{tabular}
\end{table}

The superior performance of the proposed mRNN model in the basic and non-Markovian experiments across various values of $\gamma$ can be attributed to its ability to overcome intrinsic limitations present in other baseline models. Conventional machine learning models are not well-equipped to effectively capture the temporal dependencies inherent in sequential data, which leads to suboptimal results in scenarios with time series criteria. In turn, traditional deep learning models like RNN and GRU can model temporal dependencies but typically compute the comprehensive value of each sample only at the final timestamp, potentially discarding crucial information from earlier timestamps. Additionally, these models output probabilities for each category, necessitating a manual decision on classification by, e.g., fixing a threshold at 0.5 for binary classification tasks. Nevertheless, such pre-determined thresholds may not be universally applicable, thus potentially yielding suboptimal performance in some cases.

Given these limitations, the proposed mRNN model is a promising solution that effectively addresses the abovementioned challenges. The proposed adaptive loss function also empowers the model to dynamically learn and determine classification thresholds based on the underlying data distribution, imparting flexibility and precision in decision-making. The outcomes of the Basic experiment affirm these assertions. Additionally, the non-Markovian experiment serves to reinforce these findings, as it elucidates that the performance of mRNN remains unaltered in the presence of a non-Markovian sub-marginal value function. In contrast, the performance of all comparative baselines experiences a decline.

Recall that the proposed mRNN satisfies the preference independence and monotonicity assumptions, which may decline the model performance when the data become complex. This is confirmed by the results of Non-monotonic and Non-independent experiments (see Table \ref{tab:simulation_res_1}). When the sub-marginal value functions are not monotonic, our model's performance is not the highest. However, it closely follows XGB, which is ranked as the top-performing method. It demonstrates that although we have constrained the monotonicity of the sub-marginal value function, the proposed model can provide remarkable robustness in non-monotonic cases, achieving comparable performance with the baseline methods. As for the preference-dependent case, GRU and RNN models outperform all other methods, highlighting the superior performance of deep learning-based sequence models in capturing interactive information between criteria. In contrast, the proposed model assigns an independent hidden state to each criterion and utilizes separate parameters for training. While this design choice ensures complete independence between criteria, it also leads to a potential loss of interactive information, deteriorating our model's effectiveness.

\subsection{Ability for Reconstructing Sub-marginal Value Functions}

\begin{figure}[h]
\centering
    \begin{subfigure}{\textwidth}
      \includegraphics[width=\columnwidth]{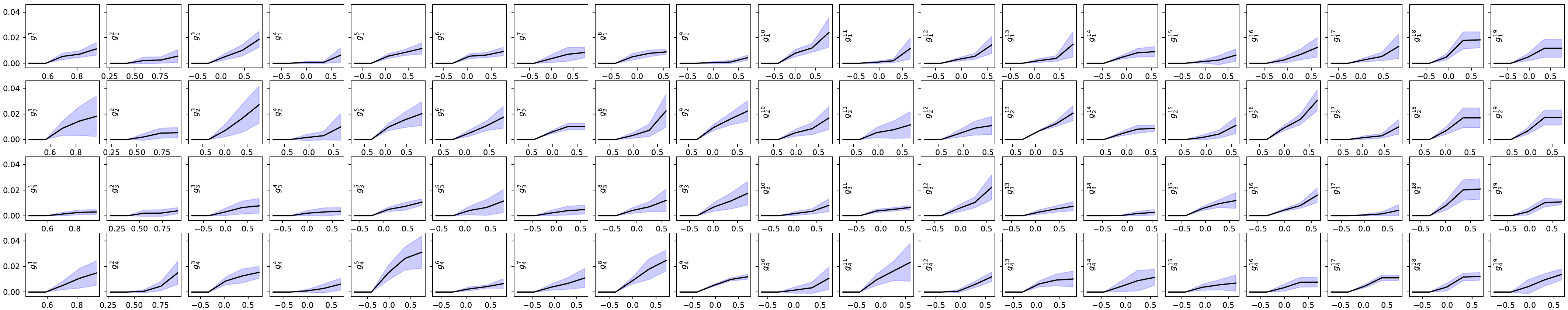}
    \caption{The marginal value functions learned by UTADIS on basic simulation dataset ($\gamma=4$).}
      \label{fig:sim_utadis_gamma4}
    \end{subfigure}
    \begin{subfigure}{\textwidth}
    \includegraphics[width=\columnwidth]{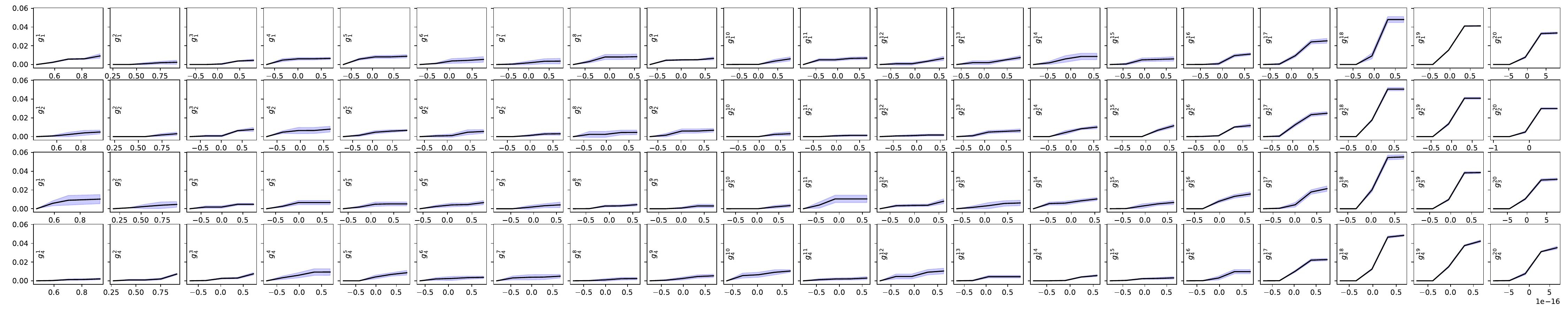}
    \caption{The marginal value functions learned by TPL on basic simulation dataset ($\gamma=4$).}
    \label{fig:sim_stpl_gamma4}
    \end{subfigure}
    \begin{subfigure}{\textwidth}
    \includegraphics[width=\columnwidth]{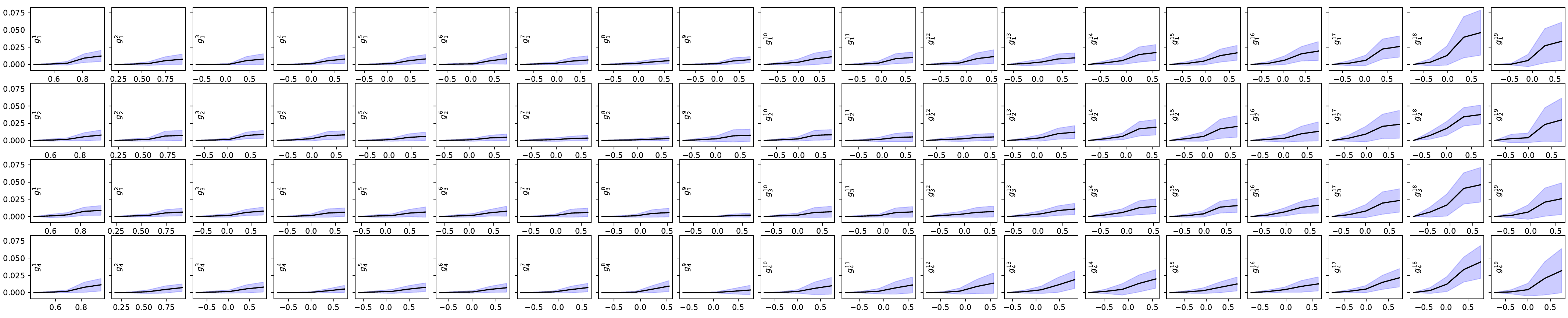}
    \caption{The marginal value functions learned by mRNN on basic simulation dataset ($\gamma=4$).}
            \label{fig:sim_mrnn_gamma4}
    \end{subfigure}%
    \caption{The marginal value functions learned by the proposed approaches and MCDA baselines on the basic simulation dataset. The solid line is the average marginal value, and the shaded region is one standard deviation across ten folds. For the purpose of fairness comparison, we present $\gamma=4$ for all the models as it is the optimal hyper-parameter for the UTADIS and TPL models. Note that the sub-marginal value functions are normalized using Eq.~(\ref{eq-normalization}). }
    
    \label{fig:sim_mgn_func}
\end{figure}

\noindent In addition to demonstrating superior classification performance, the methods proposed in this paper offer the advantage of interpretability owing to the incorporation of prior knowledge regarding monotonic constraints. During DGP, we impose a $tanh$ function as the sub-marginal value function for all time series. We can use this prior knowledge to visualize and analyze the learned sub-marginal value functions at different timestamps. The results, depicted in Figures \ref{fig:sim_stpl_gamma4} and~\ref{fig:sim_mrnn_gamma4}, reveal that all the learned sub-marginal value functions from proposed TPL and mRNN models closely resemble the $tanh$ function's characteristic shape. This observation confirms that the model effectively captures and learns the inherent monotonic constraints presented in the data.

Furthermore, the analysis of Figures~\ref{fig:sim_stpl_gamma4} and~\ref{fig:sim_mrnn_gamma4} indicates that the values from the marginal value function at later timestamps consistently exhibit higher values than those from the function at preceding timestamps. This observation implies that the later time points exert a pronounced influence on the classification outcomes, which align with the underlying DGP, wherein the marginal values associated with earlier timestamps progressively diminish over time, resulting in a reduced impact on classification decisions when compared to more recent timestamps. In contrast, the conventional UTADIS approach, lacking the capacity to capture temporal effects, manifests a static marginal value function across all timestamps.

\subsection{Model Performance for Different Numbers of Sub-intervals}

\noindent The performance of the mRNN model in the simulation experiments is influenced by the number of pre-defined sub-intervals, as shown in Table~\ref{tab:simulation_res_1}. Interestingly, increasing the number of sub-intervals does not consistently lead to improved mRNN performance. In the first two simulation experiments, the mRNN model achieves the best results when $\gamma=6$, and smaller numbers of sub-intervals yield better performance. However, in the last two simulation experiments, where the data complexity is higher due to non-monotonic and non-independent time series, mRNN performs better with smaller values of $\gamma$.

This behavior can be explained by considering the trade-off between model complexity and overfitting. More sub-intervals lead to a more complex model, which has the potential to capture intricate patterns in the data. However, this increased complexity also raises the risk of overfitting, especially when the dataset size is limited. In contrast, a~simpler model with fewer sub-intervals may generalize better when the data is noisy or exhibits non-linear patterns.

The observation that a smaller pre-defined number of sub-intervals and a slightly higher $\gamma$ can be effective aligns with the findings from the real-world case study, where $\gamma=6$ achieved the best average $F-score$. This suggests that in practice, it may be advisable to start with a small number of pre-defined sub-intervals and adjust $\gamma$ based on the specific requirements and characteristics of the decision problem and dataset. This approach allows for flexibility in model complexity and can help avoid overfitting while still meeting the DM's needs.

\section{Conclusions}

\noindent This study considered the sorting problems in the presence of temporal criteria. Such criteria involve time series data that may represent preference patterns changing over time. The capability to handle such non-static features is essential for solving numerous real-world problems in economy, energy, and medicine. 

We proposed two value-based approaches for handling such problems, employing additive piecewise linear functions as the basic preference model for each timestamp. On the one hand, we formulated a convex quadratic programming model with a fixed time discount factor under a~regularization framework. On the one hand, we proposed a~deep preference learning model that allows the time discount factors to be learnable, hence investigating the dynamics of changing preferences over time. It satisfies monotonicity and preference independence conditions and captures natural order between classes by adapting the recurrent neural network. 

We applied the proposed models to a real-world decision-making problem where the mobile game users are assigned to different classes based on their historical behavioral sequences. The study provides an in-depth analysis of users' evolving preferences captured via marginal value functions at all timestamps as well as the learned time discount factors, affording a more profound comprehension of temporal behaviors. These outcomes may be used for recommendation triggering and churn intervention, which is essential in the mobile gaming industry.

We also tested the proposed methods on synthetic data with various temporal characteristics. We demonstrated the superior performances of the proposed optimization- and neural network-based models over the baseline multiple criteria decision aiding, machine learning, and deep learning approaches. Among the novel methods, the one based on suitably adapted recurrent networks proved more computationally efficient and expressive, leading to better predictive performance than the optimization model.

Future work can focus on more complex time series. For instance, the interval of timestamps can be irregularly spaced, which cannot be directly handled by the standard RNN model. Another direction is relaxing the monotonicity and preference independence in temporal criteria, which can describe a more complicated decision scenario while simultaneously making the preference model highly complex and difficult to optimize. At last, it is interesting to analyze time series where the performances on some criteria exhibit various types of concept drifts.

\section*{Acknowledgments}
\noindent Mi{\l}osz Kadzi{\'n}ski acknowledges support from the Polish National Science Center under the SONATA BIS project (grant no. DEC-2019/34/E/HS4/00045). 


\bibliographystyle{elsarticle-harv}
\bibliography{mybib}

\appendix 
\section{Proof of the Transformation from $P0$ to $P1$}
\label{app-proof}
\begin{proof}
The Lagrange function of $P0$ is:
\begin{equation}
    \mathcal{L}(\mathbf{u}, \boldsymbol{\xi}, \boldsymbol{\mu}, \boldsymbol{\eta}, \boldsymbol{\sigma}) = \frac{1}{2} \left\Vert \mathbf{u}\right\Vert ^2_2 + C\cdot\sum_{i=1}^N \xi_i - \sum_{i=1}^N  \mu_i \left[y_i \mathbf{u}' \mathbf{v}_i + \xi_i -1 \right] - \sum_{i=1}^N \eta_i \xi_i - \sum_{j=1}^m\sum_{t=1}^{T_j}\sum_{k=1}^{\gamma_j^t}\Delta f_j^{t,k}\cdot\sigma_j^{t,k}
\end{equation}
where $\boldsymbol{\xi}=(\xi_1,\ldots,\xi_N)', \boldsymbol{\mu}=(\mu_1,\ldots,\mu_N)', \boldsymbol{\eta}=(\eta_1, \ldots,\eta_N)', \boldsymbol{\sigma}=(\underbrace{\sigma_1^{1,1},\ldots,\sigma_1^{1,\gamma_1^1}}_{\text{1st timestamp on criterion } g_1  } \ldots, \sigma_m^{T_m,\gamma_m^{T_m}} )'$, and $\mathbf{v}_i=(V_{i,1}^{1,1}, \ldots, V_{i,j}^{t,k}, \ldots, V_{i,m}^{T_m, \gamma_m^{T_m}})'$. We set derivatives of $\mathcal{L}(\mathbf{u}, \boldsymbol{\xi}, \boldsymbol{\mu}, \boldsymbol{\eta}, \boldsymbol{\sigma})$ w.r.t. $\mathbf{u}$ and $\boldsymbol{\xi}$ to be zero, respectively:
\begin{align}
    &\frac{\partial \mathcal{L}(\mathbf{u}, \boldsymbol{\xi}, \boldsymbol{\mu}, \boldsymbol{\eta}, \boldsymbol{\sigma})}{\partial \Delta f_j^{t,k}} = \Delta f_j^{t,k} -\sum_{i=1}^N y_i \mu_i V_{i,j}^{t,k} - \sigma_j^{t,k} =0 \\
    &\Rightarrow \Delta f_j^{t,k} = \sum_{i=1}^N y_i \mu_i V_{i,j}^{t,k} + \sigma_j^{t,k} \label{eq-der1} \\
    & \frac{\partial \mathcal{L}(\mathbf{u}, \boldsymbol{\xi}, \boldsymbol{\mu}, \boldsymbol{\eta}, \boldsymbol{\sigma})}{\partial \xi_i} = C - \mu_i - \eta_i = 0 \\
    & \Rightarrow C\cdot \sum_{i=1}^N \xi_i= \sum_{i=1}^N \mu_i\xi_i - \sum_{i=1}^N \eta_i\xi_i \label{eq-der2}
\end{align}
Given Eq.~(\ref{eq-der2}), the Lagrange function can be simplified:
\begin{equation}
    \mathcal{L}(\mathbf{u},\boldsymbol{\mu}, \boldsymbol{\sigma}) =  \frac{1}{2} \left\Vert \mathbf{u}\right\Vert ^2_2 - \sum_{i=1}^N \mu_i \left(y_i \mathbf{u}' \mathbf{v}_i -1 \right) - \sum_{j=1}^m\sum_{t=1}^{T_j}\sum_{k=1}^{\gamma_j^t}\Delta f_j^{t,k}\cdot\sigma_j^{t,k} \label{eq-lagfunc2}
\end{equation}
Given Eq.~(\ref{eq-der1}), the last term in Eq.~(\ref{eq-lagfunc2}) can be further expressed as:
\begin{align}
    \sum_{j=1}^m\sum_{t=1}^{T_j}\sum_{k=1}^{\gamma_j^t}\Delta f_j^{t,k}\cdot\sigma_j^{t,k} &= \sum_{j=1}^m\sum_{t=1}^{T_j}\sum_{k=1}^{\gamma_j^t} \Delta f_j^{t,k}\left(\Delta f_j^{t,k} - \sum_{i=1}^N y_i \mu_i V_{i,j}^{t,k}  \right) \\
    & = \sum_{j=1}^m\sum_{t=1}^{T_j}\sum_{k=1}^{\gamma_j^t} (\Delta f_j^{t,k})^2 - \sum_{j=1}^m\sum_{t=1}^{T_j}\sum_{k=1}^{\gamma_j^t} \Delta f_j^{t,k} (\sum_{i=1}^N y_i \mu_i V_{i,j}^{t,k}) \\ 
    & = \left\Vert \mathbf{u}\right\Vert ^2_2 - \sum_{i=1}^N y_i\mu_i \left( \sum_{j=1}^m\sum_{t=1}^{T_j}\sum_{k=1}^{\gamma_j^t} \Delta f_j^{t,k} V_{i,j}^{t,k}  \right) \\
    & = \left\Vert \mathbf{u}\right\Vert ^2_2 - \sum_{i=1}^N y_i \mu_i \mathbf{u}' \mathbf{v}_i \label{eq-lagfunc3}
\end{align}
Hence, Eq.~(\ref{eq-lagfunc2}) can be rewritten as:
\begin{equation}
     \mathcal{L}(\boldsymbol{\mu}, \boldsymbol{\sigma}) = \sum_{i=1}^N\mu_i - \frac{1}{2} \sum_{j=1}^m\sum_{t=1}^{T_j}\sum_{k=1}^{\gamma_j^t} \left( \sum_{i=1}^N y_i \mu_i  V_{i,j}^{t,k} + \sigma_j^{t,k} \right)
\end{equation}
The dual problem of P0 can be:
\begin{align}
    \max g(\boldsymbol{\mu}, \boldsymbol{\eta}, \boldsymbol{\sigma}) &= \inf_{\mathbf{u}, \boldsymbol{\xi}} \mathcal{L}(\mathbf{u}, \boldsymbol{\xi}, \boldsymbol{\mu}, \boldsymbol{\eta}, \boldsymbol{\sigma}) \\
    & = \sum_{i=1}^N \mu_i - \frac{1}{2} \sum_{j=1}^m\sum_{t=1}^{T_j}\sum_{k=1}^{\gamma_j^t} \left( \sum_{i=1}^N y_i \mu_i  V_{i,j}^{t,k} + \sigma_j^{t,k} \right) \\
    s.t.: \quad & \boldsymbol{\sigma} \ge {\mathbf{0}} \\
    & C\mathbf{e} \ge \boldsymbol{\mu} \ge \mathbf{0}
\end{align}
From Eq.~(\ref{eq-der1}), we can replace $ \sigma_j^{t,k} $ by $ \Delta f_j^{t,k} - \sum_{i=1}^N y_i \mu_i V_{i,j}^{t,k} $ and obtain P1:
\begin{align}
    \text{(P1)} \quad & \min \frac{1}{2} \left\Vert {\mathbf{u}} \right\Vert^2_2 - \sum\nolimits_{i=1}^N \mu_i \nonumber \\
    s.t.:\quad & \Delta f_j^{t,k} - \sum\nolimits_i^N y_i \mu_i V_{i,j}^{t,k} \ge 0, j=1,\ldots,m, t=1,\ldots, T_j, k = 1,\ldots,\gamma_j^{t} \\
    & C\ge \mu_i  \ge 0, i=1,\ldots,N 
\end{align}
\end{proof}

\section{Proof of Proposition \ref{prop-1}}
\label{app-prop1}
\begin{proof}
    Recall the definition of the sub-marginal values of alternative $a$ in Eq.~(\ref{eq-9}), we can rewrite the sub-marginal value at $t$-th timestamp in $j$-th criterion:
    \begin{align}
        f_j^t(g_j^t(a)) = f_j^t(x_j^{t,k}) + \frac{g_j^t(a) - x_j^{t,k}}{x_j^{t,k+1}-x_j^{t,k}}\left( f_j^t(x_j^{t,k+1}) - f_j^t(x_j^{t,k}) \right) 
    \end{align}
Following the transformation in Eq.~(\ref{eq-normalization}), we can easily get:
\begin{align}
    f_j^t(g_j^t(a))' &= \frac{f_j^t(x_j^{t,k})-f_j^t(x_j^{t,0})}{\sum\nolimits_{j,t} (f_j^t(x_j^{t,\gamma_j^t}) - f_j^t(x_j^{t,0})) } + \frac{g_j^t(a) - x_j^{t,k}}{x_j^{t,k+1}-x_j^{t,k}}\left[ \frac{f_j^t(x_j^{t,k+1})-f_j^t(x_j^{t,0})}{\sum\nolimits_{j,t} (f_j^t(x_j^{t,\gamma_j^t}) - f_j^t(x_j^{t,0})) } - \frac{f_j^t(x_j^{t,k})-f_j^t(x_j^{t,0})}{\sum\nolimits_{j,t} (f_j^t(x_j^{t,\gamma_j^t}) - f_j^t(x_j^{t,0})) } \right] \\
    &= \frac{f_j^t(x_j^{t,k}) + \frac{g_j^t(a) - x_j^{t,k}}{x_j^{t,k+1}-x_j^{t,k}}\left( f_j^t(x_j^{t,k+1}) - f_j^t(x_j^{t,k}) \right) - f_j^t(x_j^{t,0})}{\sum\nolimits_{j,t} (f_j^t(x_j^{t,\gamma_j^t}) - f_j^t(x_j^{t,0})) } \\
    &= \frac{f_j^t(g_j^t(a)) - f_j^t(x_j^{t,0})}{\sum\nolimits_{j,t} (f_j^t(x_j^{t,\gamma_j^t}) - f_j^t(x_j^{t,0})) } \label{eq-transmargin}
\end{align}
Using Eq.~(\ref{eq-transmargin}) to express the transformed comprehensive value $U'(a)$, we can get:
\begin{align}
    U'(a) &= \frac{\sum_{j=1}^m \left[ f_j^{T_j}(g_j^{T_j}(a)) - f_j^{T_j}(x_j^{T_j,0}) \right]}{\sum\nolimits_{j,t} (f_j^t(x_j^{t,\gamma_j^t}) - f_j^t(x_j^{t,0})) } + \frac{\sum_{t=1}^{T_j-1} \tau^{T_j-t}\left[ f_j^t(g_j^t(a)) - f_j^t(x_j^{t,0}) \right] }{\sum\nolimits_{j,t} (f_j^t(x_j^{t,\gamma_j^t}) - f_j^t(x_j^{t,0})) }\\
    &= \frac{\sum_{j+1}^{m} \left[ f_j^{T_j}(g_j^{T_j}(a)) + \sum_{t=1}^{T_j-1}\tau^{T_j-t}f_j^t(g_j^t(a)) \right] }{\sum\nolimits_{j,t} (f_j^t(x_j^{t,\gamma_j^t}) - f_j^t(x_j^{t,0}))}\\
    & \quad + \frac{\sum_{j=1}^m{\left[ -f_j^{T_j}(x_j^{T_j,0}) - \sum_{t=1}^{T_j-1} \tau^{T_j-t}f_j^t(x_j^{t,0}) \right]}}{\sum\nolimits_{j,t} (f_j^t(x_j^{t,\gamma_j^t}) - f_j^t(x_j^{t,0}))} \\
    &= \frac{U(a) - \sum_{j=1}^m{\left[ f_j^{T_j}(x_j^{T_j,0}) + \sum_{t=1}^{T_j-1} \tau^{T_j-t}f_j^t(x_j^{t,0}) \right]}}{\sum\nolimits_{j,t} (f_j^t(x_j^{t,\gamma_j^t}) - f_j^t(x_j^{t,0}))}
\end{align}
Thus, the transformation in Proposition \ref{prop-1} can ensure classifications are not changed when the sub-marginal value functions are normalized.
\end{proof}

\section{A Dummy Example of Loss Function with Ordinal Thresholds}
\label{app-dummyexam}
\noindent Assume that in the context of an ordinal classification problem, the objective is to categorize alternatives into three classes: $Cl_1, Cl_2$, and $Cl_3$, with $Cl_3$ representing the most preferred category. The task involves learning ordinal thresholds during training to maximize the probability of alternatives remaining in their true classes.

Consider a hypothetical scenario with three alternatives: $a_1, a_2$, and $a_3$, assigned to $Cl_1, Cl_2$, and $Cl_3$ respectively. The computed comprehensive values for these alternatives are 0.5, 1.5, and 5. In Table \ref{tab:Thresholds}, various values of thresholds $\theta_1$ and $\theta_2$ are presented, along with the corresponding probabilities of each alternative being assigned to their respective classes and the resulting loss. For instance, when thresholds are set to $\theta_1=1$ and $\theta_2=4$, we can compute the probabilities for $a_1$ belonging to each class using Eq.~\eqref{eq:threshold}:
\begin{equation}
    \begin{aligned}
    P(Cl(a)=1)&=\Phi(1-0.5)-\Phi(-\infty -0.5)=0.6225-0=0.6225,\\
    P(Cl(a)=2)&=\Phi(4-0.5)-\Phi(1 -0.5)=0.9707-0.6225=0.3482, \\
    P(Cl(a)=3)&=\Phi(\infty-0.5)-\Phi(4 -0.5)=1-0.9707=0.0293. \\
\end{aligned}
\end{equation}
Given that the ground truth for $a_1$ is $Cl_1$, we can calculate the likelihood loss for $a_1$ using Eq.~\eqref{eq:loss}:
\begin{equation}
    \mathcal{L}=-\left(1 \cdot \ln(0.6225)+0\cdot\ln(0.3482)+0\cdot\ln(0.0293)\right)=0.4741.
\end{equation}
Subsequently, optimization techniques like stochastic gradient descent are employed to minimize the loss and update the training thresholds. When compared to other threshold combinations in Table~\ref{tab:Thresholds}, it is evident that $\theta_1=1$ and $\theta_2=4$ provide a more suitable configuration.

\begin{table}[ht]
    \centering\footnotesize
    \caption{The example for the computation of ordinal thresholds and loss function.     \label{tab:Thresholds}}
    \begin{tabular}{ccccccccc}
        \toprule
        $\theta_1$      & Alternative    & Assignment  & $U(a)$  & $P(Cl=1)$     & $P(Cl=2)$     & $P(Cl=3)$     & $\mathcal{L}_a$         \\ 
        1               & $a_1$     & 1             & 0.5           & 0.6225        & 0.3482        & 0.0293        & 0.4741               \\ 
        $\theta_2$      & $a_2$     & 2             & 1.5           & 0.3775        & 0.5466        & 0.0759        & 0.6040               \\ 
        4               & $a_3$     & 3             & 5             & 0.0180        & 0.2510        & 0.7311        & 0.3133               \\ 
        Average Loss ($\mathcal{L}$) & ~     & ~    & ~             & ~             & ~             & ~             & 0.4638               \\ \midrule
                
        $\theta_1$      & Alternative    & Assignment  & $U(a)$   & $P(Cl=1)$ & $P(Cl=2)$         & $P(Cl=3)$     & $\mathcal{L}_a$        \\ 
        0.1             & $a_1$     & 1             & 0.5           & 0.4013        & 0.5694        & 0.0293        & 0.9130               \\ 
        $\theta_2$      & $a_2$     & 2             & 1.5           & 0.1978        & 0.7263        & 0.0759        & 0.3198                \\ 
        4               & $a_3$     & 3             & 5             & 0.0074        & 0.2615        & 0.7311        & 0.3133                \\ 
        Average Loss ($\mathcal{L}$) & ~     & ~    & ~             & ~             & ~             & ~             & 0.5153                \\\midrule
        $\theta_1$      & Alternative    & Assignment  & $U(a)$   & $P(Cl=1)$     & $P(Cl=2)$     & $P(Cl=3)$     & $\mathcal{L}_a$       \\ 
        2               & $a_1$     & 1             & 0.5           & 0.8176        & 0.1531        & 0.0293        & 0.2014               \\ 
        $\theta_2$      & $a_2$     & 2             & 1.5           & 0.6225        & 0.3017        & 0.0759        & 1.1984               \\ 
        4               & $a_3$     & 3             & 5             & 0.0474        & 0.2215        & 0.7311        & 0.3133                \\ 
        Average Loss ($\mathcal{L}$) & ~     & ~    & ~             & ~             & ~             & ~             & 0.5710               \\\midrule
        $\theta_1$      & Alternative    & Assignment & $U(a)$   & $P(Cl=1)$     & $P(Cl=2)$     & $P(Cl=3)$     & $\mathcal{L}_a$       \\ 
        1               & $a_1$     & 1             & 0.5           & 0.6225        & 0.1951        & 0.1824        & 0.4741               \\ 
        $\theta_2$      & $a_2$     & 2             & 1.5           & 0.3775        & 0.2449        & 0.3775        & 1.4068               \\ 
        2               & $a_3$     & 3             & 5             & 0.0180        & 0.0294        & 0.9526        & 0.0486               \\ 
        Average Loss ($\mathcal{L}$) & ~     & ~    & ~             & ~             & ~             & ~             & 0.6432               \\\midrule
        $\theta_1$      & Alternative    & Assignment  & $U(a)$   & $P(Cl=1)$       & $P(Cl=2) $      & $P(Cl=3)$       & $\mathcal{L}_a$  \\ 
        1               & $a_1$     & 1             & 0.5           & 0.6225        & 0.3735        & 0.0041        & 0.4741                \\ 
        $\theta_2$      & $a_2$     & 2             & 1.5           & 0.3775        & 0.6115        & 0.0110        & 0.4919                \\ 
        6               & $a_3$     & 3             & 5             & 0.0180        & 0.7131        & 0.2689        & 1.3133                \\ 
        Average Loss ($\mathcal{L}$) & ~     & ~    & ~             & ~             & ~             & ~             & 0.7597                \\
        
        \bottomrule
    \end{tabular}
\end{table}

\section{Best Model Settings in Experiments}
\label{app-parameter}

\noindent We conduct ten-fold cross-validation in the real-case experiment and four simulations to determine the best parameters for the baselines and the proposed models. The best parameters are presented in Table \ref{tab-bestpara}.

\begin{table}[htbp]
    \centering
    \caption{Best parameters in real-case and simulation experiments.}
    \label{tab-bestpara}
    \resizebox{1\textwidth}{!}{
    \begin{tabular}{clllllllllll}
    \toprule
        Experiment type & TPL  & mRNN & \makecell[l]{UTADIS\\UTADIS-T}& XOFM & Logistic & SVM & RF & XGB & GRU & RNN & MLP  \\\midrule 
       Real-case  & \makecell[l]{$\gamma=4$\\$C=0.01$} & \makecell[l]{Dim $c=$40\\ \#para=3,162 } & $\gamma=4$ & \makecell[l]{$\gamma=3$\\$K=3$} & $C=0.01$ & $C=1$ & \makecell[l]{n\_est=100\\max\_feat=6\\max\_depth=10} & \makecell[l]{n\_est=60\\max\_depth=2} & \makecell[l]{Dim $c=$32\\ \#para=4,737 } & \makecell[l]{Dim $c=$40\\ \#para=3,521 } & \makecell[l]{96-32-8-1\\ \#para=3,377 } \\ 
       Basic  & \makecell[l]{$\gamma=4$\\$C=0.001$} &\makecell[l]{Dim $c=$40\\ \#para=3,162 } & $\gamma=4$ & \makecell[l]{$\gamma=3$\\$K=2$} & $C=0.1$ & $C=1$ & \makecell[l]{n\_est=80\\max\_feat=8\\max\_depth=10} & \makecell[l]{n\_est=80\\max\_depth=4} & \makecell[l]{Dim $c=$32\\ \#para=4,737 } & \makecell[l]{Dim $c=$40\\ \#para=3,521 } & \makecell[l]{80-32-8-1\\ \#para=2,865 } \\
       
       Non-Markovian  & \makecell[l]{$\gamma=3$\\$C=0.001$} & \makecell[l]{Dim $c=$40\\ \#para=3,162 } & $\gamma=3$ & \makecell[l]{$\gamma=3$\\$K=2$}  & $C=0.1$ & $C=1$ & \makecell[l]{n\_est=60\\max\_feat=8\\max\_depth=10} & \makecell[l]{n\_est=80\\max\_depth=2} & \makecell[l]{Dim $c=$32\\ \#para=4,737 } & \makecell[l]{Dim $c=$40\\ \#para=3,521 } & \makecell[l]{80-32-8-1\\ \#para=2,865 } \\
       Non-monotonic  & \makecell[l]{$\gamma=4$\\$C=0.01$} &\makecell[l]{Dim $c=$40\\ \#para=3,162 } & $\gamma=4$& \makecell[l]{$\gamma=3$\\$K=3$}  & $C=0.1$ & $C=1$ & \makecell[l]{n\_est=60\\max\_feat=10\\max\_depth=6} & \makecell[l]{n\_est=80\\max\_depth=2} & \makecell[l]{Dim $c=$32\\ \#para=4,737 } & \makecell[l]{Dim $c=$40\\ \#para=3,521 } & \makecell[l]{80-32-8-1\\ \#para=2,865 } \\
Non-independent  & \makecell[l]{$\gamma=4$\\$C=0.001$} &\makecell[l]{Dim $c=$40\\ \#para=3,162 } & $\gamma=3$& \makecell[l]{$\gamma=3$\\$K=4$}  & $C=1$ & $C=1$ & \makecell[l]{n\_est=80\\max\_feat=10\\max\_depth=8} & \makecell[l]{n\_est=80\\max\_depth=4} & \makecell[l]{Dim $c=$32\\ \#para=4,737 } & \makecell[l]{Dim $c=$40\\ \#para=3,521 } & \makecell[l]{80-32-8-1\\ \#para=2,865 } \\
       \bottomrule
    \end{tabular}
    }
\end{table}

\section{Completed Results for Simulation Experiments}
\label{app-compresult}

\noindent We have provided all experimental results in simulation in Tables \ref{app-sim1} to 
\ref{app-sim4}.

\begin{table}[h]
    \centering\footnotesize
    \small{
    \caption{Experimental results for Basic experiment across four metrics. Means and standard deviations are calculated via ten-fold cross-validation. Bold type highlights the top-performing results, while stars ($^*$) denote that the performance is significantly lower than the best ($^{***} p<0.01, ^{**} p<0.05, ^{*} p<0.1$).\label{app-sim1}}
    \begin{tabular}{cccccc}
            \toprule
        Method                 & Precision  & Recall & F-score & Accuracy  \\ \midrule
        SVM               & 0.909 $\pm$ 0.013$^{***}$           & 0.916 $\pm$ 0.015$^{***}$     & 0.913 $\pm$ 0.009$^{***}$     & 0.906 $\pm$ 0.008$^{***}$  \\ 
        Logistic          & 0.906 $\pm$ 0.010$^{***}$           & 0.905 $\pm$ 0.011$^{***}$     & 0.906 $\pm$ 0.008$^{***}$     & 0.899 $\pm$ 0.008$^{***}$  \\ 
        RF                & 0.892 $\pm$ 0.017$^{***}$           & 0.958 $\pm$ 0.013$^{**}$      & 0.924 $\pm$ 0.007$^{***}$     & 0.915 $\pm$ 0.009$^{***}$  \\ 
        XGB               & 0.920 $\pm$ 0.014$^{***}$           & 0.951 $\pm$ 0.012$^{***}$     & 0.935 $\pm$ 0.009$^{***}$     & 0.929 $\pm$ 0.010$^{***}$  \\ 
        MLP               & 0.910 $\pm$ 0.013$^{***}$           & 0.917 $\pm$ 0.010$^{***}$     & 0.914 $\pm$ 0.006$^{***}$     & 0.906 $\pm$ 0.006$^{***}$  \\ 
        RNN               & 0.922 $\pm$ 0.010$^{***}$           & 0.903 $\pm$ 0.011$^{***}$     & 0.912 $\pm$ 0.008$^{***}$     & 0.907 $\pm$ 0.008$^{***}$  \\ 
        GRU               & 0.917 $\pm$ 0.014$^{***}$           & 0.908 $\pm$ 0.010$^{***}$     & 0.912 $\pm$ 0.008$^{***}$     & 0.906 $\pm$ 0.009$^{***}$  \\ 
        UTADIS            & 0.889 $\pm$ 0.012$^{***}$           & 0.894 $\pm$ 0.023$^{***}$     & 0.891 $\pm$ 0.015$^{***}$     & 0.902 $\pm$ 0.015$^{***}$  \\ 
        UTADIS-T          & 0.886 $\pm$ 0.015$^{***}$           & 0.908 $\pm$ 0.022$^{***}$     & 0.897 $\pm$ 0.018$^{***}$     & 0.906 $\pm$ 0.011$^{***}$  \\     
        XOFM              & 0.893 $\pm$ 0.013$^{***}$           & 0.915 $\pm$ 0.021$^{***}$     & 0.904 $\pm$ 0.019$^{***}$     & 0.909 $\pm$ 0.009$^{***}$  \\ 
        TPL               & 0.877 $\pm$ 0.079$^{**}$            & 0.923 $\pm$ 0.073             & 0.893 $\pm$ 0.018$^{***}$     & 0.900 $\pm$ 0.023$^{***}$  \\ \midrule
        mRNN $(\gamma=2$) & 0.930 $\pm$ 0.013$^{***}$            & 0.958 $\pm$ 0.018$^{**}$     & 0.943 $\pm$ 0.011$^{***}$     & 0.938 $\pm$ 0.011$^{***}$  \\ 
        mRNN $(\gamma=4$) & 0.949 $\pm$ 0.013                   & 0.961 $\pm$ 0.022$^{*}$       & 0.952 $\pm$ 0.017$^{*}$       & 0.948 $\pm$ 0.020  \\ 
        mRNN $(\gamma=6$) & 0.949 $\pm$ 0.012                   & \textbf{0.973 $\pm$ 0.008}    & \textbf{0.961 $\pm$ 0.005}    & \textbf{0.957 $\pm$ 0.007}  \\ 
        mRNN $(\gamma=8$) & \textbf{0.949 $\pm$ 0.018}          & 0.969 $\pm$ 0.012             & 0.959 $\pm$ 0.008             & 0.955 $\pm$ 0.010  \\ 
                \bottomrule
    \end{tabular}}
\end{table}

\begin{table}[h]
    \centering\footnotesize
    \caption{Experimental results for Non-Markovian experiment across four metrics.}
    \small{\begin{tabular}{cccccc}    
            \toprule
        Method                & Precision  & Recall & F-score & Accuracy  \\ \midrule
        SVM               & 0.884 $\pm$ 0.010$^{***}$           & 0.864 $\pm$ 0.028$^{***}$     & 0.874 $\pm$ 0.016$^{***}$     & 0.888 $\pm$ 0.013$^{***}$  \\ 
        Logistic          & 0.889 $\pm$ 0.013$^{***}$           & 0.873 $\pm$ 0.017$^{***}$     & 0.881 $\pm$ 0.009$^{***}$     & 0.894 $\pm$ 0.008$^{***}$  \\ 
        RF                & 0.918 $\pm$ 0.018$^{***}$           & 0.856 $\pm$ 0.027$^{***}$     & 0.886 $\pm$ 0.019$^{***}$     & 0.901 $\pm$ 0.015$^{***}$  \\ 
        XGB               & 0.924 $\pm$ 0.014$^{***}$           & 0.896 $\pm$ 0.011$^{***}$     & 0.910 $\pm$ 0.007$^{***}$     & 0.920 $\pm$ 0.007$^{***}$  \\ 
        MLP               & 0.878 $\pm$ 0.014$^{***}$           & 0.877 $\pm$ 0.018$^{***}$     & 0.878 $\pm$ 0.009$^{***}$     & 0.890 $\pm$ 0.009$^{***}$  \\ 
        RNN               & 0.884 $\pm$ 0.016$^{***}$           & 0.880 $\pm$ 0.019$^{***}$     & 0.882 $\pm$ 0.013$^{***}$     & 0.894 $\pm$ 0.012$^{***}$  \\ 
        GRU               & 0.881 $\pm$ 0.011$^{***}$           & 0.888 $\pm$ 0.013$^{***}$     & 0.884 $\pm$ 0.009$^{***}$     & 0.896 $\pm$ 0.009$^{***}$  \\ 
        UTADIS            & 0.889 $\pm$ 0.014$^{***}$           & 0.896 $\pm$ 0.020$^{***}$     & 0.892 $\pm$ 0.011$^{***}$     & 0.903 $\pm$ 0.013$^{***}$  \\ 
        UTADIS-T          & 0.896 $\pm$ 0.015$^{***}$           & 0.904 $\pm$ 0.021$^{***}$     & 0.900 $\pm$ 0.016$^{***}$     & 0.906 $\pm$ 0.017$^{***}$  \\ 
        XOFM              & 0.905 $\pm$ 0.013$^{***}$           & 0.918 $\pm$ 0.014$^{***}$     & 0.911 $\pm$ 0.010$^{***}$     & 0.911 $\pm$ 0.015$^{***}$  \\ 
        TPL               & 0.919 $\pm$ 0.062$^{*}$             & 0.898 $\pm$ 0.078             & 0.909 $\pm$ 0.030$^{***}$     & 0.914 $\pm$ 0.025$^{***}$  \\ \midrule
        mRNN $(\gamma=2$) & 0.928 $\pm$ 0.014$^{***}$           & 0.912 $\pm$ 0.024$^{**}$      & 0.920 $\pm$ 0.013$^{***}$     & 0.929 $\pm$ 0.011$^{***}$  \\ 
        mRNN $(\gamma=4$) & 0.959 $\pm$ 0.014                   & 0.923 $\pm$ 0.025$^{**}$      & 0.940 $\pm$ 0.016$^{**}$      & 0.947 $\pm$ 0.016$^{**}$  \\ 
        mRNN $(\gamma=6$) & \textbf{0.965 $\pm$ 0.014}          & 0.931 $\pm$ 0.020             & 0.948 $\pm$ 0.012             & 0.954 $\pm$ 0.010  \\ 
        mRNN $(\gamma=8$) & 0.960 $\pm$ 0.013                   & \textbf{0.945 $\pm$ 0.012}    & \textbf{0.952 $\pm$ 0.006}    & \textbf{0.958 $\pm$ 0.006 } \\ 
                \bottomrule
    \end{tabular}}
\end{table}

\begin{table}[h]
    \centering\footnotesize
    \caption{Experimental results for Non-monotonic experiment across four metrics.}
        \small{
    \begin{tabular}{cccccc}
            \toprule
        Method                 & Precision  & Recall & F-score & Accuracy  \\ \midrule
        SVM               & 0.623 $\pm$ 0.034$^{***}$           & 0.559 $\pm$ 0.044$^{***}$     & 0.588 $\pm$ 0.028$^{***}$     & 0.645 $\pm$ 0.025$^{***}$  \\ 
        Logistic          & 0.652 $\pm$ 0.022$^{***}$           & 0.592 $\pm$ 0.035$^{***}$     & 0.620 $\pm$ 0.024$^{***}$     & 0.671 $\pm$ 0.018$^{***}$  \\ 
        RF                & 0.672 $\pm$ 0.027                   & 0.582 $\pm$ 0.035$^{***}$     & 0.623 $\pm$ 0.027$^{***}$     & 0.681 $\pm$ 0.022$^{**}$  \\ 
        XGB               & \textbf{0.683 $\pm$ 0.030}          & \textbf{0.642 $\pm$ 0.032}    & \textbf{0.661 $\pm$ 0.024}    & \textbf{0.702 $\pm$ 0.017}  \\ 
        MLP               & 0.647 $\pm$ 0.024$^{***}$           & 0.574 $\pm$ 0.034$^{***}$     & 0.608 $\pm$ 0.021$^{***}$     & 0.664 $\pm$ 0.017$^{***}$  \\ 
        RNN               & 0.665 $\pm$ 0.037                   & 0.565 $\pm$ 0.033$^{***}$     & 0.610 $\pm$ 0.024$^{***}$     & 0.672 $\pm$ 0.019$^{***}$  \\ 
        GRU               & 0.662 $\pm$ 0.034$^{*}$             & 0.551 $\pm$ 0.031$^{***}$     & 0.600 $\pm$ 0.019$^{***}$     & 0.667 $\pm$ 0.017$^{***}$  \\ 
        UTADIS            & 0.558 $\pm$ 0.029$^{***}$           & 0.616 $\pm$ 0.038             & 0.585 $\pm$ 0.025$^{***}$     & 0.603 $\pm$ 0.021$^{***}$  \\ 
        UTADIS-T          & 0.577 $\pm$ 0.033$^{***}$           & 0.624 $\pm$ 0.039             & 0.600 $\pm$ 0.029$^{***}$     & 0.632 $\pm$ 0.019$^{***}$  \\ 
        XOFM              & 0.583 $\pm$ 0.029$^{***}$           & 0.619 $\pm$ 0.045             & 0.600 $\pm$ 0.027$^{***}$     & 0.637 $\pm$ 0.016$^{***}$  \\     
        TPL               & 0.642 $\pm$ 0.066                   & 0.607 $\pm$ 0.147             & 0.606 $\pm$ 0.048$^{**}$     & 0.652 $\pm$ 0.033$^{***}$  \\ \midrule
        mRNN $(\gamma=2$) & 0.644 $\pm$ 0.030$^{***}$           & 0.625 $\pm$ 0.043             & 0.634 $\pm$ 0.029$^{**}$      & 0.672 $\pm$ 0.021$^{***}$  \\ 
        mRNN $(\gamma=4$) & 0.649 $\pm$ 0.026$^{***}$           & 0.624 $\pm$ 0.039             & 0.636 $\pm$ 0.025$^{**}$      & 0.675 $\pm$ 0.022$^{**}$  \\ 
        mRNN $(\gamma=6$) & 0.658 $\pm$ 0.034$^{**}$            & 0.623 $\pm$ 0.044             & 0.638 $\pm$ 0.024$^{**}$      & 0.680 $\pm$ 0.020$^{**}$  \\ 
        mRNN $(\gamma=8$) & 0.660 $\pm$ 0.033$^{**}$            & 0.596 $\pm$ 0.055$^{**}$      & 0.624 $\pm$ 0.031$^{***}$     & 0.675 $\pm$ 0.021$^{***}$  \\ 
                \bottomrule
    \end{tabular}}
\end{table}

\begin{table}[h]
    \centering\footnotesize
    \caption{Experimental results for Non-independent experiment across four metrics. \label{app-sim4}}
    \small{
    \begin{tabular}{cccccc}
            \toprule
        Method                & Precision  & Recall & F-score & Accuracy  \\ \midrule
        SVM               & 0.911 $\pm$ 0.014$^{***}$           & 0.888 $\pm$ 0.026$^{***}$     & 0.899 $\pm$ 0.016$^{***}$     & 0.922 $\pm$ 0.014$^{***}$  \\ 
        Logistic          & 0.932 $\pm$ 0.009$^{***}$           & 0.917 $\pm$ 0.016$^{**}$      & 0.925 $\pm$ 0.009$^{***}$     & 0.941 $\pm$ 0.008$^{***}$  \\ 
        RF                & 0.917 $\pm$ 0.021$^{***}$           & 0.774 $\pm$ 0.026$^{***}$     & 0.839 $\pm$ 0.017$^{***}$     & 0.883 $\pm$ 0.014$^{***}$  \\ 
        XGB               & 0.905 $\pm$ 0.013$^{***}$           & 0.871 $\pm$ 0.024$^{***}$     & 0.887 $\pm$ 0.011$^{***}$     & 0.913 $\pm$ 0.009$^{***}$  \\ 
        MLP               & 0.924 $\pm$ 0.017$^{***}$           & 0.916 $\pm$ 0.022$^{**}$      & 0.920 $\pm$ 0.012$^{***}$     & 0.937 $\pm$ 0.011$^{***}$  \\ 
        RNN               & 0.964 $\pm$ 0.006                   & \textbf{0.942 $\pm$ 0.015}    & \textbf{0.953 $\pm$ 0.008}    & \textbf{0.963 $\pm$ 0.006}  \\ 
        GRU               & \textbf{0.964 $\pm$ 0.010}          & 0.940 $\pm$ 0.015             & 0.952 $\pm$ 0.009             & 0.963 $\pm$ 0.007  \\ 
        UTADIS            & 0.850 $\pm$ 0.016$^{***}$           & 0.866 $\pm$ 0.029$^{***}$     & 0.857 $\pm$ 0.013$^{***}$     & 0.887 $\pm$ 0.012$^{***}$  \\ 
        UTADIS-T          & 0.862 $\pm$ 0.019$^{***}$           & 0.869 $\pm$ 0.025$^{***}$     & 0.865 $\pm$ 0.011$^{***}$     & 0.894 $\pm$ 0.010$^{***}$  \\
        XOFM              & 0.874 $\pm$ 0.023$^{***}$           & 0.877 $\pm$ 0.031$^{***}$     & 0.875 $\pm$ 0.021$^{***}$     & 0.902 $\pm$ 0.011$^{***}$  \\
        TPL               & 0.878 $\pm$ 0.067$^{**}$            & 0.910 $\pm$ 0.070             & 0.894 $\pm$ 0.020$^{***}$     & 0.910 $\pm$ 0.017$^{***}$  \\ \midrule
        mRNN $(\gamma=2$) & 0.950 $\pm$ 0.021$^{*}$             & 0.868 $\pm$ 0.026$^{***}$     & 0.906 $\pm$ 0.011$^{***}$     & 0.929 $\pm$ 0.008$^{***}$  \\ 
        mRNN $(\gamma=4$) & 0.932 $\pm$ 0.023$^{***}$           & 0.872 $\pm$ 0.030$^{***}$     & 0.901 $\pm$ 0.014$^{***}$     & 0.924 $\pm$ 0.012$^{***}$  \\ 
        mRNN $(\gamma=6$) & 0.935 $\pm$ 0.021$^{***}$           & 0.871 $\pm$ 0.026$^{***}$     & 0.902 $\pm$ 0.015$^{***}$     & 0.925 $\pm$ 0.011$^{***}$  \\ 
        mRNN $(\gamma=8$) & 0.933 $\pm$ 0.021$^{***}$           & 0.868 $\pm$ 0.025$^{***}$     & 0.899 $\pm$ 0.016$^{***}$     & 0.923 $\pm$ 0.012$^{***}$  \\ 
                \bottomrule
    \end{tabular}}
\end{table}

\section{Example of Computing Comprehensive Value}
\label{app-compglobalvalue}
\begin{table}[ht]
    \centering\footnotesize
    \caption{The example for using sub-marginal values and time discount factors to aggregate marginal and comprehensive values. Note that the last day's time discount factors are not used in practice.\label{app-exampleofcomputingglobal}}
    \resizebox{1\textwidth}{!}{
    \begin{tabular}{ccccccccccccccccc}
    \toprule
        Day & $g_1$ & $g_2$ & $g_3$ & $g_4$ & $\tau_1$ & $\tau_2$ & $\tau_3$ & $\tau_4$ & $f_1$ & $f_2$ & $f_3$ & $f_4$ & $u_1$ & $u_2$ & $u_3$ & $u_4$  \\ \hline 
        1 & 861.36 & 2 & 15.62 & 8 & 0.10 & 0.55 & 0.60 & 0.87 & 1.99 & 0.54 & 0.10 & 0.13 & 1.99 & 0.54 & 0.10 & 0.38  \\ 
        2 & 532.00 & 4 & 12.20 & 6 & 0.33 & 0.77 & 0.73 & 0.95 & 0.23 & 1.02 & 0.21 & 0.08 & 0.43 & 1.32 & 0.27 & 0.40  \\ 
        3 & 0.00 & 0 & 4.20 & 4 & 0.46 & 0.90 & 0.80 & 0.95 & 0.00 & 0.00 & 0.06 & 0.05 & 0.14 & 1.02 & 0.26 & 0.43  \\ 
        4 & 6.00 & 1 & 6.39 & 5 & 0.45 & 0.86 & 0.83 & 0.96 & 0.00 & 0.92 & 0.15 & 0.04 & 0.06 & 1.84 & 0.35 & 0.45  \\ 
        5 & 0.00 & 0 & 9.03 & 7 & 0.44 & 0.89 & 0.85 & 0.97 & 0.00 & 0.00 & 0.16 & 0.10 & 0.03 & 1.58 & 0.45 & 0.53  \\ 
        6 & 0.00 & 0 & 3.12 & 4 & 0.43 & 0.86 & 0.83 & 0.97 & 0.00 & 0.00 & 0.04 & 0.09 & 0.01 & 1.39 & 0.42 & 0.60  \\ 
        7 & 2642.00 & 17 & 9.14 & 4 & 0.50 & 0.94 & 0.83 & 0.96 & 1.93 & 1.38 & 0.00 & 0.12 & 1.94 & 2.58 & 0.35 & 0.70  \\ 
        8 & 0.00 & 0 & 5.90 & 6 & 0.74 & 0.97 & 0.81 & 0.96 & 0.00 & 0.06 & 0.00 & 0.22 & 0.96 & 2.48 & 0.29 & 0.88  \\ 
        9 & 0.00 & 0 & 10.85 & 4 & 0.75 & 0.91 & 0.84 & 0.96 & 0.00 & 0.00 & 0.29 & 0.14 & 0.71 & 2.42 & 0.53 & 0.99  \\ 
        10 & 0.00 & 0 & 11.39 & 4 & 0.75 & 0.84 & 0.88 & 0.95 & 0.00 & 0.00 & 0.40 & 0.22 & 0.54 & 2.20 & 0.84 & 1.17  \\ 
        11 & 80.00 & 3 & 5.56 & 3 & 0.80 & 0.91 & 0.89 & 0.94 & 0.00 & 0.00 & 0.41 & 0.23 & 0.40 & 1.84 & 1.15 & 1.34  \\ 
        12 & 0.00 & 0 & 10.97 & 6 & 0.83 & 0.96 & 0.91 & 0.95 & 0.00 & 0.00 & 1.05 & 0.39 & 0.32 & 1.67 & 2.07 & 1.66  \\ 
        13 & 68.00 & 1 & 10.54 & 4 & 0.86 & 0.91 & 0.92 & 0.96 & 0.00 & 0.12 & 1.38 & 0.41 & 0.26 & 1.73 & 3.27 & 2.00  \\ 
        14 & 80.00 & 3 & 8.29 & 6 & 0.86 & 0.93 & 0.93 & 0.96 & 0.00 & 0.15 & 1.16 & 0.29 & 0.23 & 1.72 & 4.18 & 2.21  \\ 
        15 & 19.00 & 4 & 3.22 & 4 & 0.89 & 0.96 & 0.90 & 0.95 & 0.00 & 0.08 & 0.92 & 0.38 & 0.20 & 1.68 & 4.79 & 2.50  \\ 
        16 & 99.00 & 8 & 8.59 & 5 & 0.90 & 0.97 & 0.89 & 0.95 & 0.01 & 0.15 & 1.01 & 0.26 & 0.18 & 1.78 & 5.34 & 2.63  \\ 
        17 & 1519.50 & 5 & 6.17 & 4 & 0.96 & 0.98 & 0.87 & 0.95 & 1.13 & 0.15 & 0.68 & 0.31 & 1.29 & 1.87 & 5.44 & 2.81  \\ 
        18 & 198.00 & 1 & 7.89 & 5 & 0.96 & 0.97 & 0.84 & 0.95 & 0.00 & 0.15 & 0.59 & 0.31 & 1.25 & 1.98 & 5.30 & 2.98  \\ 
        19 & 0.00 & 0 & 11.26 & 6 & 0.98 & 0.96 & 0.81 & 0.96 & 0.00 & 0.09 & 0.42 & 0.15 & 1.20 & 2.01 & 4.84 & 2.99  \\ 
        20 & 0.00 & 0 & 6.97 & 4 & 0.96 & 0.92 & 0.77 & 0.96 & 0.00 & 0.04 & 0.18 & 0.11 & 1.17 & 1.96 & 4.12 & 2.99  \\ 
        21 & 75.00 & 3 & 4.49 & 5 & 0.90 & 0.91 & 0.70 & 0.96 & 0.00 & 0.19 & 0.01 & 0.11 & 1.12 & 2.01 & 3.18 & 2.98  \\ 
        22 & 94.00 & 2 & 11.21 & 5 & 0.97 & 0.94 & 0.71 & 0.96 & 0.00 & 0.32 & 0.17 & 0.13 & 1.01 & 2.14 & 2.42 & 2.98  \\ 
        23 & 0.00 & 0 & 14.81 & 6 & 0.93 & 0.95 & 0.77 & 0.96 & 0.00 & 0.08 & 0.20 & 0.11 & 0.98 & 2.10 & 1.91 & 2.98  \\ 
        24 & 74.00 & 2 & 7.74 & 4 & - & - & - & - & 0.00 & 0.58 & 0.28 & 0.17 & 0.91 & 2.57 & 1.76 & 3.04  \\ \hline
        Comprehensive value &  (Threshold) & ~ & ~ & ~ & ~ & ~ & ~ & ~ & ~ & ~ & ~ & ~ & ~ & ~ & 8.28 & (1.64)  \\ 
            \bottomrule
    \end{tabular}
    }
\end{table}

\noindent In Table \ref{app-exampleofcomputingglobal}, we delve into an illustrative example extracted from the user value evaluation experiment to explicate the computation of three pivotal functions: the sub-marginal value function, the marginal value function, and the comprehensive value function.

Commencing with the sub-marginal value vector, we employ Eq.~\eqref{eq:sub-mgn} in conjunction with an mRNN. On the initial day (Day 1), the sub-marginal values corresponding to the four criteria stand at 1.99, 0.54, 0.10, and 0.13, respectively. Consequently, these values serve as the marginal values for each criterion on Day 1.

Progressing to Day 2, we posit that the marginal value vector is a composite of two components: the sub-marginal value vector for Day 2 and the discounted marginal value vector from the preceding timestamp (Day 1). The calculation of the discount factor for each of the four criteria is determined by Eq.~\eqref{eq:discount}, yielding factors of 0.10, 0.55, 0.60, and 0.87. Armed with these factors, the computation of the marginal value vector for Day 2 unfolds as follows:
\begin{equation}
    \begin{aligned}
    u_1^2 &= f_1^2 + \tau_1^1 \cdot u_1^1=0.23+0.10 \cdot 1.99=0.43, \\
    u_2^2 &= f_2^2 + \tau_2^1 \cdot u_2^1=1.02+0.55 \cdot 0.54=1.32, \\
    u_3^2 &= f_3^2 + \tau_3^1 \cdot u_3^1=0.21+0.60 \cdot 0.10=0.27, \\
    u_4^2 &= f_4^2 + \tau_4^1 \cdot u_4^1=0.08+0.87 \cdot 0.13=0.40. \\
\end{aligned}
\end{equation}
This iterative process extends until the final timestamp, culminating in a marginal value vector of 0.91, 2.57, 1.76, and 3.04 for the respective criteria. Subsequently, these individual marginal values are aggregated to compute the comprehensive marginal value, resulting in a total of 8.28. Note that we can normalize the obtained sub-marginal value functions to re-scale the comprehensive values into range $[0,1]$, following the Eq.~\eqref{eq-normalization}, which is required in some decision scenarios. To render a classification determination, we ascertain the optimized threshold ($\theta=1.64$) through Eqs.~\eqref{eq:threshold} and~\eqref{eq:loss}. In this specific scenario, given that the calculated marginal value of 8.28 surpasses the threshold of 1.64, the alternative is classified into the high-value category.















\end{document}






\TITLE{Data-driven Preference Learning Methods for Multiple Criteria Sorting with Temporal Criteria}

\ARTICLEAUTHORS{%
\AUTHOR{Yijun LI\thanks{Both authors contributed equally to this work.}}
\AFF{School of Data Science, City University of Hong Kong, \EMAIL{yijunli5-c@my.cityu.edu.hk} \URL{}}
\AUTHOR{Mengzhuo GUO\footnotemark[1]$^,$\thanks{Corresponding author}}
\AFF{Business School, Sichuan University, \EMAIL{mengzhguo@scu.edu.cn} \URL{}}
\AUTHOR{Qingpeng ZHANG}
\AFF{Musketeers Foundation Institute of Data Science and LKS Faculty of Medicine, The University of Hong Kong, \EMAIL{qpzhang@hku.hk} \URL{}}

} 

\ABSTRACT{%
This is the supplementary material of the study entitled \textit{Data-driven Preference Learning Methods for Multiple Criteria Sorting with Temporal Criteria}. 
}%


\KEYWORDS{Preference learning; Decision analysis; Sorting; Additive value function; Deep learning; Recurrent neural networks; Temporal criteria}
\HISTORY{}

\maketitle
%
\begin{APPENDICES}

\section{Proof of the Transformation from $P0$ to $P1$}
\label{app-proof}
\proof{Proof}
The Lagrange function of $P0$ is:
\begin{equation}
    \mathcal{L}(\mathbf{u}, \boldsymbol{\xi}, \boldsymbol{\mu}, \boldsymbol{\eta}, \boldsymbol{\sigma}) = \frac{1}{2} \left\Vert \mathbf{u}\right\Vert ^2_2 + C\cdot\sum_{i=1}^N \xi_i - \sum_{i=1}^N  \mu_i \left[y_i \mathbf{u}^T \mathbf{x}_i + \xi_i -1 \right] - \sum_{i=1}^N \eta_i \xi_i - \sum_{j=1}^m\sum_{t=1}^{T_j}\sum_{k=1}^{\gamma_j^t}\Delta f_j^{t,k}\cdot\sigma_j^{t,k}
\end{equation}
where $\boldsymbol{\xi}=(\xi_1,\ldots,\xi_N)^T, \boldsymbol{\mu}=(\mu_1,\ldots,\mu_N)^T, \boldsymbol{\eta}=(\eta_1, \ldots,\eta_N)^T, \boldsymbol{\sigma}=(\underbrace{\sigma_1^{1,1},\ldots,\sigma_1^{1,\gamma_1^1}}_{\text{1st timestamp on criterion } g_1  } \ldots, \sigma_m^{T_m,\gamma_m^{T_m}} )$. We set derivatives of $\mathcal{L}(\mathbf{u}, \boldsymbol{\xi}, \boldsymbol{\mu}, \boldsymbol{\eta}, \boldsymbol{\sigma})$ w.r.t. $\mathbf{u}$ and $\boldsymbol{\xi}$ to be zero, respectively:
\begin{align}
    &\frac{\partial \mathcal{L}(\mathbf{u}, \boldsymbol{\xi}, \boldsymbol{\mu}, \boldsymbol{\eta}, \boldsymbol{\sigma})}{\partial \Delta f_j^{t,k}} = \Delta f_j^{t,k} -\sum_{i=1}^N y_i \mu_i X_{i,j}^{t,k} - \sigma_j^{t,k} =0 \\
    &\Rightarrow \Delta f_j^{t,k} = \sum_{i=1}^N y_i \mu_i X_{i,j}^{t,k} + \sigma_j^{t,k} \label{eq-der1} \\
    & \frac{\partial \mathcal{L}(\mathbf{u}, \boldsymbol{\xi}, \boldsymbol{\mu}, \boldsymbol{\eta}, \boldsymbol{\sigma})}{\partial \xi_i} = C - \mu_i - \eta_i = 0 \\
    & \Rightarrow C\cdot \sum_{i=1}^N \xi_i= \sum_{i=1}^N \mu_i\xi_i - \sum_{i=1}^N \eta_i\xi_i \label{eq-der2}
\end{align}
Given Eq.(\ref{eq-der2}), the Lagrange function can be simplified:
\begin{equation}
    \mathcal{L}(\mathbf{u},\boldsymbol{\mu}, \boldsymbol{\sigma}) =  \frac{1}{2} \left\Vert \mathbf{u}\right\Vert ^2_2 - \sum_{i=1}^N \mu_i \left(y_i \mathbf{u}^T \mathbf{x}_i -1 \right) - \sum_{j=1}^m\sum_{t=1}^{T_j}\sum_{k=1}^{\gamma_j^t}\Delta f_j^{t,k}\cdot\sigma_j^{t,k} \label{eq-lagfunc2}
\end{equation}
Given Eq.(\ref{eq-der1}), the last term in Eq.(\ref{eq-lagfunc2}) can be further expressed as:
\begin{align}
    \sum_{j=1}^m\sum_{t=1}^{T_j}\sum_{k=1}^{\gamma_j^t}\Delta f_j^{t,k}\cdot\sigma_j^{t,k} &= \sum_{j=1}^m\sum_{t=1}^{T_j}\sum_{k=1}^{\gamma_j^t} \Delta f_j^{t,k}\left(\Delta f_j^{t,k} - \sum_{i=1}^N y_i \mu_i X_{i,j}^{t,k}  \right) \\
    & = \sum_{j=1}^m\sum_{t=1}^{T_j}\sum_{k=1}^{\gamma_j^t} (\Delta f_j^{t,k})^2 - \sum_{j=1}^m\sum_{t=1}^{T_j}\sum_{k=1}^{\gamma_j^t} \Delta f_j^{t,k} (\sum_{i=1}^N y_i \mu_i X_{i,j}^{t,k}) \\ 
    & = \left\Vert \mathbf{u}\right\Vert ^2_2 - \sum_{i=1}^N y_i\mu_i \left( \sum_{j=1}^m\sum_{t=1}^{T_j}\sum_{k=1}^{\gamma_j^t} \Delta f_j^{t,k} X_{i,j}^{t,k}  \right) \\
    & = \left\Vert \mathbf{u}\right\Vert ^2_2 - \sum_{i=1}^N y_i \mu_i \mathbf{u}^T \mathbf{x}_i \label{eq-lagfunc3}
\end{align}
Hence, Eq.(\ref{eq-lagfunc2}) can be rewritten as:
\begin{equation}
     \mathcal{L}(\boldsymbol{\mu}, \boldsymbol{\sigma}) = \sum_{i=1}^N\mu_i - \frac{1}{2} \sum_{j=1}^m\sum_{t=1}^{T_j}\sum_{k=1}^{\gamma_j^t} \left( \sum_{i=1}^N y_i \mu_i  X_{i,j}^{t,k} + \sigma_j^{t,k} \right)
\end{equation}
The dual problem of P0 can be:
\begin{align}
    \max g(\boldsymbol{\mu}, \boldsymbol{\eta}, \boldsymbol{\sigma}) &= \inf_{\mathbf{u}, \boldsymbol{\xi}} \mathcal{L}(\mathbf{u}, \boldsymbol{\xi}, \boldsymbol{\mu}, \boldsymbol{\eta}, \boldsymbol{\sigma}) \\
    & = \sum_{i=1}^N \mu_i - \frac{1}{2} \sum_{j=1}^m\sum_{t=1}^{T_j}\sum_{k=1}^{\gamma_j^t} \left( \sum_{i=1}^N y_i \mu_i  X_{i,j}^{t,k} + \sigma_j^{t,k} \right) \\
    s.t.: \quad & \boldsymbol{\sigma} \ge {\mathbf{0}} \\
    & C\mathbf{e} \ge \boldsymbol{\mu} \ge \mathbf{0}
\end{align}
From Eq.(\ref{eq-der1}), we can replace $ \sigma_j^{t,k} $ by $ \Delta f_j^{t,k} - \sum_{i=1}^N y_i \mu_i X_{i,j}^{t,k} $ and obtain P1:
\begin{align}
    \text{(P1)} \quad & \min \frac{1}{2} \left\Vert {\mathbf{u}} \right\Vert^2_2 - \sum\nolimits_{i=1}^N \mu_i \nonumber \\
    s.t.:\quad & \Delta f_j^{t,k} - \sum\nolimits_i^N y_i \mu_i X_{i,j}^{t,k} \ge 0, j=1,\ldots,m, t=1,\ldots, T_j, k = 1,\ldots,\gamma_j^{t} \\
    & C\ge \mu_i  \ge 0, i=1,\ldots,N 
\end{align}
\endproof

\section{Proof of Proposition \ref{prop-1}}
\label{app-prop1}

\begin{proposition}
    If an alternative $a$ is assigned to $Cl_h$, i.e., $\theta_{h-1}\le U(a) < \theta_h$, then the transformed threshold should satisfy $\theta_h'=\frac{\theta_h - \sum_{j=1}^m{\left[ f_j^{T_j}(x_j^{T_j,0}) + \sum_{t=1}^{T_j-1} \tau^{T_j-t}f_j^t(x_j^{t,0}) \right]}}{\sum\nolimits_{j,t} (f_j^t(x_j^{t,\gamma_j^t}) - f_j^t(x_j^{t,0})) }$ to ensure $\theta_{h-1}'\le U'(a) < \theta_h'$, where $U'(\cdot)$ is the global value using the transformed sub-marginal value functions. \label{prop-1}
\end{proposition}

\proof{Proof}
    Recall the definition of the sub-marginal values of an alternative $a$ in $f_j^{t}(g_j^{t}(a))=\sum_{k=1}^{k_j} \Delta f_j^{t,k} + \frac{g^t_j(a)-x^{t, k_j}_j}{x^{t, k_j+1}_j-x^{t, k_j}_j} \Delta f_j^{t,k_j+1} \quad if \quad g_j^t(a)\in \left[x_j^{t,k_j},x_j^{t,k_j+1} \right]$, we can rewrite the sub-marginal value at $t$-th timestamp in $j$-th criterion:
    \begin{align}
        f_j^t(g_j^t(a)) = f_j^t(x_j^{t,k}) + \frac{g_j^t(a) - x_j^{t,k}}{x_j^{t,k+1}-x_j^{t,k}}\left( f_j^t(x_j^{t,k+1}) - f_j^t(x_j^{t,k}) \right) 
    \end{align}
Following the transformation in Eq.(\ref{eq-normalization}):
\begin{equation}
    \begin{aligned}
     f_j^t(x_j^{t,k_j})' &= \frac{f_j^t(x_j^{t,\gamma_j^t}) - f_j^t(x_j^{t,0})}{\sum\nolimits_{j,t} (f_j^t(x_j^{t,\gamma_j^t}) - f_j^t(x_j^{t,0})) } \times \frac{f_j^t(x_j^{t,k_j}) - f_j^t(x_j^{t,0})}{f_j^t(x_j^{t,\gamma_j^t}) - f_j^t(x_j^{t,0})} \\
     &=\frac{f_j^t(x_j^{t,k_j}) - f_j^t(x_j^{t,0})}{\sum\nolimits_{j,t} (f_j^t(x_j^{t,\gamma_j^t}) - f_j^t(x_j^{t,0})) }.
\end{aligned}\label{eq-normalization}
\end{equation}
We can easily get 
\begin{align}
    f_j^t(g_j^t(a))' &= \frac{f_j^t(x_j^{t,k})-f_j^t(x_j^{t,0})}{\sum\nolimits_{j,t} (f_j^t(x_j^{t,\gamma_j^t}) - f_j^t(x_j^{t,0})) } + \frac{g_j^t(a) - x_j^{t,k}}{x_j^{t,k+1}-x_j^{t,k}}\left[ \frac{f_j^t(x_j^{t,k+1})-f_j^t(x_j^{t,0})}{\sum\nolimits_{j,t} (f_j^t(x_j^{t,\gamma_j^t}) - f_j^t(x_j^{t,0})) } - \frac{f_j^t(x_j^{t,k})-f_j^t(x_j^{t,0})}{\sum\nolimits_{j,t} (f_j^t(x_j^{t,\gamma_j^t}) - f_j^t(x_j^{t,0})) } \right] \\
    &= \frac{f_j^t(x_j^{t,k}) + \frac{g_j^t(a) - x_j^{t,k}}{x_j^{t,k+1}-x_j^{t,k}}\left( f_j^t(x_j^{t,k+1}) - f_j^t(x_j^{t,k}) \right) - f_j^t(x_j^{t,0})}{\sum\nolimits_{j,t} (f_j^t(x_j^{t,\gamma_j^t}) - f_j^t(x_j^{t,0})) } \\
    &= \frac{f_j^t(g_j^t(a)) - f_j^t(x_j^{t,0})}{\sum\nolimits_{j,t} (f_j^t(x_j^{t,\gamma_j^t}) - f_j^t(x_j^{t,0})) } \label{eq-transmargin}
\end{align}
Using Eq.(\ref{eq-transmargin}) to express the transformed global value $U'(a)$, we can get:
\begin{align}
    U'(a) &= \frac{\sum_{j=1}^m \left[ f_j^{T_j}(g_j^{T_j}(a)) - f_j^{T_j}(x_j^{T_j,0}) \right]}{\sum\nolimits_{j,t} (f_j^t(x_j^{t,\gamma_j^t}) - f_j^t(x_j^{t,0})) } + \frac{\sum_{t=1}^{T_j-1} \tau^{T_j-t}\left[ f_j^t(g_j^t(a)) - f_j^t(x_j^{t,0}) \right] }{\sum\nolimits_{j,t} (f_j^t(x_j^{t,\gamma_j^t}) - f_j^t(x_j^{t,0})) }\\
    &= \frac{\sum_{j+1}^{m} \left[ f_j^{T_j}(g_j^{T_j}(a)) + \sum_{t=1}^{T_j-1}\tau^{T_j-t}f_j^t(g_j^t(a)) \right] }{\sum\nolimits_{j,t} (f_j^t(x_j^{t,\gamma_j^t}) - f_j^t(x_j^{t,0}))}\\
    & \quad + \frac{\sum_{j=1}^m{\left[ -f_j^{T_j}(x_j^{T_j,0}) - \sum_{t=1}^{T_j-1} \tau^{T_j-t}f_j^t(x_j^{t,0}) \right]}}{\sum\nolimits_{j,t} (f_j^t(x_j^{t,\gamma_j^t}) - f_j^t(x_j^{t,0}))} \\
    &= \frac{U(a) - \sum_{j=1}^m{\left[ f_j^{T_j}(x_j^{T_j,0}) + \sum_{t=1}^{T_j-1} \tau^{T_j-t}f_j^t(x_j^{t,0}) \right]}}{\sum\nolimits_{j,t} (f_j^t(x_j^{t,\gamma_j^t}) - f_j^t(x_j^{t,0}))}
\end{align}
Thus, the transformation in Proposition \ref{prop-1} can ensure classifications are not changed when the sub-marginal value functions are normalized.
\endproof

\section{Baseline Models and Their Settings in Experiments}
\label{app-parameter}

The baseline models' descriptions are listed as follows:
\begin{itemize}
    \item Single temporal preference learning model (Stpl). This model corresponds to the optimization-based approach introduced in problem statement $P1$. It optimizes the model parameters using all available data simultaneously. The optimization problem is solved using the CVXOPT Python package with the Mosek optimizer.
    
    \item Parallel temporal preference learning model (Ptpl). Ptpl is an extension of the Stpl model and is implemented using Algorithm 1. It leverages a parallel learning approach to optimize model parameters efficiently.
    
    \item The UTilit\'es Additives DIScriminantes (UTADIS) method \citep{zopounidis2002multicriteria}. It is a classic approach for MCS problems. It utilizes linear programming to minimize misclassification errors, defined as the average distance of global values and thresholds. In this study, each timestamp is treated as an individual criterion in UTADIS, and a threshold-based preference model in the form of piecewise linear functions with equal segments is inferred using the Pulp Python package.
    
    \item Logistic regression (LR). LR is a linear model aggregating input attributes (criteria in this study) using weighted sums. It transforms the output using a function determining the probability of assigning an alternative to a desired class \citep{hosmer2013applied}. The hyper-parameter $C$ controls the degree of regularization applied to the model.
    
    \item Support vector machines (SVM) \citep{cortes1995support}. SVM is a supervised learning algorithm that maps training examples to points in space to maximize the margin between two classes. This study uses the criteria values for all timestamps as inputs to SVM. A Radial Basis Function kernel is employed to learn a non-linear classifier. The hyper-parameter $C$ controls the trade-off between maximizing the margin and minimizing classification errors on the training data.

    \item Random forests (RF) \citep{ho1995random}. RF is an ensemble learning method that aggregates multiple decision trees during training. It is often used as a black-box model due to its ability to provide predictions across a wide range of data with minimal configuration. Hyper-parameters such as n\_estimators, max\_features, and max\_depth play critical roles in controlling the behavior and performance of the ensemble of decision trees in RF.
        
    \item Extreme gradient boosting (XGB) \citep{chen2016xgboost}. XGB is a sequential model based on a gradient-boosting algorithm. It differs from bagging algorithms and can be parallelized. XGB incorporates regularization techniques to generate smaller trees, mitigating overfitting issues. Hyper-parameters like n\_estimators and max\_depth are crucial for controlling the behavior and performance of the ensemble of decision trees in XGB.
    
    \item Monotonic recurrent neural network (mRNN). The mRNN is a proposed deep preference learning model designed for temporal criteria with learnable time discount factors. Several adaptations are made to ensure the criteria monotonicity, preference independence, and natural order between classes assumptions are satisfied. This model captures the temporal dynamics in the criteria while maintaining these critical properties.
    
    \item Multi-layer perceptron (MLP) \citep{rosenblatt1958perceptron}. The MLP is a fully connected ANN. It can capture nonlinear transformations and interactions between criteria, making it suitable for distinguishing data that is not linearly separable. This study employs a three-layered MLP as a deep learning-based baseline, with criteria values from all timestamps directly used as input.
    
    \item Recurrent neural network (RNN) \citep{schuster1997bidirectional}. RNN is a type of sequential ANN that considers internal states to process sequential data. The output from the current state can influence subsequent inputs. The proposed mRNN is adapted from the standard RNN architecture to better describe and address MCS problems with temporal criteria.
    
    \item Gated recurrent unit (GRU) \citep{cho2014learning}. GRU is a variant of RNN that incorporates a gating mechanism. It features a forget gate, allowing only some information from previous states to pass to the subsequent states. GRU is designed to capture long-range dependencies in sequential data efficiently.
\end{itemize}

For the implementation of experiments, the Python programming language is utilized. The Scikit-Learn\footnote{\url{https://scikit-learn.org/stable/}} and PyTorch\footnote{\url{https://pytorch.org/}} libraries are employed to implement conventional machine learning algorithms and deep learning models, respectively. Pulp\footnote{\url{https://coin-or.github.io/pulp/}} and Mosek\footnote{\url{https://www.mosek.com/}} libraries are used to implement the optimization-based models. 

We conduct five-fold cross-validation in the real-case experiment and four simulations to determine the best parameters for the baselines and the proposed models. The best parameters are presented in Table \ref{tab-bestpara}.

\begin{table}[htbp]
    \centering
    \resizebox{1\textwidth}{!}{
    \begin{tabular}{clllllllllll}
    \hline
         & Stpl & Ptpl & mRNN & UTADIS & Logistic & SVM & RF & XGB & GRU & RNN & MLP  \\\hline 
       Real-case  & \makecell[l]{$\gamma=4$\\$C=0.01$} &\makecell[l]{$\gamma=4$\\$C=0.1$} & \makecell[l]{Dim $c=$40\\ \#para=3,162 } & $\gamma=4$ & $C=0.01$ & $C=1$ & \makecell[l]{n\_est=100\\max\_feat=6\\max\_depth=10} & \makecell[l]{n\_est=60\\max\_depth=2} & \makecell[l]{Dim $c=$32\\ \#para=4,737 } & \makecell[l]{Dim $c=$40\\ \#para=3,521 } & \makecell[l]{96-32-8-1\\ \#para=3,377 } \\ 
       Basic  & \makecell[l]{$\gamma=4$\\$C=0.001$} &\makecell[l]{$\gamma=4$\\$C=0.001$} & \makecell[l]{Dim $c=$40\\ \#para=3,162 } & $\gamma=4$ & $C=0.1$ & $C=1$ & \makecell[l]{n\_est=80\\max\_feat=8\\max\_depth=10} & \makecell[l]{n\_est=80\\max\_depth=4} & \makecell[l]{Dim $c=$32\\ \#para=4,737 } & \makecell[l]{Dim $c=$40\\ \#para=3,521 } & \makecell[l]{80-32-8-1\\ \#para=2,865 } \\
       
       Non-Markovian  & \makecell[l]{$\gamma=3$\\$C=0.001$} &\makecell[l]{$\gamma=3$\\$C=0.001$} & \makecell[l]{Dim $c=$40\\ \#para=3,162 } & $\gamma=3$ & $C=0.1$ & $C=1$ & \makecell[l]{n\_est=60\\max\_feat=8\\max\_depth=10} & \makecell[l]{n\_est=80\\max\_depth=2} & \makecell[l]{Dim $c=$32\\ \#para=4,737 } & \makecell[l]{Dim $c=$40\\ \#para=3,521 } & \makecell[l]{80-32-8-1\\ \#para=2,865 } \\
       Non-monotonic  & \makecell[l]{$\gamma=4$\\$C=0.01$} &\makecell[l]{$\gamma=2$\\$C=0.001$} & \makecell[l]{Dim $c=$40\\ \#para=3,162 } & $\gamma=4$ & $C=0.1$ & $C=1$ & \makecell[l]{n\_est=60\\max\_feat=10\\max\_depth=6} & \makecell[l]{n\_est=80\\max\_depth=2} & \makecell[l]{Dim $c=$32\\ \#para=4,737 } & \makecell[l]{Dim $c=$40\\ \#para=3,521 } & \makecell[l]{80-32-8-1\\ \#para=2,865 } \\
Non-independent  & \makecell[l]{$\gamma=4$\\$C=0.001$} &\makecell[l]{$\gamma=4$\\$C=0.001$} & \makecell[l]{Dim $c=$40\\ \#para=3,162 } & $\gamma=3$ & $C=1$ & $C=1$ & \makecell[l]{n\_est=80\\max\_feat=10\\max\_depth=8} & \makecell[l]{n\_est=80\\max\_depth=4} & \makecell[l]{Dim $c=$32\\ \#para=4,737 } & \makecell[l]{Dim $c=$40\\ \#para=3,521 } & \makecell[l]{80-32-8-1\\ \#para=2,865 } \\
       \hline
    \end{tabular}
    }
    \caption{Best parameters in real-case and simulation experiments.}
    \label{tab-bestpara}
\end{table}
\end{APPENDICES}

\bibliographystyle{informs2014} 
\bibliography{mybib.bib} 